\documentclass[10pt,twocolumn,letterpaper]{article}

\usepackage{iccv}
\usepackage{times}
\usepackage{epsfig}
\usepackage{graphicx}
\usepackage{amsmath}
\usepackage{amssymb}

\usepackage{float}

\usepackage{enumitem}

\usepackage{multirow}
\usepackage[normalem]{ulem}
\useunder{\uline}{\ul}{}

\usepackage{subfigure}
\usepackage{caption}
\usepackage{graphicx}

\usepackage[square,sort,comma,numbers]{natbib}
\usepackage{authblk}

\setcitestyle{comma}

\usepackage{placeins}

\usepackage[pagebackref=true,breaklinks=true,letterpaper=true,colorlinks,bookmarks=false]{hyperref}

\iccvfinalcopy 

\def\iccvPaperID{8099} 

\ificcvfinal\fi

\graphicspath{ {2020-09-15/fig/}, {2020-10-15/fig/}, {2020-11-15/fig/}, {2020-12-15/fig/}, {2021-2-17/fig/} }

\newcommand*{\affaddrMC}[1]{#1} 
\newcommand*{\affmarkMC}[1][*]{\textsuperscript{#1}}

\begin{document}

\title{\Large\bfseries SeeTheSeams: Localized Detection of Seam Carving based Image Forgery in Satellite Imagery}


\author{%
Chandrakanth Gudavalli\affmarkMC[1], Erik Rosten\affmarkMC[1], Lakshmanan Nataraj \affmarkMC[1], Shivkumar Chandrasekaran \affmarkMC[1, 2], and ~~~B. S. Manjunath\affmarkMC[1,2]\\
\affaddrMC{\affmarkMC[1]Mayachitra, Inc.}\\
\affaddrMC{\affmarkMC[2]Electrical and Computer Engineering Depatment, UC Santa Barbara}\\
\affaddrMC{Santa Barbara, California, USA}%
}



\maketitle
\ificcvfinal\thispagestyle{empty}\fi

\renewcommand{\thefootnote}{\fnsymbol{footnote}}

\begin{abstract}
\vspace{-0.3cm}
Seam carving is a popular technique for content aware image retargeting. It can be used to deliberately manipulate images, for example, change the GPS locations of a building or insert/remove roads in a satellite image. This paper proposes a novel approach for detecting and localizing seams in such images. While there are methods to detect seam carving based manipulations, this is the first time that robust localization and detection of seam carving forgery is made possible. We also propose a seam localization score (SLS) metric to evaluate the effectiveness of localization. The proposed method is evaluated extensively on a large collection of images from different sources, demonstrating a high level of detection and localization performance across these datasets. The datasets curated during this work will be released to the public. %
\vspace{-0.6cm}
\end{abstract}

\section{Introduction}
\label{sec:intro}

\begin{figure}[t]
\begin{center}
\newcommand*{\factor}{0.42}
\subfigure[]{ \includegraphics[height=\factor\columnwidth,keepaspectratio]{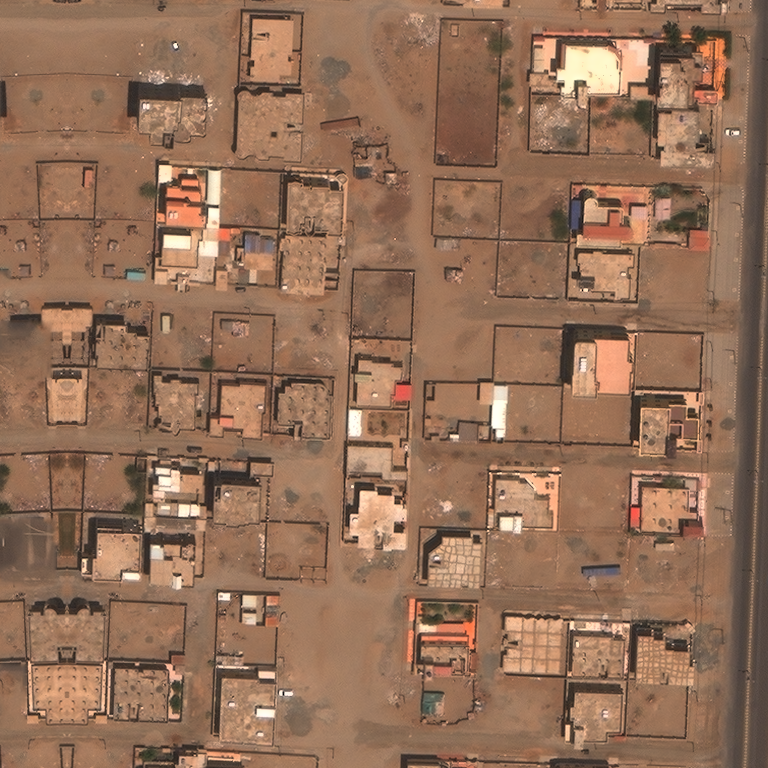}}
\subfigure[]{ \includegraphics[height=\factor\columnwidth,keepaspectratio]{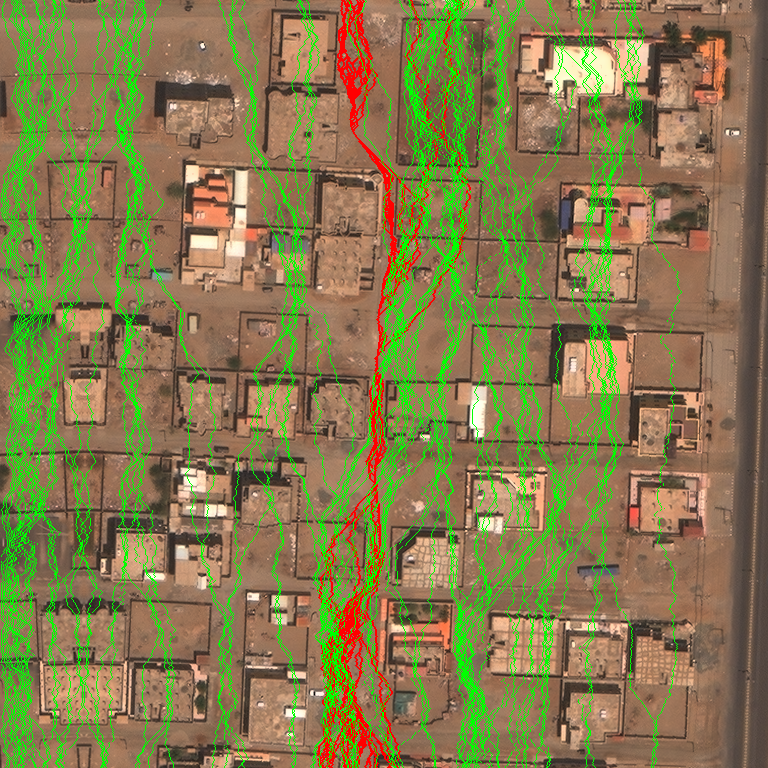}}
\newline
\vspace{-0.3cm}
\subfigure[]{ \includegraphics[height=\factor\columnwidth,keepaspectratio]{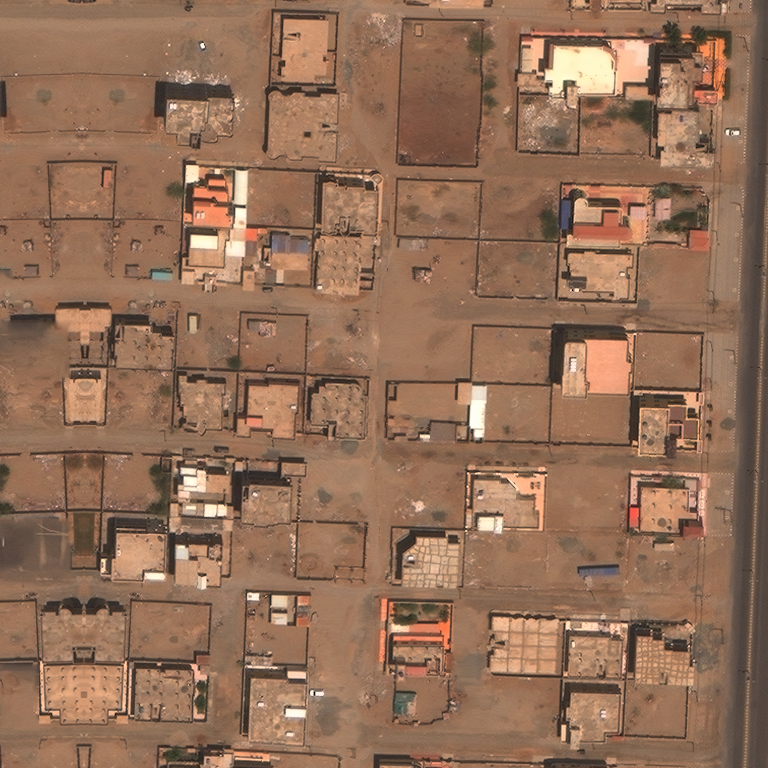}}
\subfigure[]{ \includegraphics[height=\factor\columnwidth,keepaspectratio]{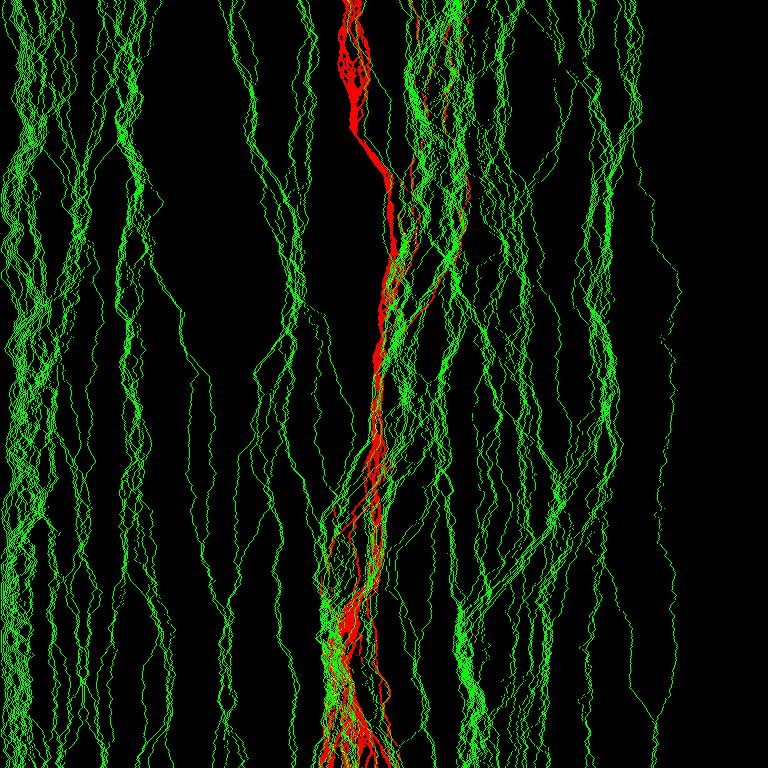}}
\newline
\vspace{-0.4cm}
\end{center}
\caption{ \textbf{(a)}. Pristine satellite image. \textbf{(b)}. Seam carved image with a few buildings removed overlaid with ground truth seam mask, where red seams indicate the pixels where seams are removed while green seams indicate the pixels where seams are inserted (red) seams. \textbf{(c)}. Seam carved image after removing a few objects (buildings). \textbf{(d)}. Seam prediction mask generated by our detector.}
\label{fig:xview1_obj_dislocation}
\vspace{-0.6cm}
\end{figure}

Seam carving is a popular image manipulation technique ~\cite{AS07,Rubinstein08,OTHER_SC_METHODS_SALIENCY} that is  effective for content aware image retargetting. In seam carving, the input image is resized by removing or inserting ``seams'' which are defined as connected pixel paths from top-to-bottom or left-to-right. These seams are chosen by their optimality according to an energy function computed for each pixel. That is, the optimal seam is the seam with the lowest energy along its path. Commonly used energy functions in seam carving are computed by measuring the contrast of a pixel with its neighbors. Removing an optimal seam has fewer artifacts in resized images than a randomly chosen seam, and protects image content that is highly textured. Seam carving can be extended to remove entire objects from images by assigning the energy of object pixels to a low value such that the seams forcibly pass through the object marked for removal. Since seam carving leaves a large percentage of pixel values untampered, it poses a challenge to image forgery detection.


Seam carving is included as a feature in popular image editing software such as GIMP~\cite{gimp_sc} and Photoshop~\cite{photoshop_sc}. The ease of access to these programs along with increasing availability of satellite image data from commercial satellites presents a growing problem for organizations that rely on accurate satellite data. Satellite images have been manipulated to influence public opinion such as in the Malaysia Airlines flight incident~\cite{sat_motivation_russia}, nighttime flyovers of India during the festivals~\cite{sat_motivation_india}, and fake spliced images of bridges~\cite{sat_motivation_china}. Furthermore, satellite imagery is particularly susceptible to seam carving based image manipulations. One reason is that in satellite imagery, objects of importance generally occupy fewer pixels than conventional images and consequently require fewer seams to remove, reducing the potential for visual artifacts in seam carved satellite images. In Figure~\ref{fig:xview1_obj_dislocation}, we show an example of object removal where we remove a set of buildings from the original image. Moreover, objects can be displaced to change their geographical coordinates (latitude, longitude) using seam carving much more easily in satellite imagery than typical images. Since satellite images are captured from high altitudes, they often have large, smooth regions that are ideal for seams to pass through, making them good candidates for object displacement while preserving visual image quality. In Figure~\ref{fig:top_level_example}, we show an example of object displacement accomplished by removing seams on one side of the building and inserting on the other. Displacing objects can be especially malicious in satellite imagery, where a single pixel can correspond with as much as $60m^{2}$ of ground area. In applications such as military target acquisition, this can be the difference between a successful and unsuccessful mission.

\begin{figure*}[!ht]
\begin{center}
\newcommand*{\factor}{0.40}
\subfigure[]{  \includegraphics[height=\factor\columnwidth,keepaspectratio]{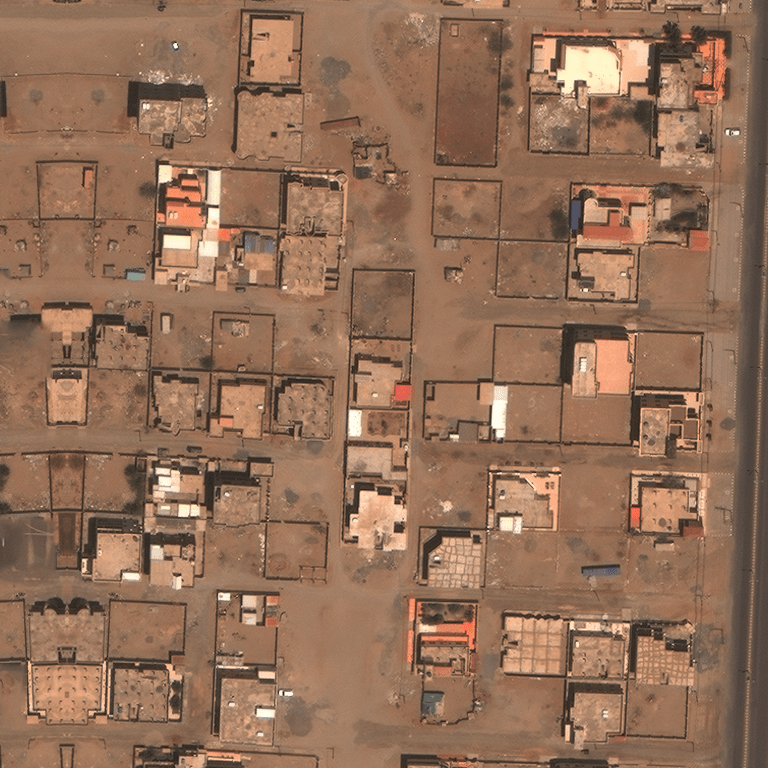}}
\subfigure[]{  \includegraphics[height=\factor\columnwidth,keepaspectratio]{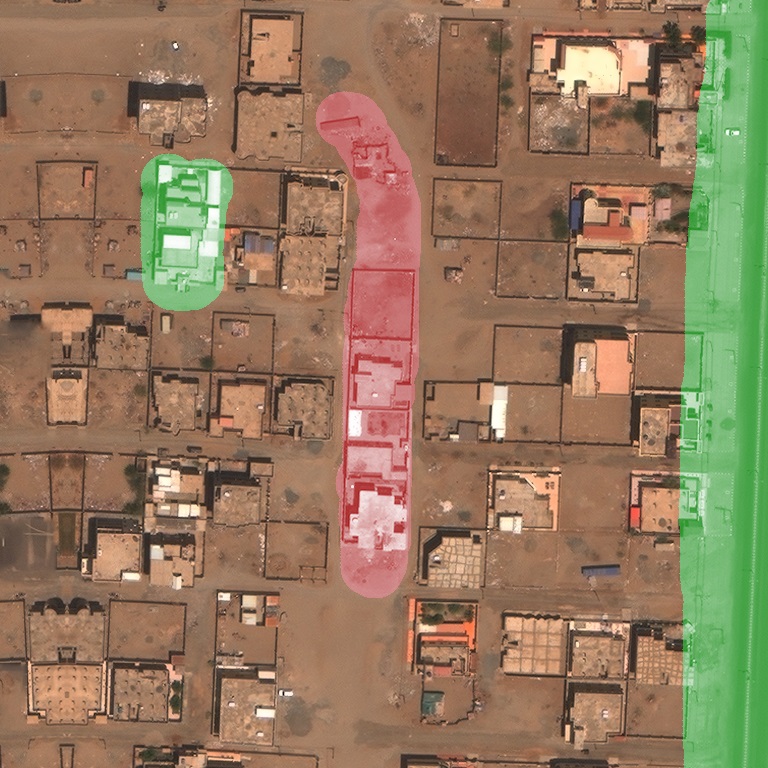}}
\subfigure[]{  \includegraphics[height=\factor\columnwidth,keepaspectratio]{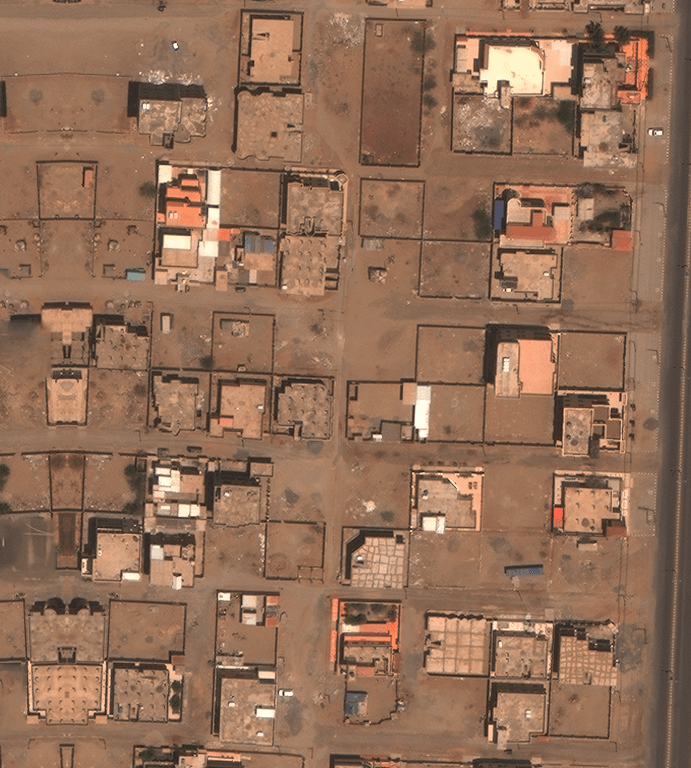}}
\subfigure[]{  \includegraphics[height=\factor\columnwidth,keepaspectratio]{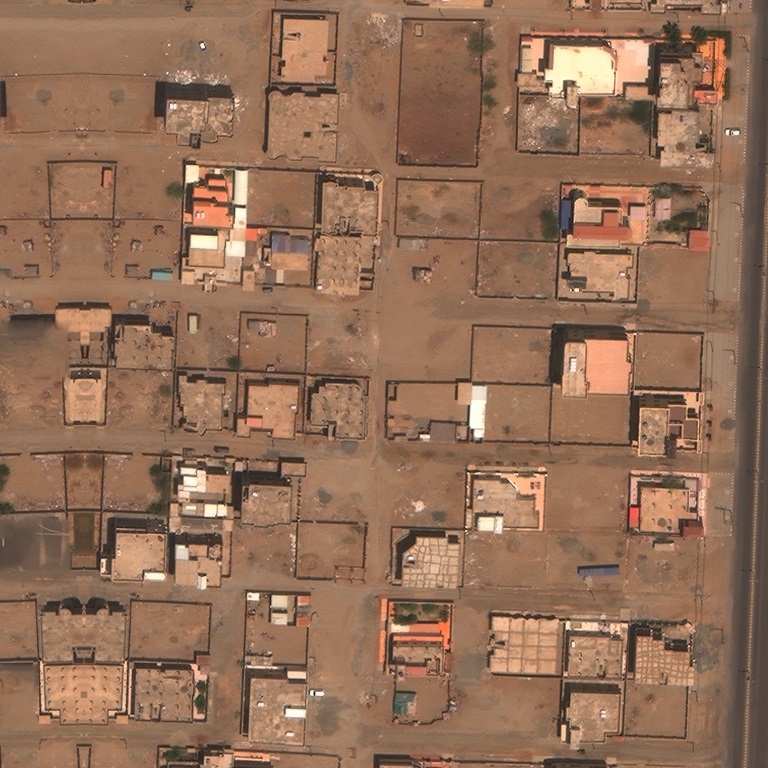}}
\subfigure[]{  \includegraphics[height=\factor\columnwidth,keepaspectratio]{SC-ICCV/images/xview1_obj_rem/gt_on_manip_enh.png}}
\vspace{-0.5cm}
\caption{ \textbf{(a)}. Original xView image \textbf{(b)}. Original image overlaid with masks on objects to be removed (red) and protected (green). \textbf{(c)}. Seam carved image with objects removed and the dimension of the image is reduced by 77 columns in the seam removal process. \textbf{(d)}. Final image after reinserting 77 seams into (c) to match dimensions of original image (a). \textbf{(e)}. Seam carved image overlaid with ground truth seam mask of inserted (green) seams and removal (red) seams.}
\label{fig:sc_obj_rem}
\end{center}
\vspace{-0.8cm}
\end{figure*}

While there are several works proposed to detect seam carving based image forgery~\cite{sarkar2009detection,lu2011seam,fei2015detection,seam_yin2015detecting,zhang2017detection,zhang2020detecting,cieslak2018seam,han2018exploring,ye2018convolutional,nam2019content,Nam2020DeepCN}, to our knowledge, none of these methods have investigated localization of removed and inserted seams at the pixel level, and their design is restricted to classification at the image level, or classification and localization at a patch level. We develop and evaluate our method on satellite imagery as a case study due to the potential ramifications of seam carving based manipulations. 



Towards addressing the above challenges, we propose a two stage, deep learning based seam carving detector with two key advantages over existing methods. The first advantage is the ability to \textit{localize the seams} at pixel level resolution. This is invaluable in discovering the extent of potential manipulations in satellite imagery. By considering the location of seams, one may be able to discover not only that an object may have been removed in an image, but where that removed object used to be. The second advantage is \textit{generalizability}. Since image level classifiers are trained on original satellite images, they are prone to become specialized to the distribution of data they are trained on. We demonstrate that our method is generalizable to not only training dataset distribution but different seam carving techniques. 
The main contributions of the paper are:
\vspace{-0.2cm}
\begin{enumerate}[wide]
\itemsep0em 
    \item We propose a method for robust detection of seam carving manipulations and accurate localization of seams removed or inserted by seam carving in satellite imagery.
    \item We develop the seam localization score (SLS), a specialized metric to better evaluate the localization performance of seam carving detectors on specific seams.
    \item We have created unique, large seam carving datasets that we plan to release to the public \footnote{Links to the dataset will be shared in near future.}.
\end{enumerate}

The rest of this paper is organized as follows. In Section~\ref{sec:rel-work} we review existing forensic approaches related to our work. Section \ref{sec:seam_carving_overview} gives an overview of seam carving and several variations on its original formulation. Section~\ref{sec:prop_method} details our proposed framework for localized detection of seam carving forgeries. Section~\ref{sec:datasets} describes the datasets that are curated to carry out our seam carving forensic experiments. Metrics that are used to evaluate our models are described in Section~\ref{sec:eval_mets}. Experimental setups and results are detailed in Section~\ref{sec:exp}. Finally, we conclude the paper in Section~\ref{sec:conclusion} by reviewing the pros and cons of our method and possible research areas that merit further exploration.
\vspace{-0.2cm}

\section{Related Work}
\label{sec:rel-work}
\vspace{-0.2cm}

Several works have been proposed to detect digital image manipulations (see~\cite{forgery_2,forgery_5} for an overview).
These works include the detection of specific image manipulations such as resampling ~\cite{popescu-farid-resampling, ryu2014estimation}, morphing~\cite{neubert2017face, jassim2018automatic}, copy-move ~\cite{li2015segmentation, cozzolino2015efficient}, splicing~\cite{guillemot2014image,salloum2018image}, seam carving~\cite{gong2018detecting,li2020identification}, and inpainting based object removal~\cite{wu2008detection,liang2015efficient}.
Several approaches also exploit JPEG blocking artifacts to detect tampered regions~\cite{farid2009exposing,luo2010jpeg}, while more recent efforts tend to exploit deep learning based approaches~\cite{bayar2016deep,rao2016deep}.

Most image forensics techniques that have been developed so far target consumer images~\cite{nonsat_bondi2016first,nonsat_bondi2017tampering}, which significantly differ in nature from satellite sensors (e.g. different compression schemes, color channels, orthorectification based post processing and so on).
Furthermore, it has been observed that many of these techniques do not perform well when naively applied to overhead images ~\cite{ali2017identification,sat_van2018you,sat_yarlagadda2018satellite,sat_bartusiak2019splicing,montserrat2020generative,fouad2020detection}.
To address this issue, several forensic techniques that work for satellite/overhead imagery have recently been proposed~\cite{sat_ho_1525212,sat_yarlagadda2018satellite,sat_horvath2019anomaly,sat_bartusiak2019splicing}.
One method uses handcrafted watermarks to detect manipulations in satellite images~\cite{sat_ho_1525212}. Although this method is quite effective, it cannot be utilized if the watermark is not inserted at the time of image acquisition by a trustworthy source. 
Another technique leverages conditional GANs to detect and localize splicing forgeries in satellite images by estimating a forgery mask~\cite{sat_yarlagadda2018satellite}.
A second GAN based technique~\cite{sat_horvath2019anomaly} encodes patches from an image into a low dimensional vector space that are used as input into a support vector machine (SVM) to detect the presence of forgeries at a patch level.
Finally, Sat-SVDD~\cite{sat_bartusiak2019splicing} is a kernel-based one-class classification method that detects splicing forgery with the help of support vector data description (SVDD).
In comparison to these methods, our paper explores the challenging case of detection and localization of seam carving based manipulations in satellite images. 

There have been several works proposed over the past decade to reveal traces of seam carving in digital images~\cite{sarkar2009detection,lu2011seam,fei2015detection,seam_yin2015detecting,zhang2017detection,zhang2020detecting,cieslak2018seam,han2018exploring,ye2018convolutional,nam2019content,Nam2020DeepCN}. 
These include methods using steganalysis~\cite{sarkar2009detection}, hashing~\cite{lu2011seam,fei2015detection}, local binary patterns (LBP)~\cite{seam_yin2015detecting,zhang2017detection,zhang2020detecting}, and deep learning based methods~\cite{cieslak2018seam,ye2018convolutional,nam2019content,Nam2020DeepCN}. 
In~\cite{han2018exploring}, the authors propose a method based on block artifacts to estimate the location of an object removed using seam carving. However, this method mainly focuses on finding a region in a seam carved image where an object has been removed and does not produce localization maps where seams have been removed/inserted. 
One deep learning method proposes a Convolutional Neural Network (CNN) based approach to perform image level classification using local binary patterns~\cite{cieslak2018seam}. 
Another CNN based approach uses a customized network that learns and uses more effective features via joint optimization of feature extraction and pattern classification~\cite{ye2018convolutional}. These two proposed methods perform image level classification to detect the presence of seam carving manipulation. One method that performs localization at a patch level employs a CNN called LFNet that is specifically designed to learn low-level features, capturing local artifacts from seam carving based image resizing~\cite{nam2019content}. The authors in~\cite{nam2019content} improved their network design with a new architecture, ILFNet~\cite{Nam2020DeepCN}. The authors in~\cite{nataraj2021seam} proposes a two stage model where stage-1 performs patch level localization and stage-2 performs image level classification. The method proposed in this paper localizes seams at a pixel level as opposed to the patch level strategies in prior work and focuses on satellite imagery, which is more resistant to seam carving artifacts than consumer images. 
In contrast to these proposed methods, our approach both detects images that are seam carved and also generates a seam localization mask which highlights pixels on the manipulated image where seams have been carved or inserted.




\vspace{-0.4cm}
\section{Overview of Seam Carving}
\label{sec:seam_carving_overview}
\vspace{-0.2cm}

Seam carving is a content-aware image retargetting algorithm proposed in 2007~\cite{AS07}. A seam is defined as an optimal 8-connected path of pixels from top to bottom or left to right. The optimality of a seam is determined by assigning an energy value to each pixel in a given image. One possible energy function proposed in \cite{AS07} is the gradient magnitude, also termed as \emph{backward energy}.

\vspace{-0.2cm}
\begin{equation}
\label{eqn:1}
    E(\boldsymbol{I}) = \sqrt{\frac{\partial^{2} \boldsymbol{I}}{\partial x^{2}} + \frac{\partial^{2} \boldsymbol{I}}{\partial y^{2}}}
\vspace{-0.2cm}
\end{equation}

\noindent where $\boldsymbol{I}(x,y)$ is an image indexable by row $x$ and column $y$, and $E(\boldsymbol{I})$ is the energy map of $\boldsymbol{I}$. Then, the optimal vertical seam for an $m$ by $n$ image can be defined as 

\vspace{-0.7cm}
\begin{align}
    s^{x} &= \{s_{i}^{x}\}_{i = 1}^{m} = \{i, x(i)\}_{i=1}^{m} \nonumber\\ &\text{s.t.} \; \;  \forall i, \;\; \left| x(i) - x(i-1) \le 1\right|
\vspace{-0.5cm}
\end{align}

\vspace{-0.2cm}
The optimal seam is then the seam that minimizes the energy function over all possible seams,

\vspace{-0.35cm}
\begin{equation}
    s^{*} = \min_{\boldsymbol{s}} E(\boldsymbol{s}) = \min_{\boldsymbol{s}} \sum_{i=1}^{m} E(\boldsymbol{I}(\boldsymbol{s}))
\end{equation}
\vspace{-0.3cm}

This optimal vertical seam can be found through dynamic programming, computing a minimum cumulative energy matrix $M(x,y)$ for all possible seams through $(x,y)$ by traversing from the second row to the last row:

\vspace{-0.6cm}

\begin{equation*}
    M(x,y) = e(x,y) + \min \begin{cases}
&M(x-1,y-1)  \; \nonumber \\ 
&M(x-1, y)  \; \\ 
&M(x-1, y+1)  \;  \nonumber \end{cases}
\vspace{-0.15cm}
\end{equation*}
where $e(x,y)$ is an optional additional energy measure. 


Once the cumulative energy matrix has been computed, the minimum value in the last row of $M$ gives the index of the end of the optimal seam. The rest of the seam can be found by backtracking through $M$. 

Once we find the optimal seam, we can simply remove it to reduce the width or height of an image. To increase the dimensions of an image, we take the average of the optimal seam and a seam to the immediate right and insert it at the location of the optimal seam. By successively removing and inserting optimal seams, images can be resized without modifying many of the original pixel values of the image. These removed seams have low energy according to Eqn.~\ref{eqn:1}, and often correspond to smooth regions in the image, visually preserving image content and structure. 

Seam carving can be applied to remove objects by setting energy map values from a user-provided mask to a ``low'' value, and successively removing seams until the object is removed. The user can insert seams to restore the original image size, as well as preserve objects through a mask that tags regions to have a ``high'' energy. Figure~\ref{fig:sc_obj_rem} is one such example of this \emph{object removal} strategy and perceptually it is very difficult for a human to deduce that objects have been removed in this image.
Seam carving can also be utilized for \emph{object displacement} in a satellite image by marking the region on one side of an object for seam removal and then inserting seams on the other side of the object after the seams have been removed (Figure~\ref{fig:top_level_example}).
In this way, objects can be ``shifted'' to a different GPS location with very little perceptual difference.


\vspace{-0.2cm}
\subsection{Other Seam Carving Methods}
\label{sec:other_methods_seam_carving_overview}
\vspace{-0.2cm}

There have been several modifications to the original seam carving algorithm,. One such modification was proposed in 2008~\cite{Rubinstein08} that considers the energy introduced into the resized image by bringing previously non-adjacent pixels together by removing seams. This formulation, termed as \emph{forward energy}, computes the energy map by looking at the differences in pixel values depending on the direction of the potential seam, defining three possible cases: up, up and to the left, and up and to the right.
\vspace{-0.2cm}
\begin{small}
\begin{align*}
    C_{L} &= \left| \boldsymbol{I}(x,y+1) - \boldsymbol{I}(x,y-1) \right| + \left| \boldsymbol{I}(x-1,y) - \boldsymbol{I}(x, y-1)\right| \\
    C_{U} &= \left| \boldsymbol{I}(x,y+1) - \boldsymbol{I}(x,y-1) \right| \\
    C_{R} &= \left| \boldsymbol{I}(x,y+1) - \boldsymbol{I}(x,y-1) \right| + \left| \boldsymbol{I}(x-1,y) - \boldsymbol{I}(x, y+1)\right|
\end{align*}
\vspace{-0.25cm}
\end{small}

\noindent Then, the cumulative energy matrix can be updated as: 
\vspace{-0.2cm}
\begin{equation*}
    M(x,y) = e(x,y) + \min \begin{cases}
&M(x-1,y-1) + \; C_{L}(x,y) \nonumber \\ 
&M(x-1, y) + \; C_{U}(x,y)\\ 
&M(x-1, y+1) + \; C_{R}(x,y) \nonumber \end{cases}
\vspace{-0.15cm}
\end{equation*}
where $e(x,y)$ is an optional additional energy measure. 

Another seam carving variation proposes using an importance map based on \emph{salient region detection}~\cite{OTHER_SC_METHODS_SALIENCY}. In this particular formulation, salient regions are uniformly identified considering global contrast as opposed to local edges in the two original methods outlined above. This prevents the necessity to recompute the energy map after each optimal seam is found. The saliency map used in place of the energy map also incorporates color information, which the two methods above omit. To calculate the saliency map, the image is converted into $Lab$ color space. Then, the final map is the Euclidean distance between the average pixel vector and a Gaussian blurred version, approximated by a 5 x 5 binomial filter (both in $Lab$ space):
\vspace{-0.15cm}
\begin{equation}
    E_{Lab}(x,y) = || \boldsymbol{I}_{\mu} - \boldsymbol{I}_{Gauss}(x,y)||
    \vspace{-0.15cm}
\end{equation}

The final variation of seam carving that we explore is called \emph{seam merging}~\cite{mishiba2012image}. 
This method merges a two-pixel-width seam element into one new pixel during image reduction and inserts a new pixel between the two pixels during image enlargement. This algorithm utilizes importance and structure energies to define seam optimality, as well as an additional energy term that suppresses artifacts generated by excessive reduction or enlargement from repeated merging or inserting.



\begin{figure*}[t]
    \centering
    \includegraphics[width=0.99\textwidth]{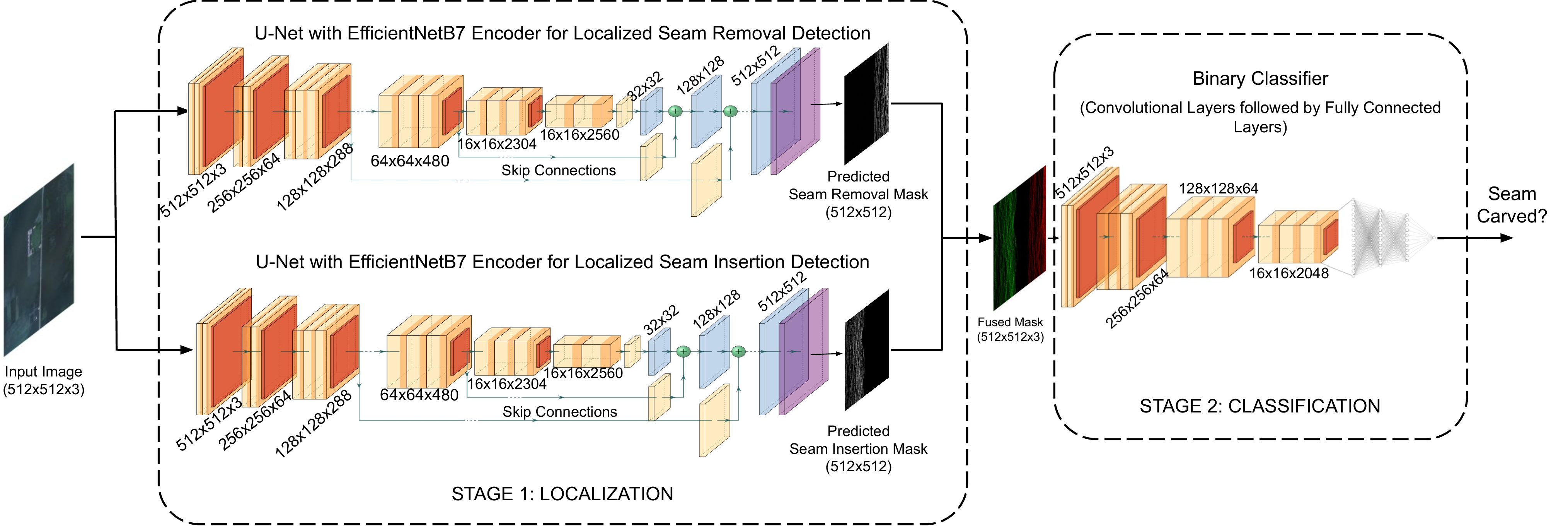}
    \caption{Overview of proposed two stage framework for localized seam carving detection.}
    \label{fig:overview}
\end{figure*}

\vspace{-0.3cm}
\section{Proposed Approach}
\label{sec:prop_method}
\vspace{-0.1cm}
Here, we describe our approach for localized seam detection.
At its core, the problem can be formulated as binary classification, where we would ultimately like to predict whether the given image is seam carved or not.
In order to achieve this, we propose a two stage framework.
In Section~\ref{sec:stage1}, we develop localization models that flag regions of the image that are seam carved. 
In Section~\ref{sec:stage2}, we perform a final classification whether the image has been manipulated by seam carving by fusing the outputs of stage 1. 

\vspace{-0.2cm}
\subsection{Stage 1 - Localization of Seams}
\label{sec:stage1}
\vspace{-0.2cm}



In order to obtain localized detection of seams in an image we implement a fully convolutional network to learn a mapping from satellite image to seam mask, where the seam mask contains the locations of removed or inserted seams as described in Section \ref{sec:gt_mask_generation}. U-Net has been extensively used in image segmentation ~\cite{UNET}, and its architecture is composed of a contracting path (encoder) followed by an expanding path (decoder). Encoder maps the input image to a feature vector, which will then be used by the decoder to flag seam carved regions of the input image.
The basic encoder-decoder model's localization power is enhanced in U-Net by applying skip connections between encoder feature maps and decoder outputs.
We chose EfficientNetB7~\cite{efficientnet} as our baseline network since it performed the best when compared to other standard networks (see Section~\ref{sec:stg1_pxl_wise} for comparative experiments) 
EfficientNet is a baseline network composed of sequential mobile inverted bottleneck convolutions (MBConv) blocks that can be scaled up to improve accuracy at the cost of increased computation.
This generates a family of models from EfficientNetB0 to EfficientNetB7 with the EfficientNetB7 model being the most accurate, but also computationally the most expensive. 
EfficientNetB7 also outperforms other encoder architectures when used for pixelwise classification, such as in~\cite{baheti2020eff}.
For the decoder architecture, we use four sets of transposed convolutional, batch norm, dropout, and ReLU layers to upsample encoded features back to the size of input image.
In Figure~\ref{fig:overview}, we show an overview of our proposed method, where we see that the localization stage is comprised of two pixelwise classifiers - one model trained for seam removal detection and another model trained for seam insertion detection, both using U-Net and EfficientNetB7.
On a test image, these two models output predicted seam localization masks flagging the regions where seams have been removed and inserted.


\vspace{-0.2cm}
\subsection{Stage 2 - Classification}
\label{sec:stage2}
\vspace{-0.2cm}
The pixelwise classifiers described above output two seam prediction masks with the same dimensions as the input image, one localizing removed seams and the other inserted seams. 
Both these masks are concatenated and fed as input to a standard CNN (ResNet50~\cite{resnet}) to perform vanilla binary classification to identify if an image has been seam carved or not. 

\vspace{-0.2cm}
\section{Datasets}
\label{sec:datasets}
\vspace{-0.2cm}
In this section, we give a brief overview on the characteristics of source datasets used, and how we generated our forgery datasets. 
Three common satellite imagery datasets (xView~\cite{XVIEW}, xBD~\cite{XVIEW2}, and Orbview-3\cite{orbview3}) have been used to evaluate our method. The xView dataset contains $1,127$ images at varying high resolutions. Using these high resolution images, we generated a dataset of $53,943$ images by randomly cropping $512$ x $512$ regions from the original dataset. When dividing the images into a train-test-val split, all $512$ x $512$ images from a given xView sample are assigned to the same split. For xView, we allocated 70\% of the images into training set, 15\% into validation set, and remaining 15\% into test set. The xBD dataset contains a total of $22,098$ pre-disaster and post-disaster RGB images of size $1024$ x $1024$, where we've preserved the original dataset's train-test-val split of $80$:$10$:$10$ and randomly cropped $512$ x $512$ regions as with the xView dataset. The xBD dataset also contains ground truth building masks that can be used in seam carving object removal. These building masks are quite small and make up an average of $0.37\%$ of all pixels in a particular training set image. On average, a building mask occupies $916 \; pixels^{2}$. Since the xBD dataset contains satellite images at $0.3m$ per pixel, these building objects correspond with around $82m^{2}$ of ground area despite taking up a small number of pixels. In the most extreme cases, a building mask can take as little as $0.9m^{2}$ up to $15,943m^{2}$ of ground area. Finally, we curated a third dataset consisting of $48,000$  Orbview-3 images by randomly selecting non overlapping $512$ x $512$ regions. The train-test-val split is again maintained to be $80$:$10$:$10$. While the xView and xBD datasets contain 8 bit RGB images (image intensities are in range $0$ to $255$), the Orbview-3 dataset has single channel 16 bit images. These three datasets provide variation in the geographical location of the images as well as different levels of color depth. 



\vspace{-0.2cm}
\subsection{Ground Truth Seam Masks}
\label{sec:gt_mask_generation}
\vspace{-0.2cm}
Ground truth seam masks for training pixelwise classifiers are generated while seam carving original dataset images. As we remove seams from the original image, all the preserved pixel locations that were adjacent to removed seams are marked, indicating manipulation. As we insert seams into the image, all the pixel locations of the inserted seam are flagged. A visual example of a ground truth seam mask is shown overlaid on the resulting seam carved image in Figure \ref{fig:xview1_obj_dislocation}b.

\vspace{-0.2cm}
\subsection{Pixelwise Classification Datasets}
\label{ssec:data_pxl_wise_clf}
\vspace{-0.2cm}
Datasets for pixelwise classification are generated by seam carving pristine samples to remove the top 10\% of optimal seams and inserting seams to restore the original image dimension. All of our models are trained on $512$ x $512$ images, cropped from top left of the seam carved images. Ground truth seam masks are generated as described in Section~\ref{sec:gt_mask_generation} and similarly cropped. For the remainder of the paper if unspecified, forward energy is used to define the optimality of seams and examine the generalization capabilities of our models across several seam carving variants in Section~\ref{sec:generalizability_sc_techniques}.

\vspace{-0.2cm}
\subsection{Stage 2 Classification Datasets}
\vspace{-0.2cm}
Since we train models from each stage independently, stage 2 datasets are easily generated once pixelwise classifier training is complete. We obtain the prediction masks from trained model inference on cropped seam carved and original images to form a dataset of manipulated and pristine samples required for the image level classification.

\vspace{-0.3cm}
\section{Evaluation Metrics}
\label{sec:eval_mets}
\vspace{-0.2cm}
In this section, we briefly describe the evaluation metrics used to select and assess our models. 
Since we're operating in a binary classification setting at a pixel level in the localization stage and at image level in the classification stage, we use metrics based on confusion matrices. For a dataset of $N$ samples, the $i$-th sample will have a confusion matrix:

\vspace{-0.15cm}
\begin{equation}
    CM_{i} = \begin{bmatrix}
    TP_{i} & FP_{i} \\
    FN_{i} & TN_{i} \\
    \end{bmatrix}
    \vspace{-0.1cm}
\end{equation}

\noindent where TP is the number of true positives, FP is the number of false positives, FN is the number of false negatives, and TN is the number of true negatives. Then, the pixelwise accuracy for the $i$-th image is 

\begin{equation}
    \text{Accuracy}_{i} = \frac{TP_{i} + TN_{i}}{TP_{i} + TN_{i} + FP_{i} + FN_{i}}
\end{equation}

\noindent The confusion matrix and the accuracy over the entire dataset is computed from cumulative confusion matrices for every sample.
However, since seam carving based manipulations tend to remove less than 10 percent of the pixels from source datasets, pixelwise accuracy is an inadequate representation of the performance of our models as a naive method predicting all negatives will achieve above 90\% accuracy. 
To address the inherent imbalance in our generated datasets, we prioritize several more relevant confusion matrix derived metrics. For the rest of this section, we report these metric formulae at the dataset level, knowing that we can compute image level statistics as needed. A few metrics that attempt to address class imbalance are precision, recall and F1 score. 
\vspace{-0.15cm}
\begin{equation}
    \text{Precision} = \frac{TP}{TP+FP} \; \; \; \; \; \; \text{Recall} = \frac{TP}{TP + FN}
    \vspace{-0.2cm}
\end{equation}


\begin{equation}
    \text{F1 Score} = 2 \cdot \frac{\text{Precision} \cdot \text{Recall}}{\text{Precision} + \text{Recall}}
\end{equation}

While useful, these three metrics are inherently biased towards the positive class, and independent of the number of true negatives.
For seam carving localization, we would like to not only incorporate how close the predicted seams are to the true seams, but measure the efficacy of our model on correctly identifying untampered regions. 
One metric that satisfies this specification is the Matthews Correlation Coefficient (MCC), a balanced measure ranging between $\pm1$ that can be used regardless of the degree of class imbalance in a dataset due to its invariance to the choice of which class is positive or negative~\cite{MCCADVANTAGES}. 
It should be noted that in all cases where cropping resulted in all ground truth negatives for a particular sample, all metrics aside from pixelwise accuracy are set to 0.

\vspace{-0.6cm}
\begin{equation}    
\resizebox{.88\hsize}{!}{$
    \text{MCC} = \frac{TP \cdot TN - FP \cdot FN}{\sqrt{(TP+FP)(TP+FN)(TN+FP)(TN+FN)}}$}
    \vspace{-0.2cm}
\end{equation}


\begin{figure}[t]
\begin{center}
\newcommand*{\factor}{0.42}
\subfigure[]{ \includegraphics[height=\factor\columnwidth,keepaspectratio]{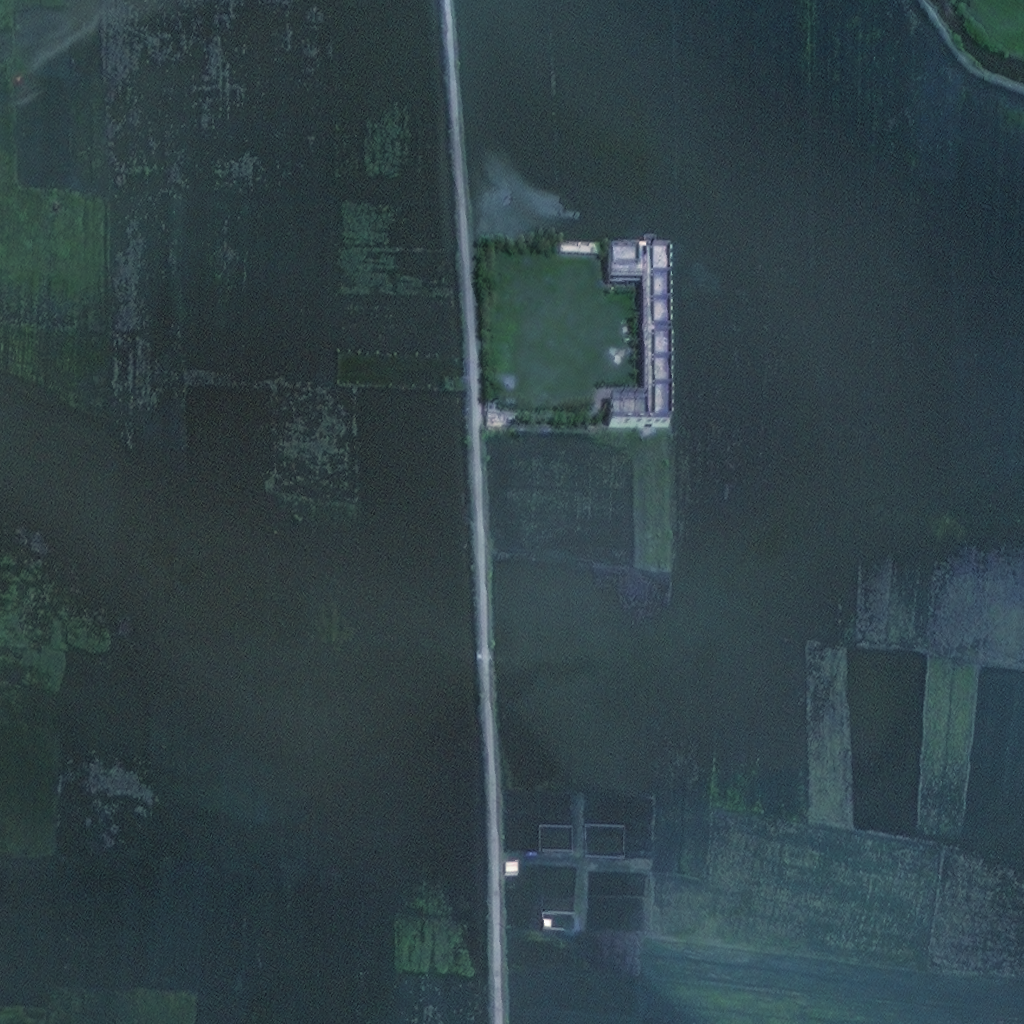}}
\subfigure[]{ \includegraphics[height=\factor\columnwidth,keepaspectratio]{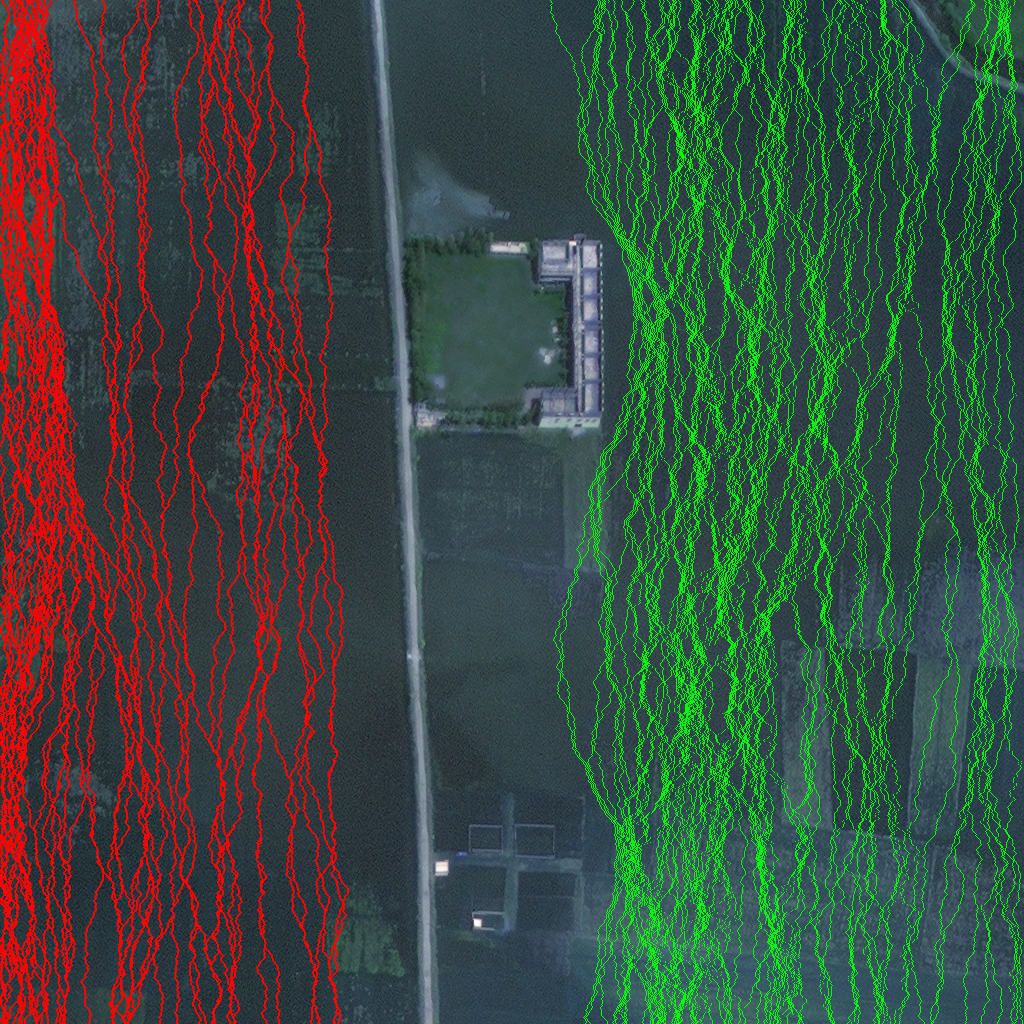}}
\vspace{-0.3cm}
\newline
\subfigure[]{ \includegraphics[height=\factor\columnwidth,keepaspectratio]{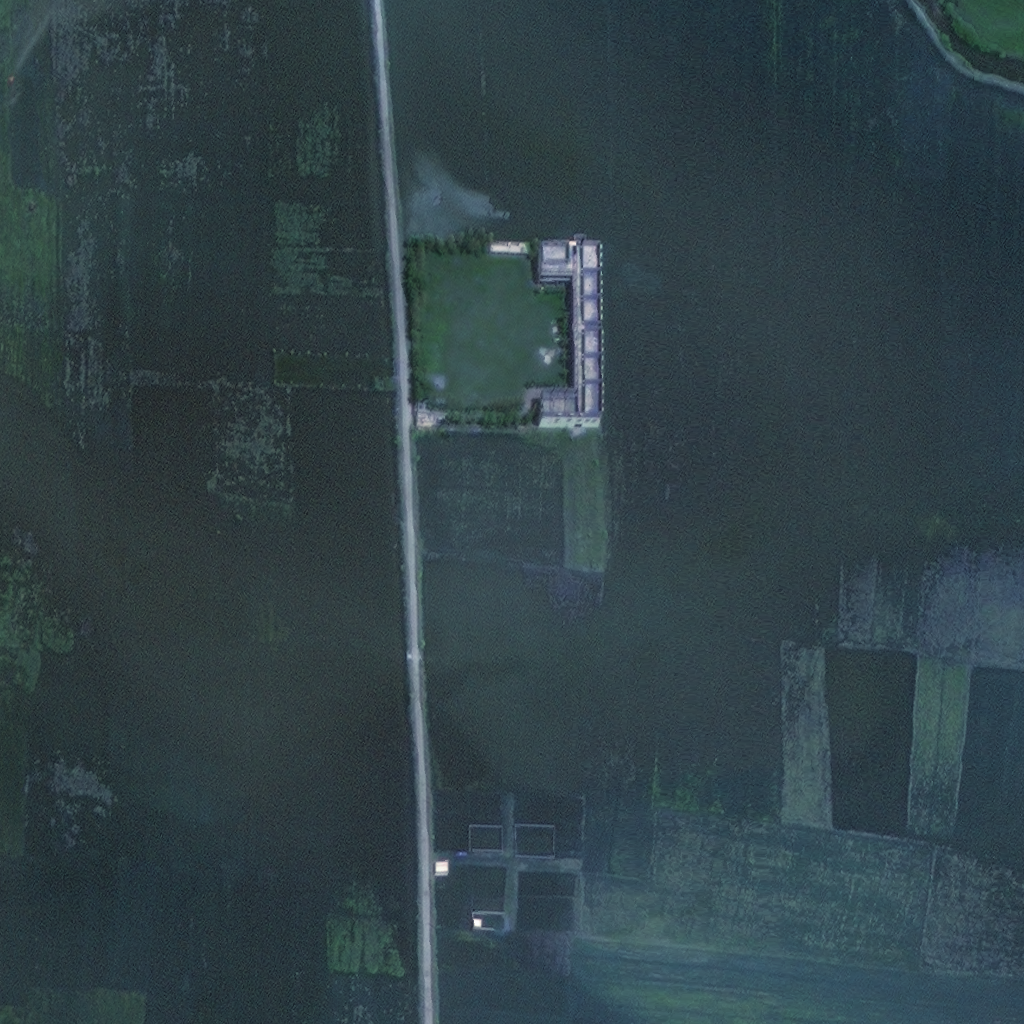}}
\subfigure[]{ \includegraphics[height=\factor\columnwidth,keepaspectratio]{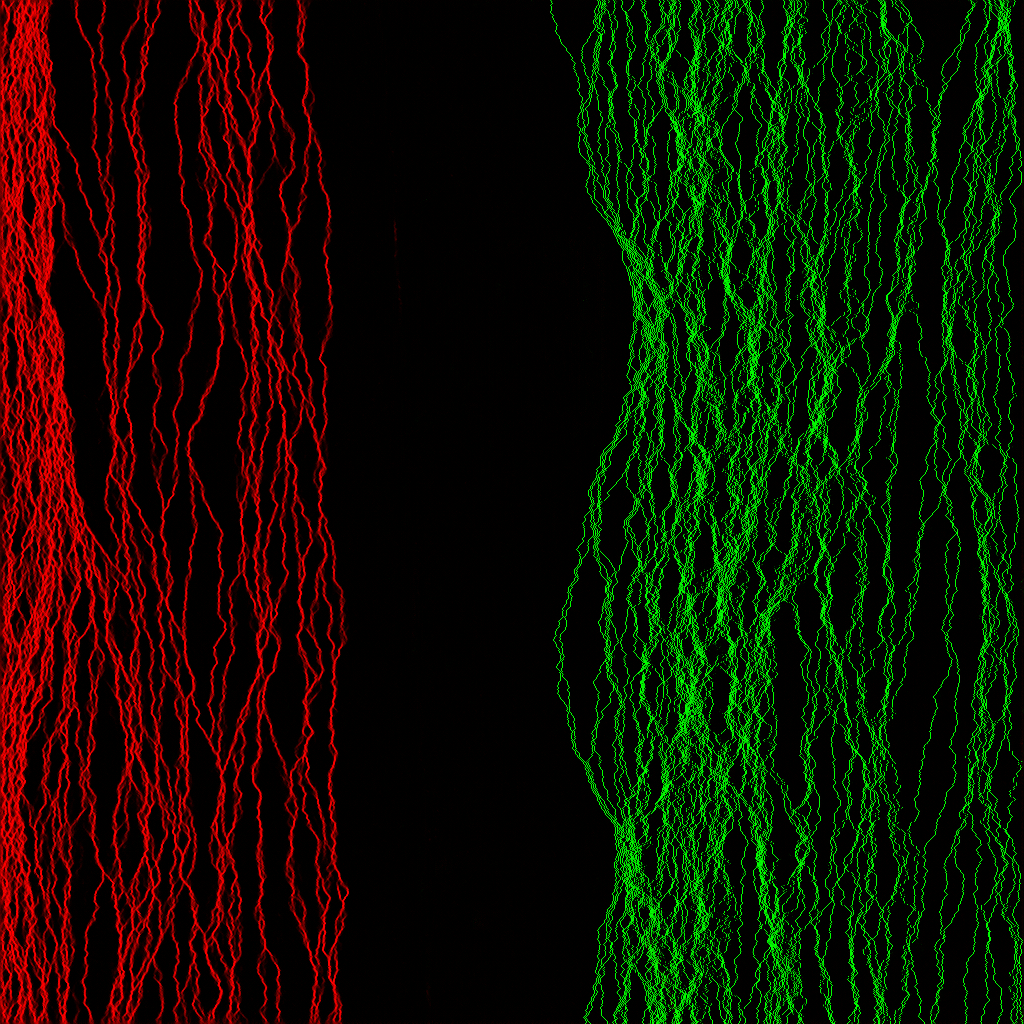}}
\vspace{-0.8cm}
\newline
\end{center}
\caption{ \textbf{(a)}. Pristine satellite image marked for ``object displacement''. \textbf{(b)}. Seam carved image overlaid with ground truth seam mask with removal seams marked in red and inserted seams marked in green. \textbf{(c)}. Seam carved image indicating ``object displacement'' where an entire strip in the center of the image has been displaced by a few pixels to the left. \textbf{(d)}. Prediction mask generated by our detector highlighting the areas where seams have been removed and inserted.}
\label{fig:top_level_example}
\vspace{-0.6cm}
\end{figure}

\vspace{-0.2cm}
\subsection{Customizing Confusion Matrix Metrics}
\vspace{-0.2cm}
While these confusion matrix metrics are widely used to evaluate binary classification performance, we make a slight adjustment to the way we calculate confusion matrices (and their derived metrics) for our specific use-case. Since seams are only one-pixel wide and our confusion matrices are calculated on a pixel-wise basis, they are extremely sensitive to the spatial distribution of the prediction mask. For example, take the seam insertion mask from Figure~\ref{fig:top_level_example}b. If we shift all the seams one pixel to the right and compare the shifted mask to the original, our confusion matrix metrics break down and report poor results. Specifically, the recall between the original and shifted masks becomes $0.194$ when the shifted version is in fact localizing the seams quite well. 

To properly represent the performance of our model's predictions, we modify the confusion matrix calculation such that we assign a true positive if our prediction is within a buffer of $p$ pixels of a ground truth positive. This relaxation is similar to a metric used in evaluating road detectors, especially in the context of aerial imagery~\cite{mnih2010learning, wegner2013higher}. It's worth noting that if we assign a true positive in this way, we do not double count the ground truth positive used to evaluate predicted negative pixels. If we predict negatively and the ground truth at that location is positive but has been used, we assign it as a true negative. We employ the strategy described here with $p=1$, and for the particular example in this section the recall between the original and shifted masks becomes $1.0$, due the $1$ pixel buffer described above. To make this clear, we refer to these customized metrics as MCC-1, F1 Score-1, Precision-1, and Recall-1 throughout the remainder of this paper.

\vspace{-0.2cm}
\subsection{Seam Localization Score (SLS)}
To fully capture the localization performance of our models, we develop a metric based on the seams that are inserted or removed. By keeping track of the seams associated with manipulated pixels, we can evaluate how well our model localizes each specific seam. For a particular vertical seam $s$ taken from an image of height $h$ and width $w$, the corresponding ground truth seam mask will associate $h$ pixels with seam $s$. Then, we sum over each seam pixel from $1:h$, the absolute distance to the nearest predicted positive in each row. If we do not predict a positive pixel in a particular row, the absolute distance is set to $w$. Finally, we normalize by the number of rows $h$. We refer to this metric the seam localization score (SLS). For an image with $N$ seams we compute an image level SLS by summing the score for each seam, and dividing by $N$. The SLS for any particular seam ranges from 0, perfect overlap, to $w$. When the SLS for a particular seam is less than one, we can interpret that on average, the seam was less than one pixel away from its ground truth location. 



\begin{table}[t]
\begin{center}
\begin{tabular}{|c|c|c|c|}
\hline
\textbf{\begin{tabular}[c]{@{}c@{}}Encoder \\ Architecture\end{tabular}} & \textbf{F1Score-1} & \textbf{MCC-1} & \textbf{SLS} \\ \hline
MobileNetV2 & 0.779 & 0.776 & 5.95 \\ \hline
ResNet50 & 0.631 & 0.651 & 17.93 \\ \hline
ResNet101 & 0.603 & 0.628 & 23.73 \\ \hline
\textbf{EfficientNetB7} & \textbf{0.911} & \textbf{0.903} & \textbf{1.56} \\ \hline
\end{tabular}
\end{center}
\vspace{-0.5cm}
\caption{Performance of seam removal detectors with various encoder architectures, trained and tested on xView.}
\label{tab:pixel_wise_stage1_res}
\vspace{-0.2cm}
\end{table}

\section{Experimental Results}
\label{sec:exp}
Here, we describe the experiments that are carried out towards localized seam carving detection. First, we cover our multistage approach towards localized detection of seam carving, using pixelwise classifiers, and then present the performance of localization models using the metrics detailed in Section \ref{sec:eval_mets}. Then, we demonstrate the generalizability of our model across different training distributions and seam carving algorithms.

\begin{table}
\begin{center}
\begin{tabular}{|c|c|c|}
\hline
\textbf{Dataset} & \textbf{MCC-1} & \textbf{F1Score-1} \\ \hline
xView            & 0.956          & 0.959              \\ \hline
xBD              & 0.938          & 0.935              \\ \hline
OrbView-3         & 0.942          & 0.943              \\ \hline
\end{tabular}
\end{center}
\caption{Generalizability of a seam insertion pixelwise classifier trained on xView, and tested on all three datasets.}
\label{tab:pixel_wise_stage1_res_si_detect}
\end{table}

\begin{table}[h]
\begin{center}
\begin{tabular}{c|c|c|c|}
\cline{2-4}
\textbf{} &
  \textbf{\begin{tabular}[c]{@{}c@{}}Tested on \\ xView\end{tabular}} &
  \textbf{\begin{tabular}[c]{@{}c@{}}Tested on \\ xBD\end{tabular}} &
  \textbf{\begin{tabular}[c]{@{}c@{}}Tested on \\ Orbview-3\end{tabular}} \\ \hline
\multicolumn{1}{|c|}{\textbf{{\begin{tabular}[c]{@{}c@{}}Trained on \\ xView\end{tabular}}}}    & 0.903 & 0.821 & 0.711 \\ \hline
\multicolumn{1}{|c|}{\textbf{\textbf{{\begin{tabular}[c]{@{}c@{}}Trained on \\ xBD\end{tabular}}}}}    & 0.722 & 0.894 & 0.708 \\ \hline
\multicolumn{1}{|c|}{\textbf{\textbf{{\begin{tabular}[c]{@{}c@{}}Trained on \\ Orbview-3\end{tabular}}}}} & 0.721 & 0.726 & 0.889 \\ \hline
\end{tabular}
\end{center}
\caption{MCC-1 scores of EfficientNetB7 seam removal pixelwise classifiers, trained and tested on different datasets.}
\label{tab:pixel_wise_stage1_gen_across_datasets}
\end{table}

\subsection{Pixelwise Classification}
\label{sec:stg1_pxl_wise}
To achieve more fine-grained localization of seams and and incorporate the spatial relationships of input images, we train a step-down, step-up fully convolutional neural network based on U-Net as described in Section~\ref{sec:stage1} to detect removed and inserted seams at pixel-level resolution (Figure~\ref{tab:pixel_wise_stage1_res}). 
We report stage 1 evaluation metrics from Section~\ref{sec:eval_mets} using a variety of encoder architectures in Table~\ref{tab:pixel_wise_stage1_res} on our xView test dataset, as generated in Section~\ref{ssec:data_pxl_wise_clf}. The best performing model used an EfficientNetB7 encoder architecture, achieving a MCC-1 score of 0.903. We note that the performance increase gained by using EfficientNet cannot be solely due to the increase in model capacity. The second best performing network was MobileNetV2, which has the smallest number of trainable parameters among all networks tested.
Our seam insertion detector of the same EfficientNetB7 architecture, achieves a MCC-1 score of 0.956 on the xView test set. We also observe that although trained on xView images, the seam insertion detector generalizes well to xBD and Orbview-3 datasets with an MCC-1 of 0.938 and 0.942 respectively.

\vspace{-0.2cm}
\subsection{Generalizability of Pixelwise Classifiers}
\label{sec:generalizability_sc_techniques}
\vspace{-0.2cm}
Here, we summarize several experiments demonstrating the generalizability of pixelwise classifiers across datasets and various seam carving techniques. Pixelwise classifiers trained on our xView dataset work well on xBD and Orbview-3 datasets and vice versa. Moreover, even though pixelwise classifiers are trained with datasets forged using forward energy, they are generalizable to different seam carving techniques.
\vspace{-0.3cm}
\subsubsection{Generalizability Across Datasets}
\vspace{-0.2cm}
\label{sec:generalizability_datasets}
Table \ref{tab:pixel_wise_stage1_res_si_detect} shows that pixelwise classifiers for seam insertion detection are generalizable across different datasets. A seam insertion detector trained on xView has only minor drops in performance on xBD and Orbview-3 test sets. 

\begin{table*}[ht]
\begin{center}
\begin{tabular}{|c|c|c|c|c|}
\hline
\multirow{2}{*}{\textbf{\begin{tabular}[c]{@{}c@{}}Seam carving techniques\end{tabular}}} &
  \multicolumn{4}{c|}{{\ul \textbf{Evaluation Metrics}}} \\ \cline{2-5} 
 &
  \textbf{Precision-1} &
  \textbf{Recall-1} &
  \textbf{F1Score-1} &
  \textbf{MCC-1} \\ \hline
\textbf{\begin{tabular}[c]{@{}c@{}}Backward Energy~\cite{AS07}\end{tabular}} &
  0.936 &
  0.834 &
  0.882 &
  0.874 \\ \hline
\textbf{\begin{tabular}[c]{@{}c@{}}Forward Energy~\cite{Rubinstein08}\end{tabular}} &
  0.956 &
  0.869 &
  0.911 &
  0.903 \\ \hline
\textbf{\begin{tabular}[c]{@{}c@{}}Frequency Tuned Saliency Map Method~\cite{OTHER_SC_METHODS_SALIENCY}\end{tabular}} &
  0.817 &
  0.425 &
  0.559 &
  0.578 \\ \hline
\textbf{\begin{tabular}[c]{@{}c@{}}Seam Merging Method~\cite{mishiba2012image}\end{tabular}} &
  0.862 &
  0.521 &
  0.649 &
  0.658 \\ \hline
\end{tabular}
\end{center}
\vspace{-0.6cm}
\caption{Seam carving technique generalizability of the seam removal pixelwise classifier, trained on forward energy seam carved xView images and tested on xView images seam carved using other techniques.}
\vspace{-0.3cm}
\label{tab:pixel_wise_stage1_gen_across_sc_techniques}
\end{table*}

We also observe that while seam removal detectors are not as generalizable as seam insertion detectors, they still perform well on different datasets as shown in Table \ref{tab:pixel_wise_stage1_gen_across_datasets}. Although the decrease in MCC-1 scores for seam removal detectors are larger when stress testing across datasets, we note that the scores themselves are still quite good, and are indicative of adequate performance for stage 2 classification.

\vspace{-0.3cm}
\subsubsection{Generalizability Across Seam Carving Methods}
\label{tab:generalizability_sc_techniques}
\vspace{-0.2cm}

We also tested the generalizability of our models across various seam carving methods. So far, all of our results have been reported on datasets that have been generated using forward energy seam carving. We report test evaluation metrics on our xView dataset in Table~\ref{tab:pixel_wise_stage1_gen_across_sc_techniques}, where we have generated test sets using backward energy, frequency tuned saliency maps, and seam merging variations of seam carving as developed in Section~\ref{sec:seam_carving_overview}. 
In this table, the model has been trained on only forward energy seam carving data, but we see good performance across different seam carving techniques, with the lowest performance on the dataset generated using saliency map based seam carving. This is most likely due to how different the seams from saliency map seam carving look compared to the other techniques.

         

\vspace{-0.2cm}
\begin{table}
\begin{center}
\begin{tabular}{|c|c|c|c|c|c|}
\hline
\textbf{Dataset} & \textbf{xView} & \textbf{xBD} & \textbf{xBD\_OR} & \textbf{Orbview-3} \\ \hline
\textbf{Accuracy}  & 99.29 & 99.08 & 99.73 & 98.46  \\ \hline
\end{tabular}
\end{center}
\vspace{-0.6cm}
\caption{Test set accuracy(\%) of stage 2 binary classifiers.}
\vspace{-0.6cm}
\label{tab:stage2_model_performance}
\end{table}

\vspace{-0.1cm}
\subsection{Stage 2 Classification}
\label{sec:stg2_pxlwise}
\vspace{-0.2cm}
In stage 2, we generate a dataset of stage 1 predictions on manipulated and pristine images to train a final binary classifier to check if the input image has been seam carved. 
We use a stage 1 model trained on the xBD dataset as described in \ref{ssec:data_pxl_wise_clf}. This stage 1 model is used to obtain predictions on a combination of xBD datasets. The first xBD dataset that we predict on is described in \ref{ssec:data_pxl_wise_clf}. We remove the 10\% most optimal seams, which are often distributed throughout the image, and reinsert seams to restore the original image dimensions. The second xBD dataset we predict on incorporates the original dataset's ground truth building masks and uses seam carving to remove a building and reinsert seams to restore the original image dimensions. We call this dataset "xBD\_OR". Thus, our stage 2 model is trained on a combination of stage 1 predictions on best seam removed and object removed images. Finally, we generate best seam removed test datasets of xView, xBD and Orbview-3 using individually trained pixelwise classifiers. For example, the Orbview-3 test dataset for stage 2 is generated using an Orbview-3 trained pixelwise classifier. Although the xView and Orbview-3 test datasets are generated using different pixelwise classifiers than the training dataset, our stage 2 model performs very well on both, achieving almost 99\% accuracy. This shows that the stage 2 final classification model is robust to the pixelwise classification model used to generate the input prediction masks.
 


\subsection{Experimental Setup}
\label{sec:exp_setup}
In this subsection, we cover the experimental setup for training our models. Seam removal and seam insertion pixelwise classifiers for the xView dataset are trained using the Adam optimizer with a learning rate of $0.0003$, momentum coefficients $\beta_{1}=0.9$ and $\beta_{2}=0.999$ and a numerical stability constant of $\epsilon=10^{-6}$. The loss function used is a pixelwise mean squared error. We find that using binary cross entropy as the loss function produces comparable results, but the mean squared error leads to smoother convergence. We use a batch size of $8$ where each batch is sampled randomly from the training set without replacement for each epoch. We train for $30$ epochs and select the model with the highest validation set accuracy, reducing the learning rate by a factor of $0.2$ if there is no validation accuracy improvement for $5$ epochs. For training on xBD and Orbview-3 datasets, we keep all hyperparameters the same except for a learning rate adjustment to $0.0001$.

Stage 2 binary classifiers are trained using identical hyperparameters as above, except that we use a cross entropy loss, a learning rate of $0.0001$, and train for $35$ epochs. These hyperparameters are the same for the stage 2 model used in Table \ref{tab:seam_carving_percentage_evalution} and Table~\ref{tab:stage2_model_performance}.

\subsection{Seam Carving Retargetting Ratios}
\label{sec:sc_ret}

In this section, we present results of two experiments where we vary the number of seams removed and inserted from the original xView dataset. 

In Table \ref{tab:seam_carving_percentage_evalution}, we provide MCC-1 and SLS scores for both seam removal and insertion detectors as well as stage 2 binary classification accuracies  on test datasets of varying seam carving retargetting ratios using a model that is trained on 10\% seam carved data. Test datasets are generated by seam carving 512x512 patches from xview test split by different percentages. We can see from this table that our seam carving detector is remarkably generalizable to other seam carving retargetting ratios despite being trained on only 10\% seam carved data. In terms of an overall image classification, our framework achieves over 99\% accuracy at detecting seam carved images on all retargetting ratios except 2\%, where it achieves the lowest accuracy at 98.56\%. In general, the SLS score of the seam removal detector decreases the further away we move from 10\% seam carving, while the MCC-1 only decreases as we increase the retargetting ratio. This showcases the usefulness of the SLS metric as a seam carving localization metric. In the case of 2\% seam carved data our seam removal detector achieves its best MCC-1 score of $0.918$ due to the large amount of non manipulated pixels in the ground truth and predicted masks. However, the SLS score of $2.05$ shows that our model is not as good at predicting seam locations as the MCC-1 score might lead us to expect. In the seam insertion case, we see good performance across the range of retargetting ratios tested. The lowest MCC-1 score for seam insertion is $0.920$ on 50\% seam carved data, which is higher than the best MCC-1 score for seam removal at $0.918$ on 2\% seam carved data. The seam insertion SLS scores show that our model is able to localize inserted seams within $1$ pixel precision across all retargetting ratios. 
Although we show results on a model that was trained using a 10\% seam carving ratio in this paper, the results were similar for models trained on other percentages of seam carving too.


\begin{table*}[h]
\begin{center}
\begin{tabular}{|c|c|c|c|c|c|}
\hline
\multirow{2}{*}{\textbf{Seam Carving Ratio}} & \multicolumn{2}{c|}{\textbf{Seam Removal Detector}} & \multicolumn{2}{c|}{\textbf{Seam Insertion Detector}} & \multirow{2}{*}{\textbf{Stage 2 Accuracy(\%)}} \\ \cline{2-5}
                   & \textbf{MCC-1} & \textbf{SLS}   & \textbf{MCC-1} & \textbf{SLS}      & \\ \hline
2\%                &    0.918   &    2.05   &   0.943    & 0.26 &    98.56   \\ \hline
4\%                &    0.913   &   1.48    &   0.951    & 0.14 &   99.02      \\ \hline
6\%                &    0.910   &    1.30   &   0.953    & 0.12  &  99.06      \\ \hline
8\%                &   0.908    &   1.18    &   0.954    &  0.12  & 99.14      \\ \hline
10\%                &    0.903   &    1.56   &   0.959    & 0.11 &   99.29      \\ \hline
20\%                &   0.886    &   1.05    &   0.953    & 0.14 &   99.23      \\ \hline
30\%                &   0.831    &   1.51    &    0.945   & 0.25  &  99.34      \\ \hline
40\%                &   0.697    &    4.48   &   0.934    & 0.42 &    99.51  \\ \hline
50\%                &     0.471  &    21.79   &    0.920   & 0.61 &   99.51  \\ \hline
\end{tabular}
\end{center}
\caption{Performance of Stage 1 and Stage 2 models, trained on a 10\% seam carved images from the xView dataset and evaluated on xView test sets with varying numbers of removed and inserted seams.}

\label{tab:seam_carving_percentage_evalution}
\end{table*}

In Table \ref{tab:seam_carving_percentages}, we show results of training on datasets of different seam carving retargetting ratios. We include the results on the 10\% seam carved test set, and provide evaluation metrics on 20\%, 30\%, 40\%, and 50\% seam carved test datasets. This table shows that our method is applicable to other seam carving retargetting ratios and achieves similar results. The SLS score is around $1$ for all values tested, and the F1 Score-1 and MCC-1 scores are high. Notably, the 20\% seam carving dataset has the highest performance while the 50\% seam carved dataset has the worst. 

\begin{table}[h]
\begin{center}
\begin{tabular}{|c|c|c|c|}
\hline
\textbf{Seam Carving Ratio} & \textbf{F1Score-1} & \textbf{MCC-1} & \textbf{SLS} \\ \hline
10\% & 0.911 & 0.903 & 1.56 \\ \hline
20\% & 0.986 & 0.893 & 0.838 \\ \hline
30\% & 0.907 & 0.891 & 0.776 \\ \hline
40\% & 0.901 & 0.883 & 0.892 \\ \hline
50\% & 0.865 & 0.847 & 1.23 \\ \hline
\end{tabular}
\end{center}
\caption{Performance of Stage 1 models trained on xview datasets generated by inserting and removing different percentage of seams.}
\label{tab:seam_carving_percentages}
\end{table}

\subsection{Seam Carving detection on post processed images}
\label{sec:sc_post_processing}
We explored the performance of our models on the dataset of images that are JPEG compressed after seam carving. Retraining the models on post processed datasets resulted in the similar image level detection, while the models that are not trained on post processed images has shown drop in performance, as shown in Table~\ref{tab:jpeg_performance}. Same trend has been observed when we replaced JPEG compression with rotation as post processing step, as shown in Table~\ref{tab:rotation_performance}.

\begin{table}[h]
\begin{center}
\begin{tabular}{|c|c|c|}
\hline
\textbf{\begin{tabular}[c]{@{}c@{}}JPEG Compression\\ Quality Factor\end{tabular}} & \textbf{\begin{tabular}[c]{@{}c@{}}Model trained\\ without JPEG\end{tabular}} & \textbf{\begin{tabular}[c]{@{}c@{}}Model trained\\ with JPEG\end{tabular}} \\ \hline
60 & 51.96 & 84.86 \\ \hline
70 & 53.88 & 90.11 \\ \hline
80 & 59.06 & 95.06 \\ \hline
90 & 74.19 & 97.81 \\ \hline
No Comp & 99.26 & 99.18 \\ \hline
\end{tabular}
\end{center}
\vspace{-0.5cm}
\caption{Stage-2 Test accuracy(\%): Improvement in detection accuracy when the model is trained on post processed images.}
\label{tab:jpeg_performance}
\vspace{-0.4cm}
\end{table}

\begin{table}[h]
\begin{center}
\begin{tabular}{|c|c|c|}
\hline
\textbf{\begin{tabular}[c]{@{}c@{}}Rotation\\ (Degrees)\end{tabular}} & \textbf{\begin{tabular}[c]{@{}c@{}}Model trained\\ without rotated\\ images\end{tabular}} & \textbf{\begin{tabular}[c]{@{}c@{}}Model trained\\ with rotated\\ images\end{tabular}} \\ \hline
45 & 52.11 & 91.26 \\ \hline
60 & 56.63 & 96.78 \\ \hline
75 & 63.14 & 97.66 \\ \hline
90 & 99.26 & 99.11 \\ \hline
\begin{tabular}[c]{@{}c@{}}No \\ Rotation\end{tabular} & 99.26 & 99.11 \\ \hline
\end{tabular}
\end{center}
\vspace{-0.5cm}
\caption{Stage-2 Test accuracy(\%): Improvement in detection accuracy when the model is trained on post processed images.}
\label{tab:rotation_performance}
\vspace{-0.4cm}
\end{table}

\vspace{-0.2cm}
\subsection{Visual Examples}
\label{sec:vis_ex}

\begin{figure*}
\setlength\tabcolsep{2pt}
\centering
\newcommand*{\factor}{0.4}
\begin{tabular}{ccccc}
 
 \includegraphics[height=\factor\columnwidth,keepaspectratio]{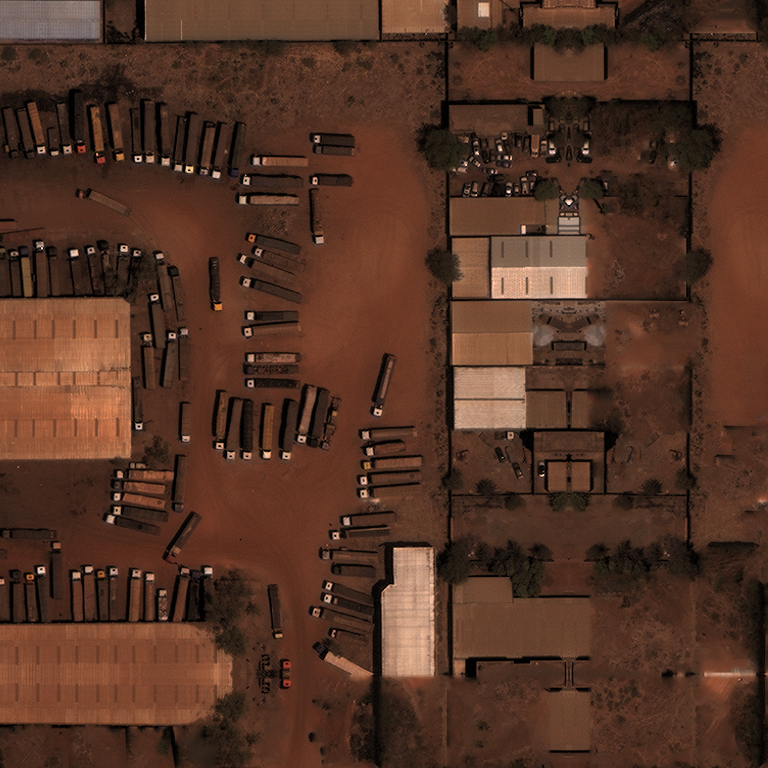} &
 \includegraphics[height=\factor\columnwidth,keepaspectratio]{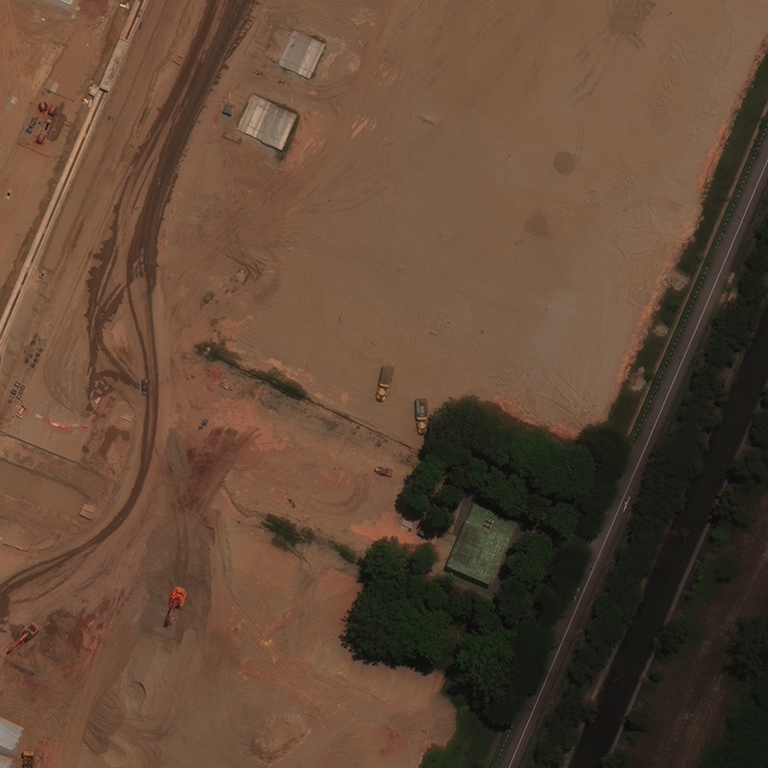} &
 \includegraphics[height=\factor\columnwidth,keepaspectratio]{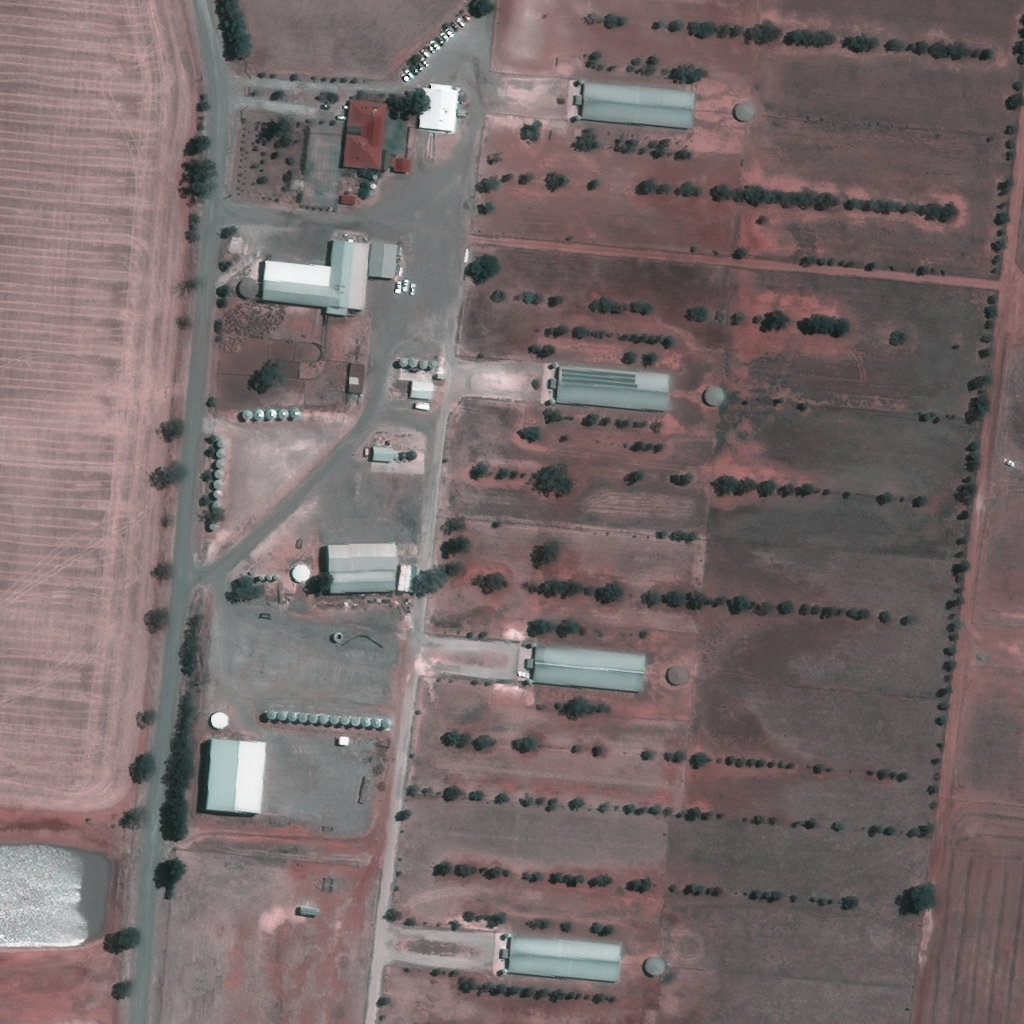} & 
 \includegraphics[height=\factor\columnwidth,keepaspectratio]{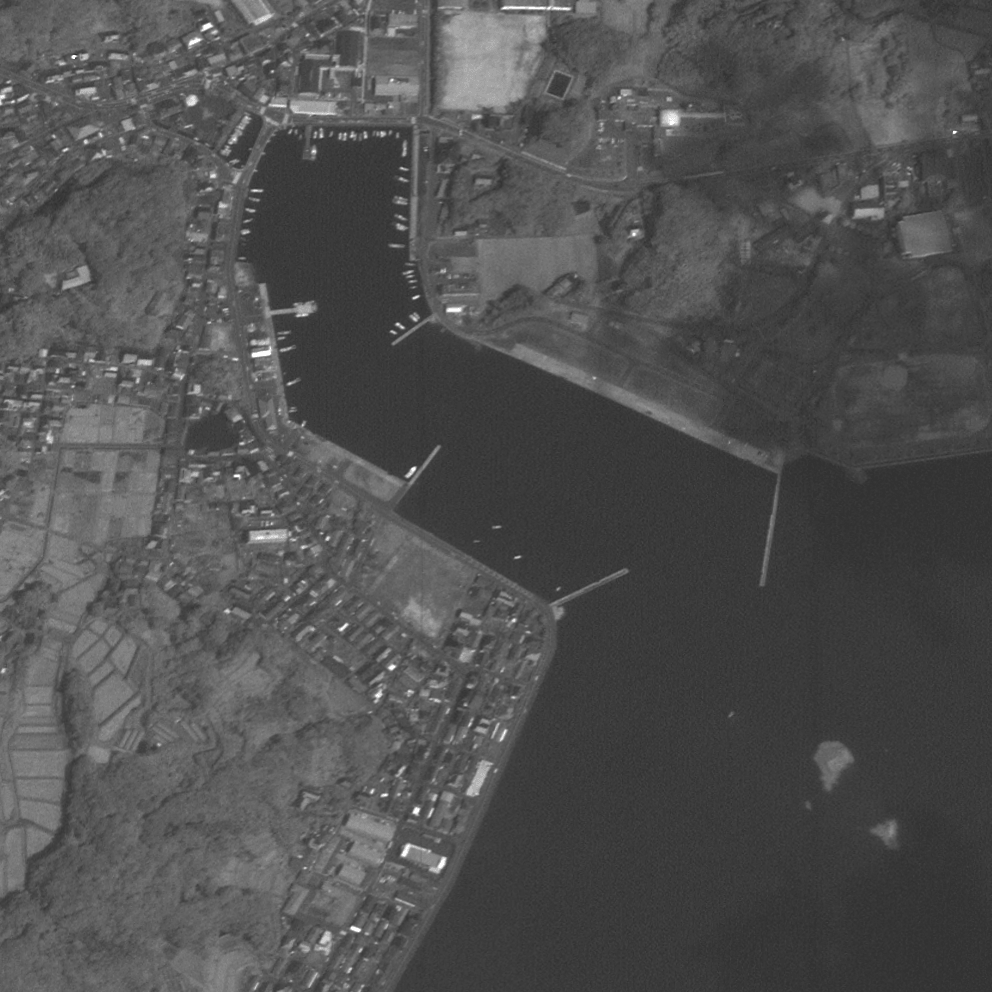} & 
 \includegraphics[height=\factor\columnwidth,keepaspectratio]{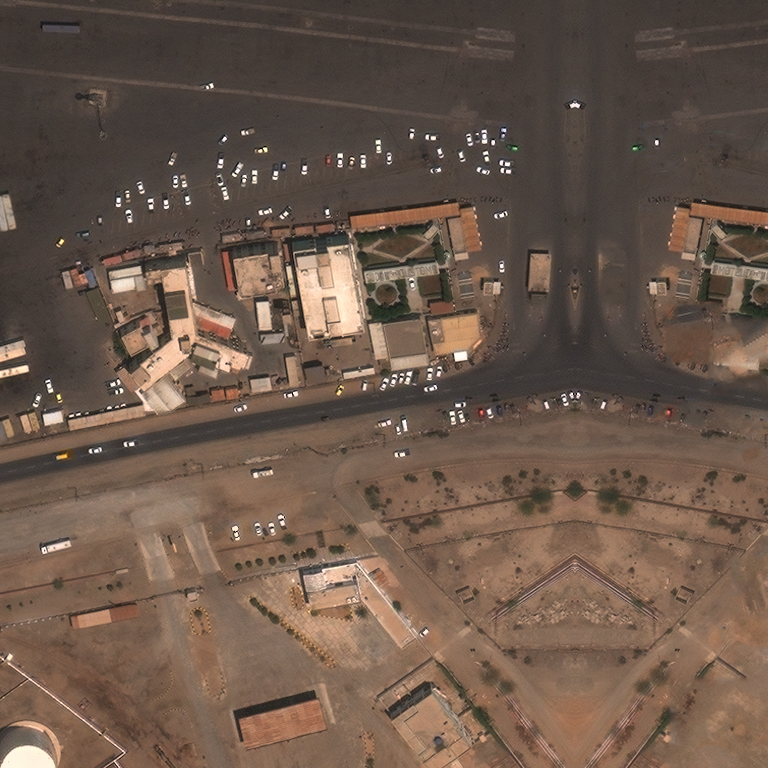} \\
 
  \includegraphics[height=\factor\columnwidth,keepaspectratio]{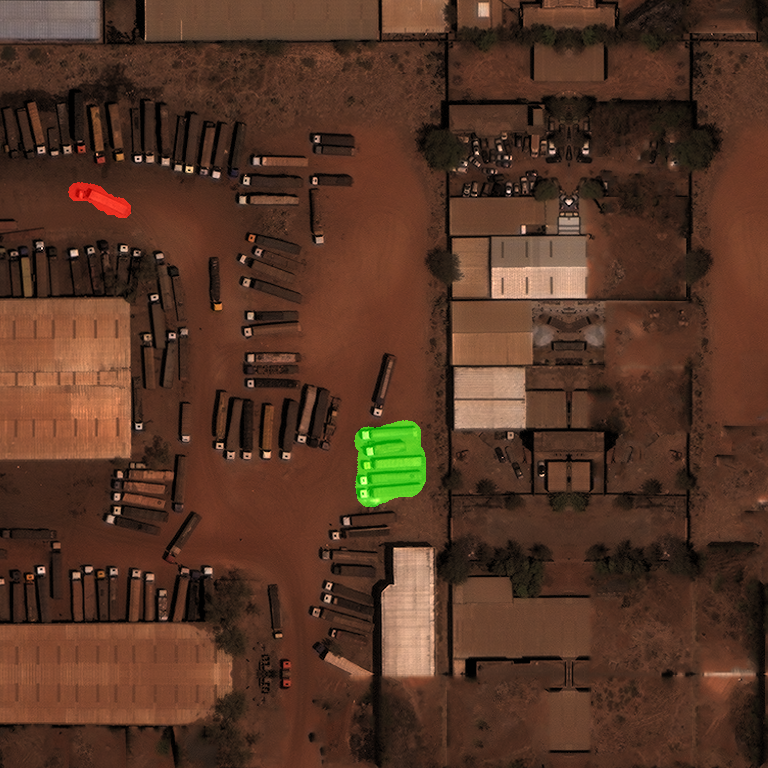}&
 \includegraphics[height=\factor\columnwidth,keepaspectratio]{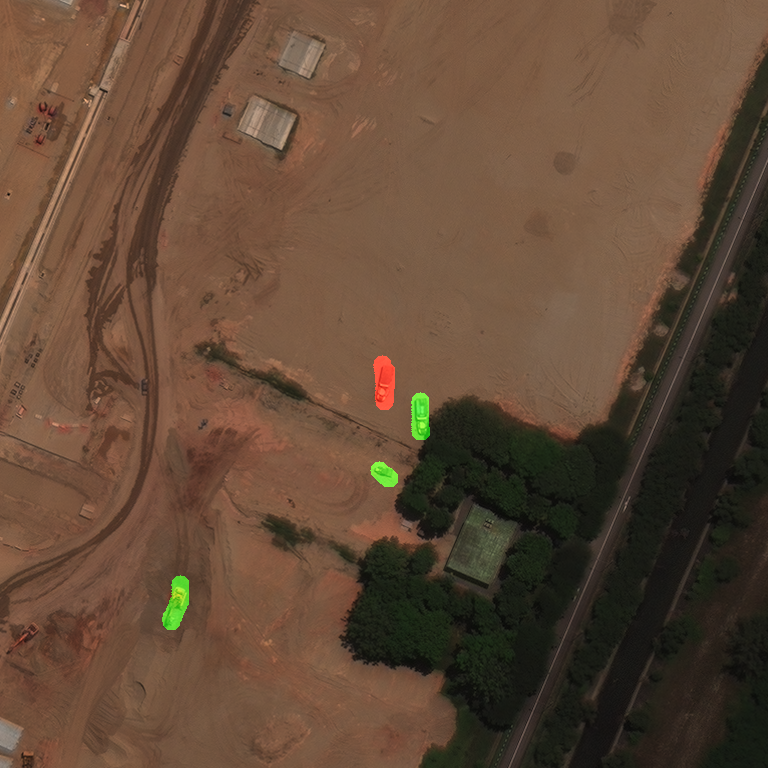} &
 \includegraphics[height=\factor\columnwidth,keepaspectratio]{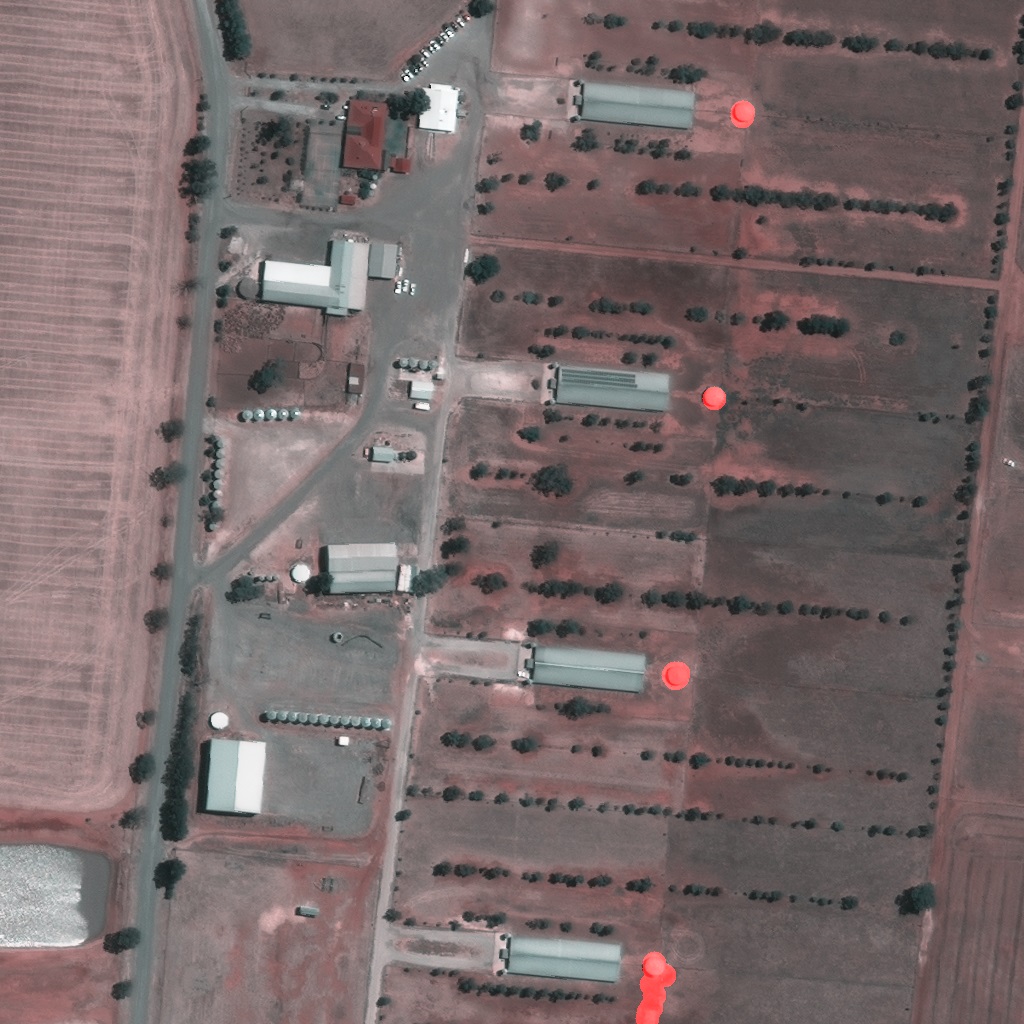} &
 \includegraphics[height=\factor\columnwidth,keepaspectratio]{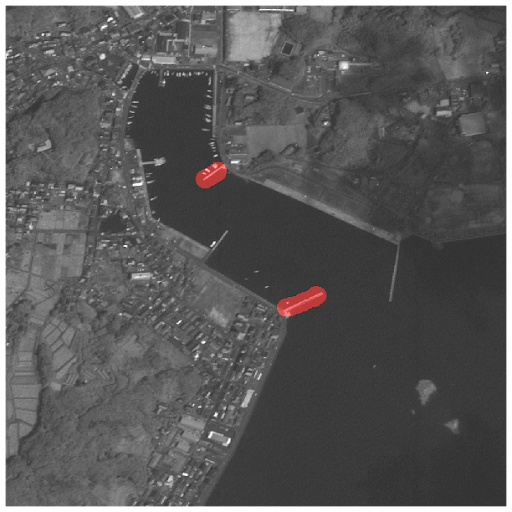} &
 \includegraphics[height=\factor\columnwidth,keepaspectratio]{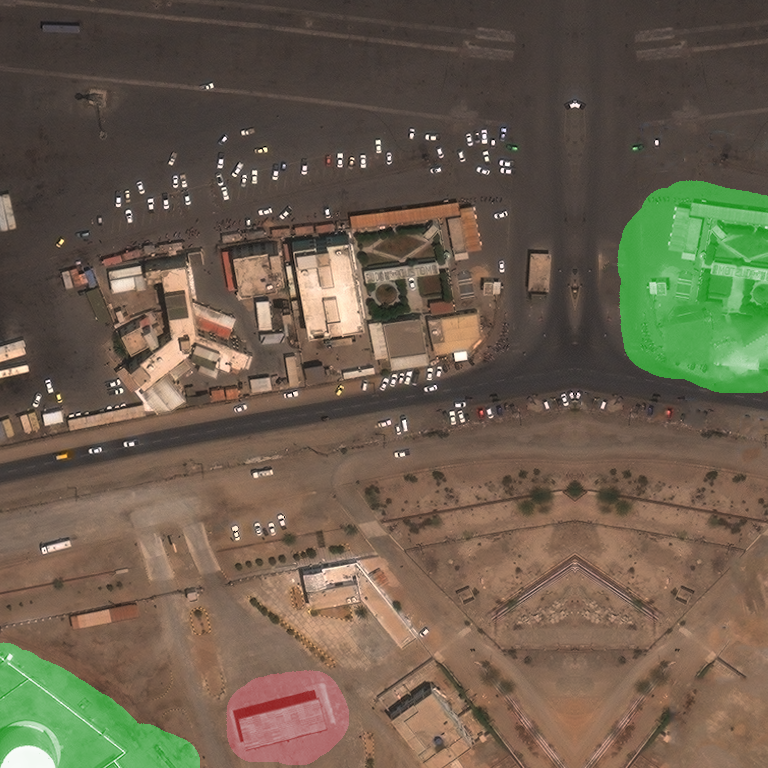} \\
 
 \includegraphics[height=\factor\columnwidth,keepaspectratio]{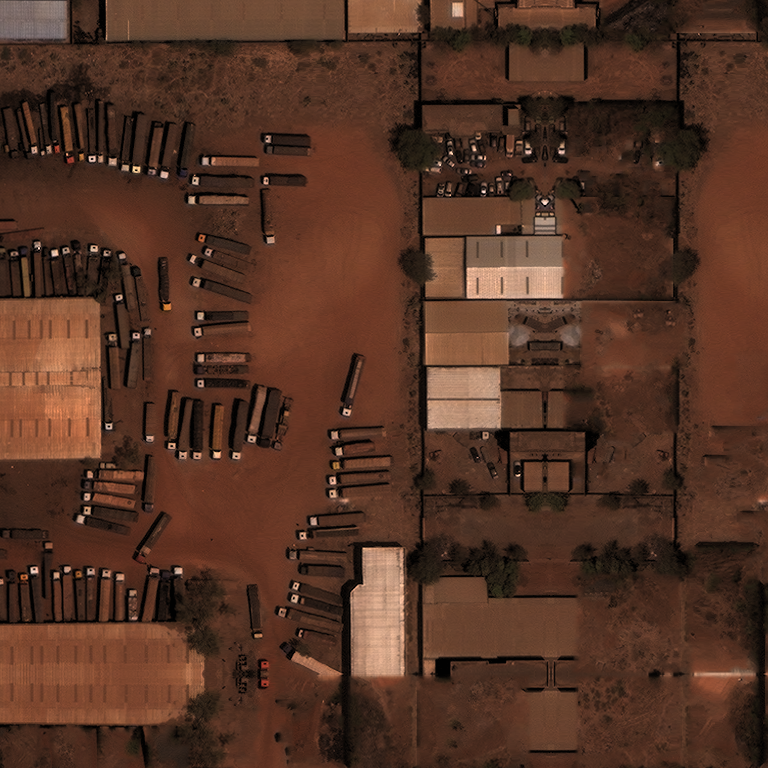}&
 \includegraphics[height=\factor\columnwidth,keepaspectratio]{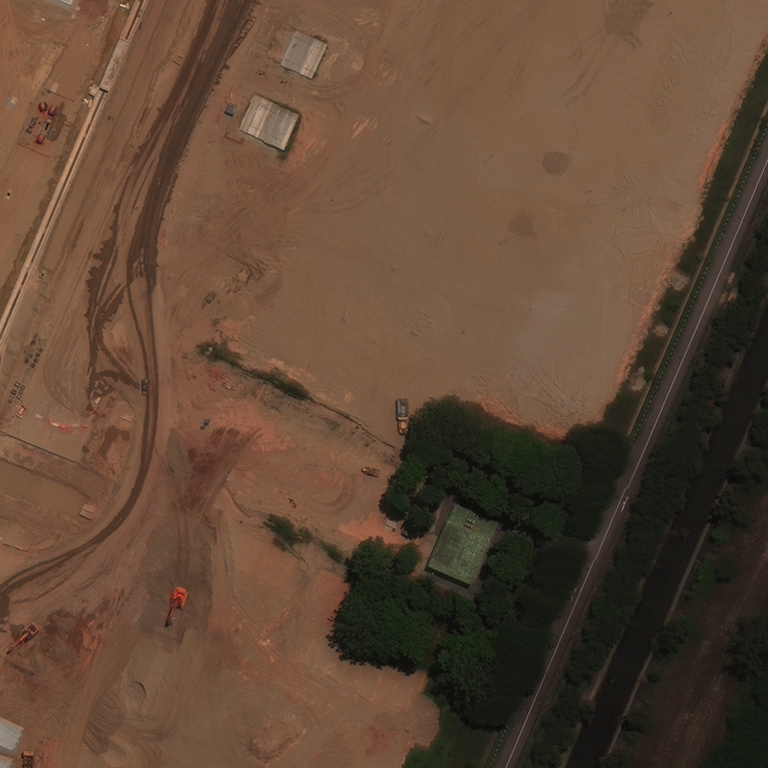}&
 \includegraphics[height=\factor\columnwidth,keepaspectratio]{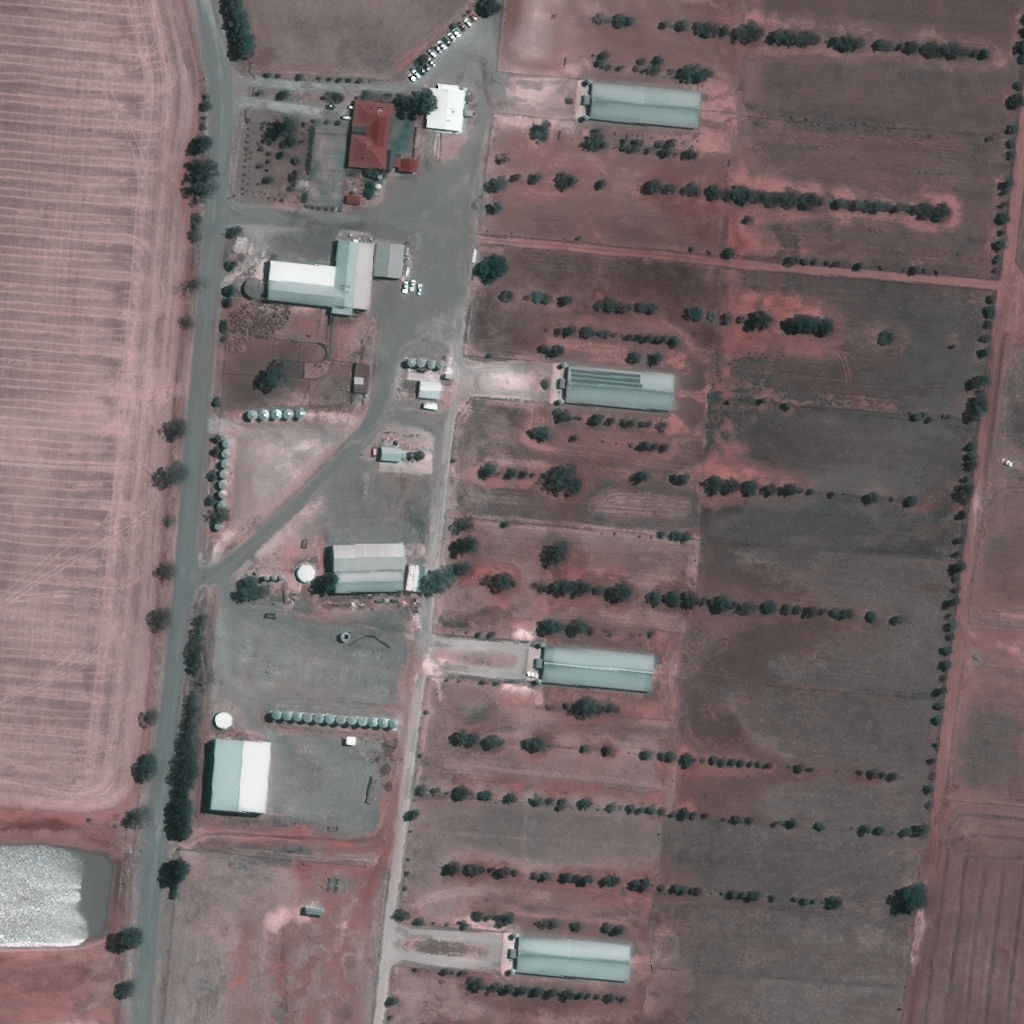}&
 \includegraphics[height=\factor\columnwidth,keepaspectratio]{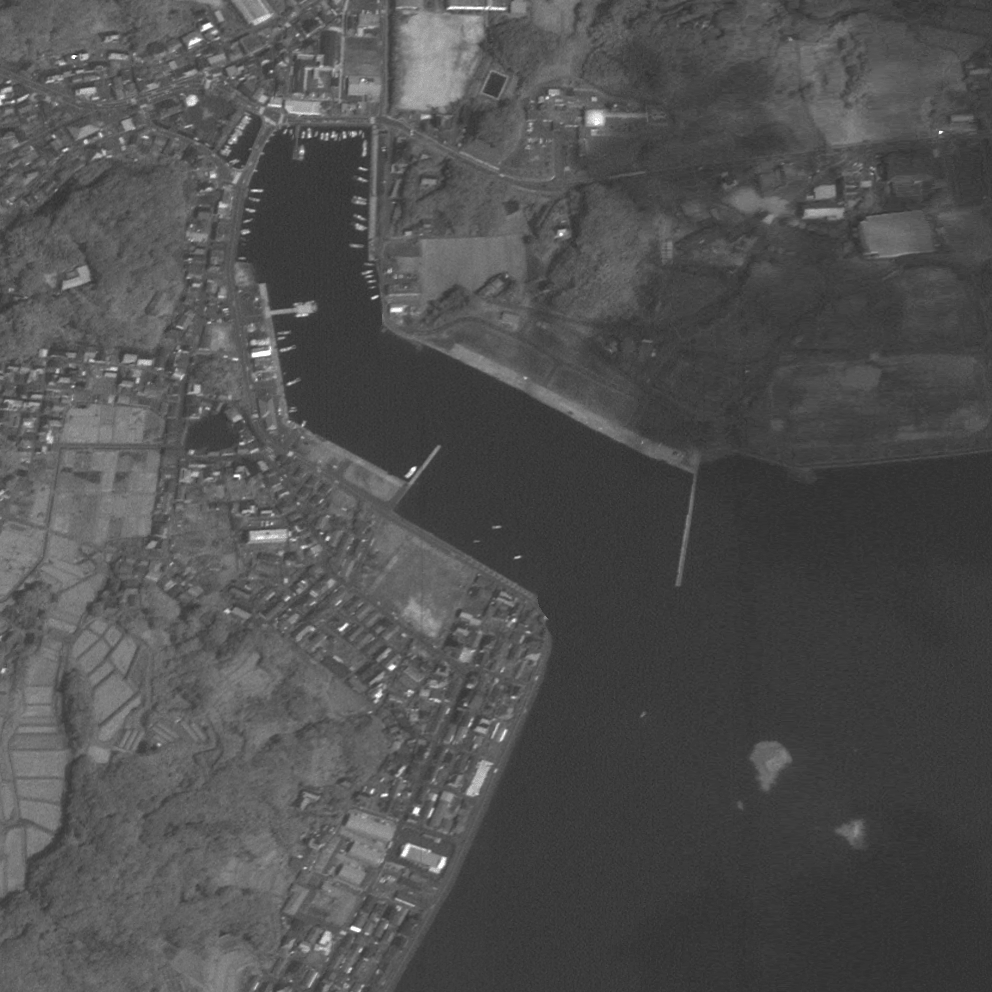}&
 \includegraphics[height=\factor\columnwidth,keepaspectratio]{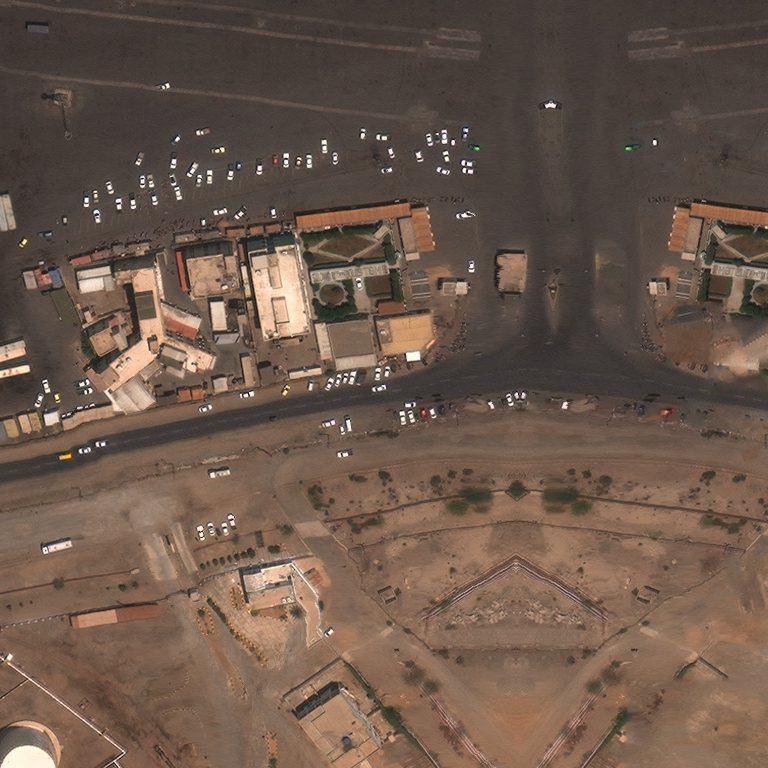}\\
 
 \includegraphics[height=\factor\columnwidth,keepaspectratio]{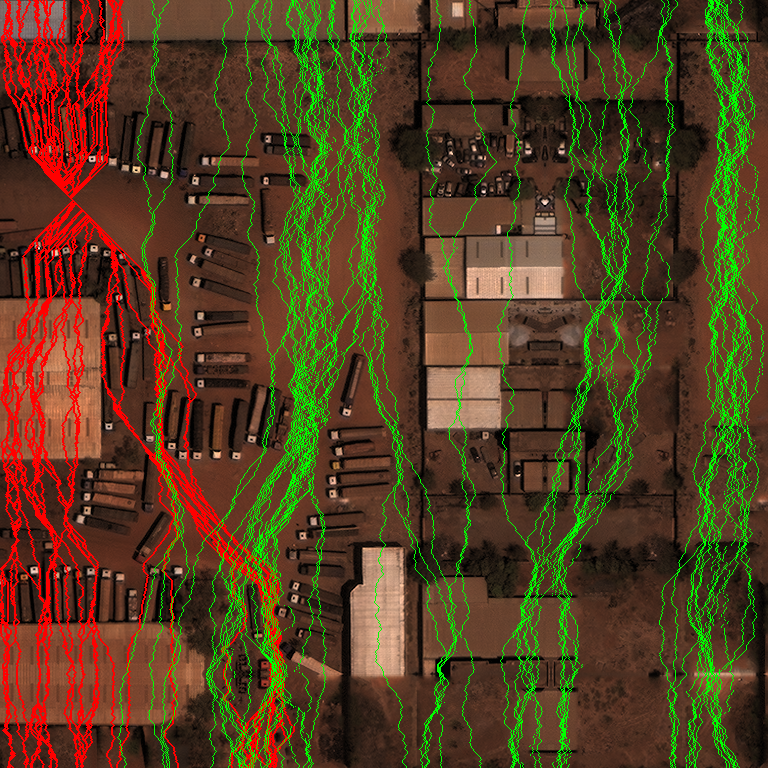}&
 \includegraphics[height=\factor\columnwidth,keepaspectratio]{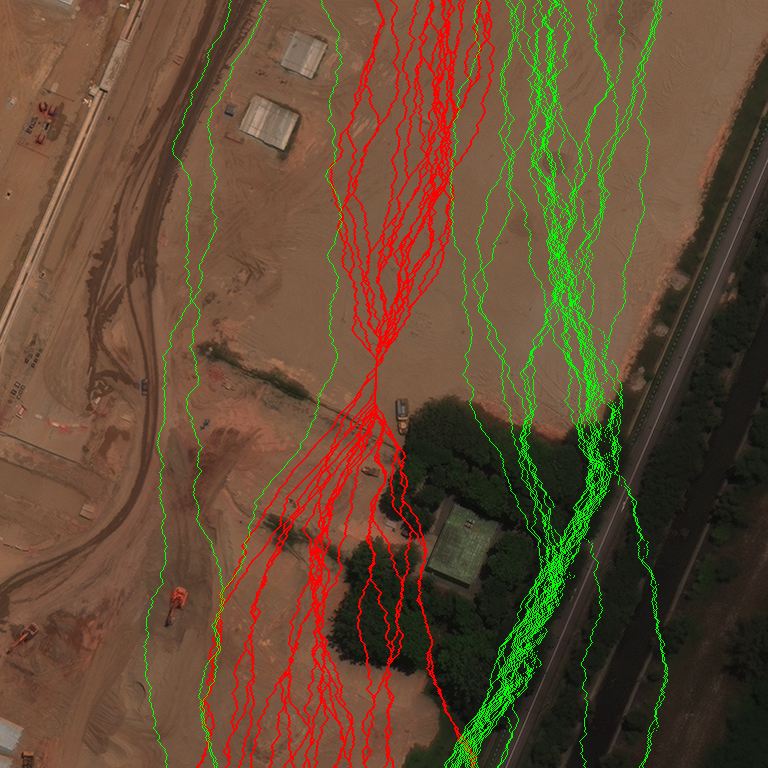}&
 \includegraphics[height=\factor\columnwidth,keepaspectratio]{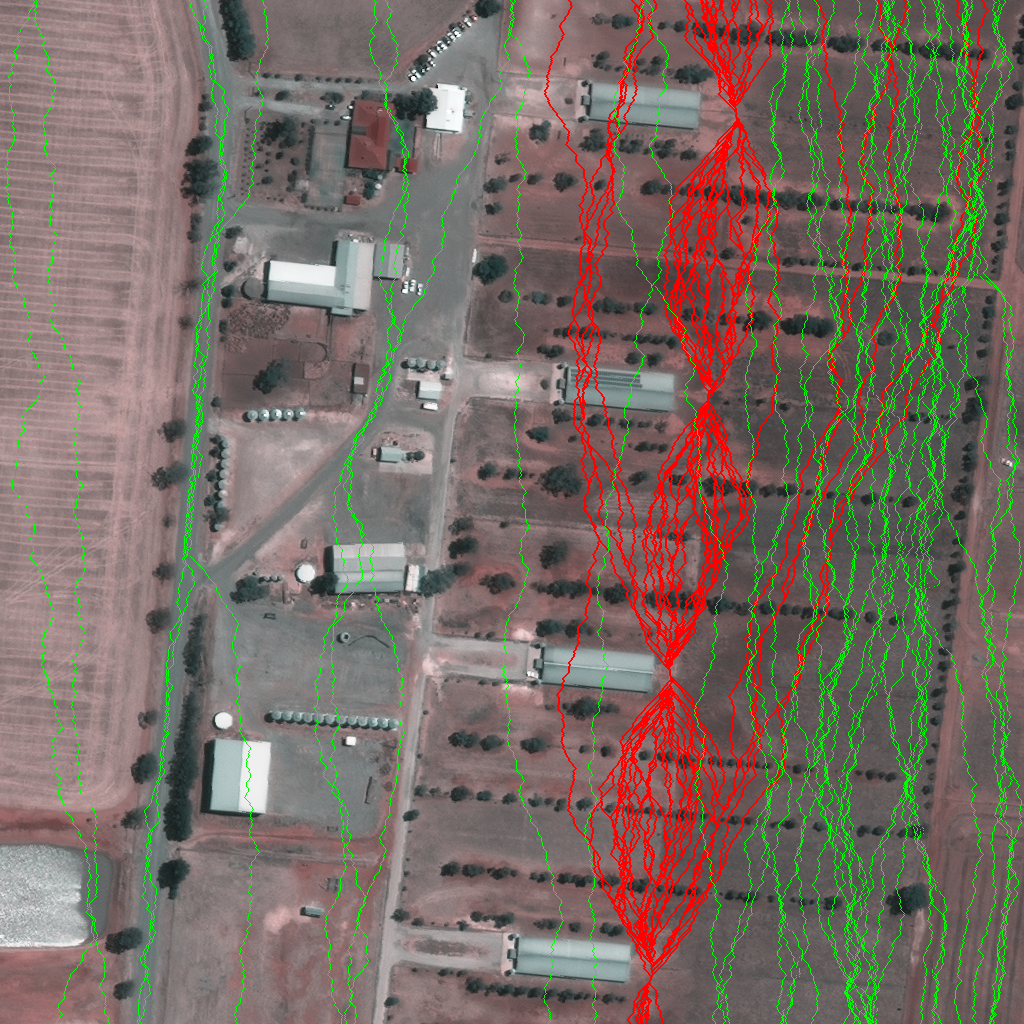}&
 \includegraphics[height=\factor\columnwidth,keepaspectratio]{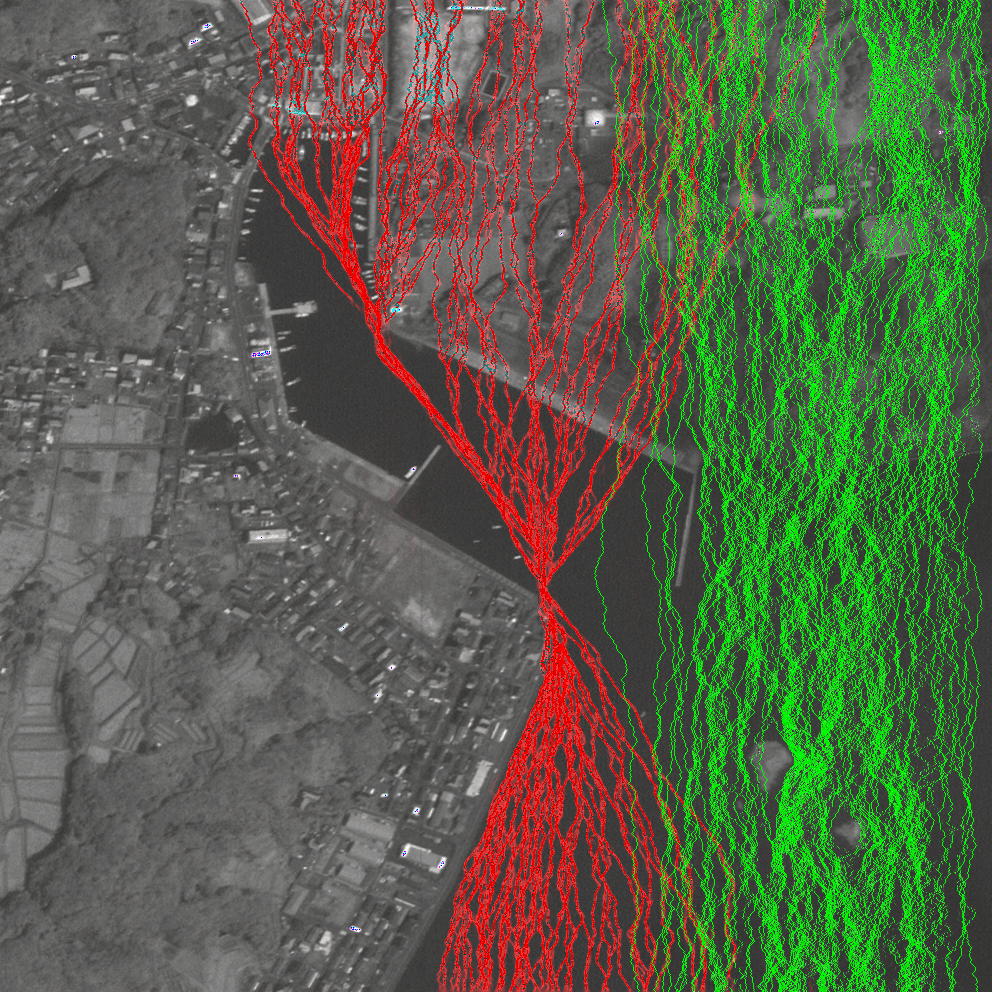}&
 \includegraphics[height=\factor\columnwidth,keepaspectratio]{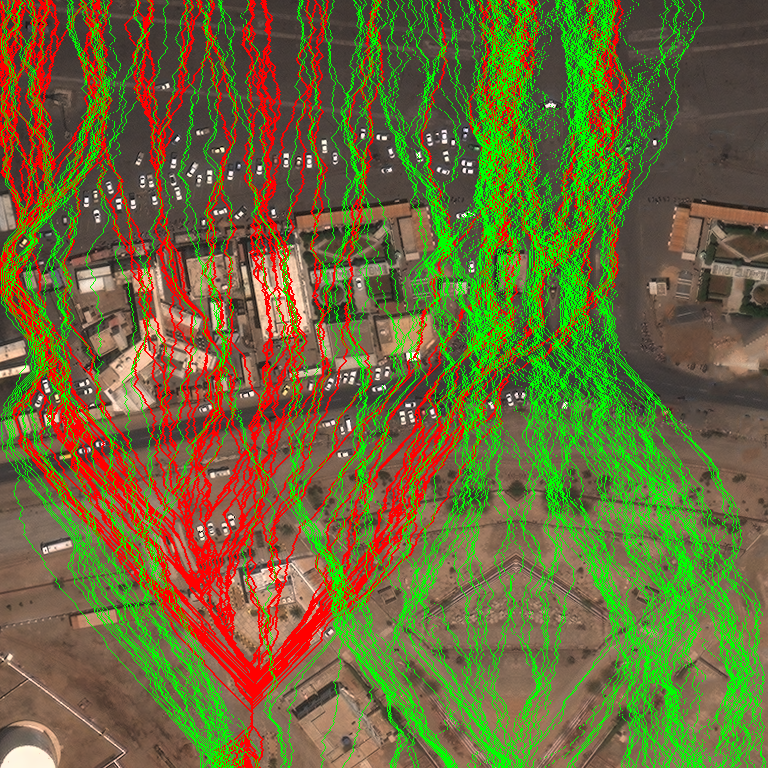} \\
 
 \includegraphics[height=\factor\columnwidth,keepaspectratio]{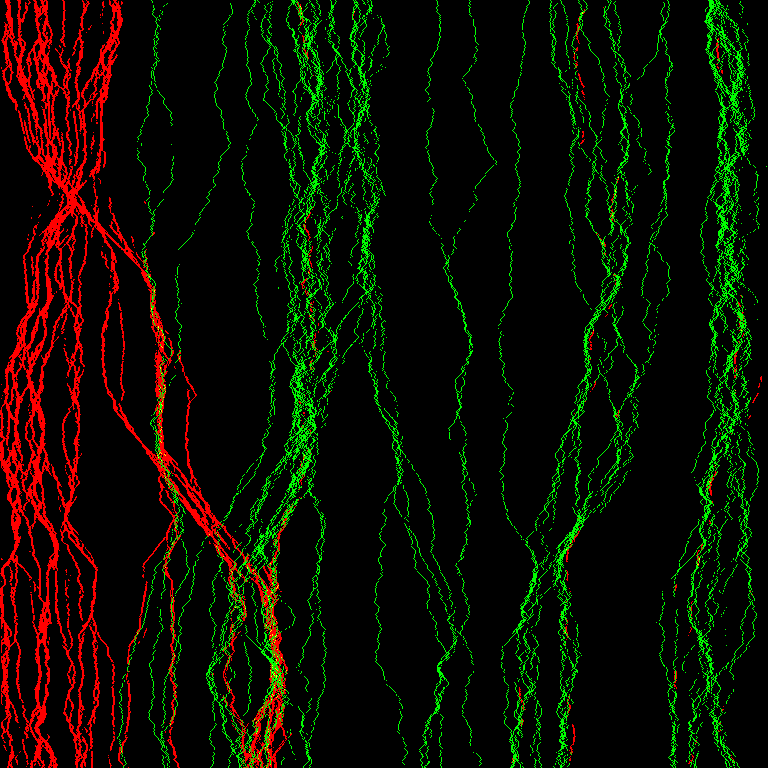}&
 \includegraphics[height=\factor\columnwidth,keepaspectratio]{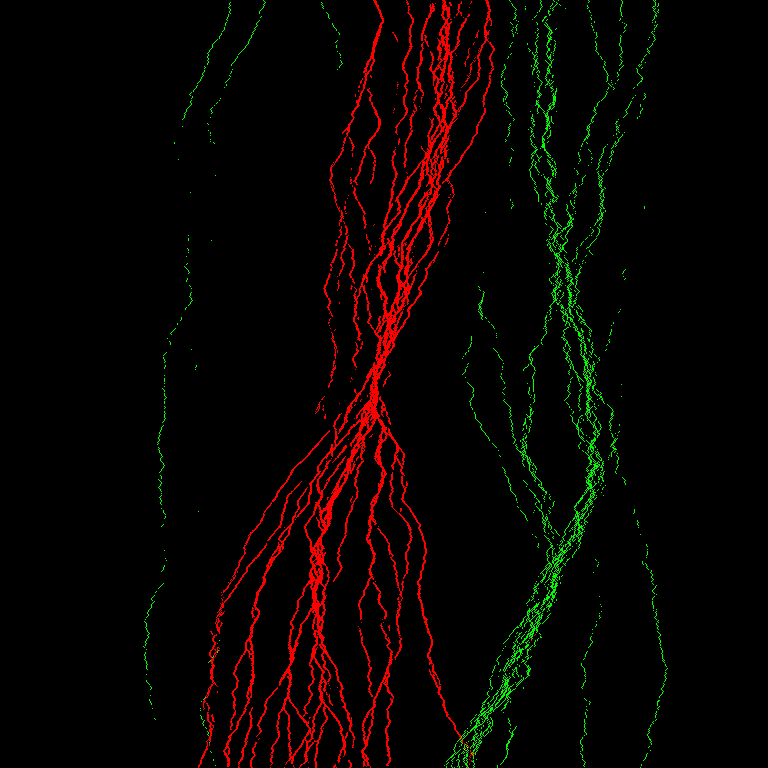}&
 \includegraphics[height=\factor\columnwidth,keepaspectratio]{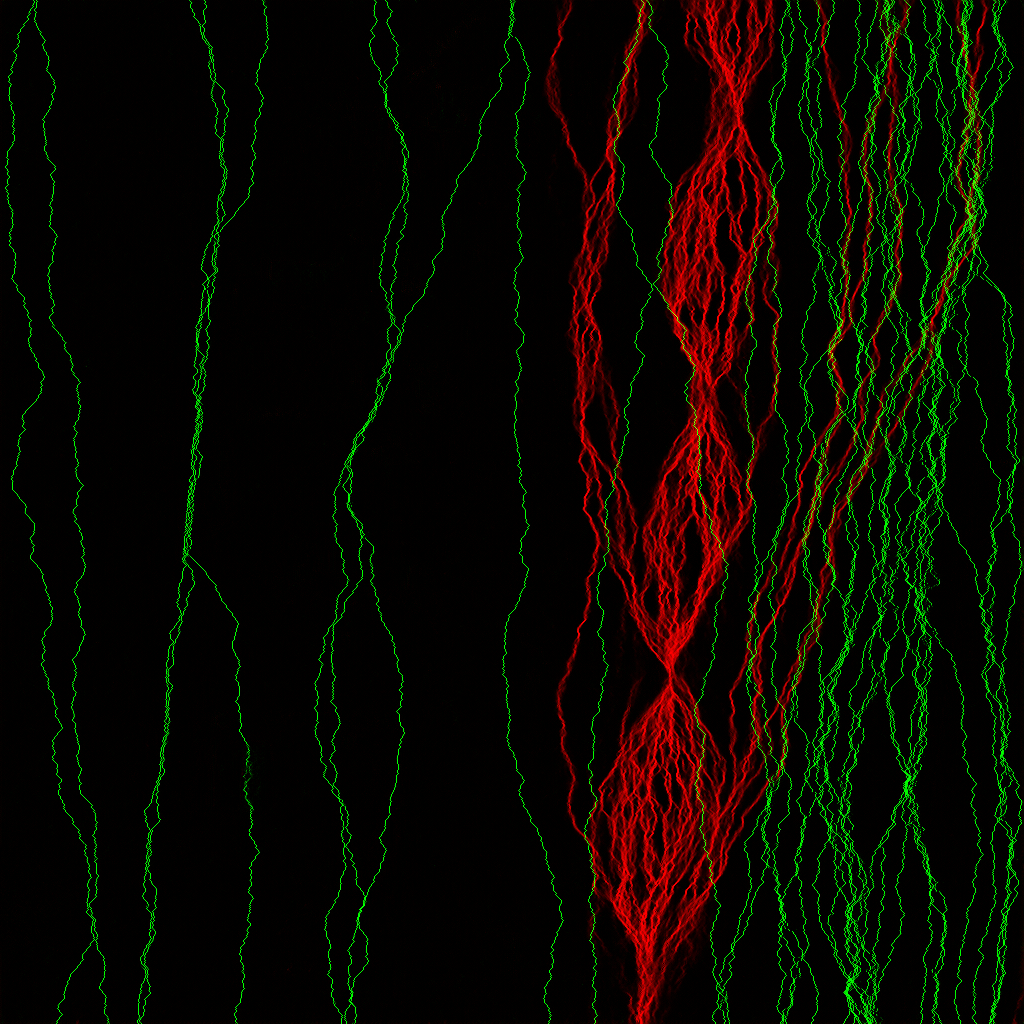}&
 \includegraphics[height=\factor\columnwidth,keepaspectratio]{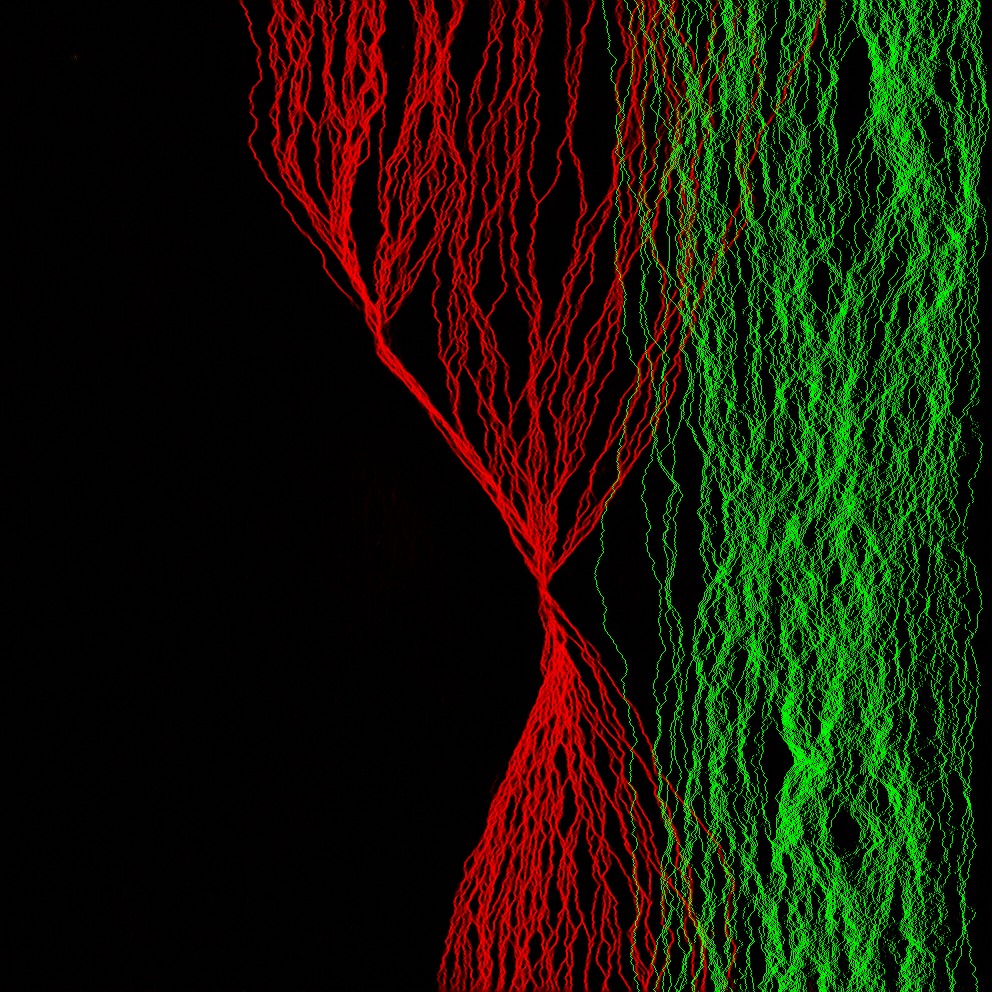}&
 \includegraphics[height=\factor\columnwidth,keepaspectratio]{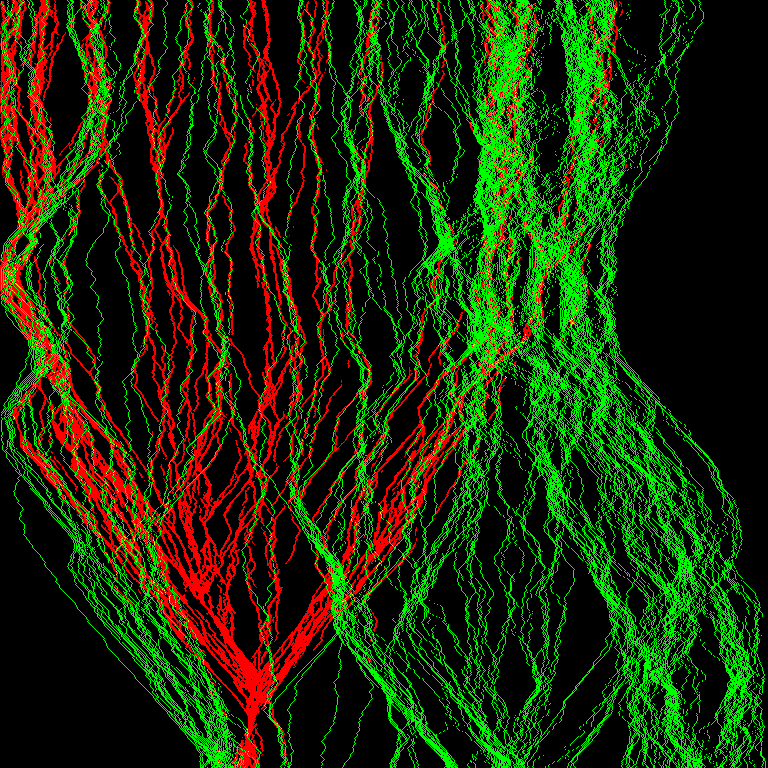}\\
 
 (a) &
 (b) &
 (c) &
 (d) &
 (e)
 
\end{tabular}

\caption{Object Removal Examples. From top to bottom: Pristine satellite image; Pristine image overlaid on removal (red) and protective (green) masks; Seam carved image with objects removed while retaining the original image size; Ground truth seam mask, with removed (red) and inserted (green) seams, overlaid on seam carved image; Predicted seam mask generated by stage1 seam removal detector (red) and seam insertion detector (green).}

\label{fig:sc_obj_rem_supp}
\end{figure*}

\begin{figure*}
\setlength\tabcolsep{2pt}
\centering
\newcommand*{\factor}{0.4}
\begin{tabular}{ccccc}
 
 \includegraphics[height=\factor\columnwidth,keepaspectratio]{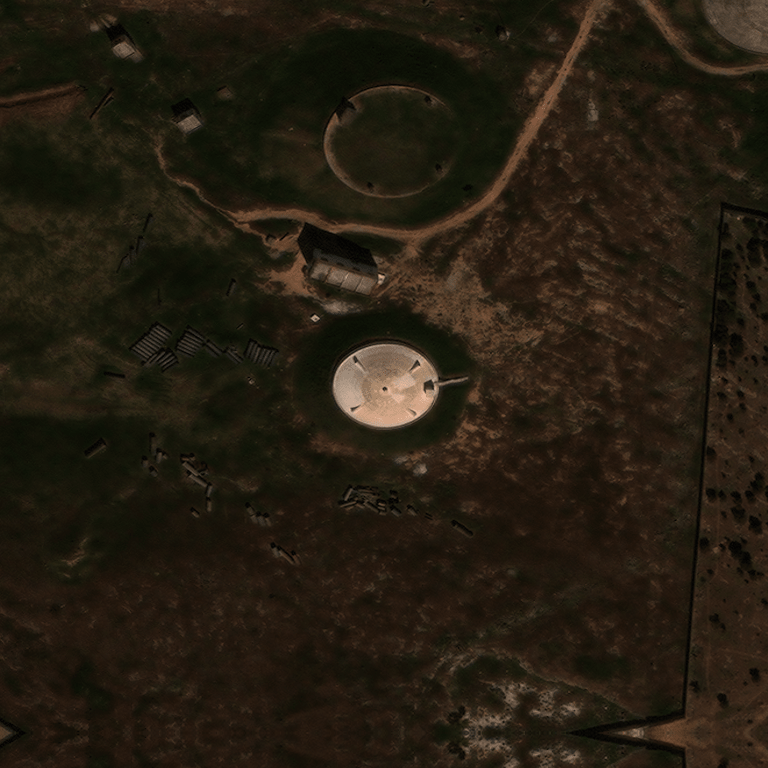}&
 \includegraphics[height=\factor\columnwidth,keepaspectratio]{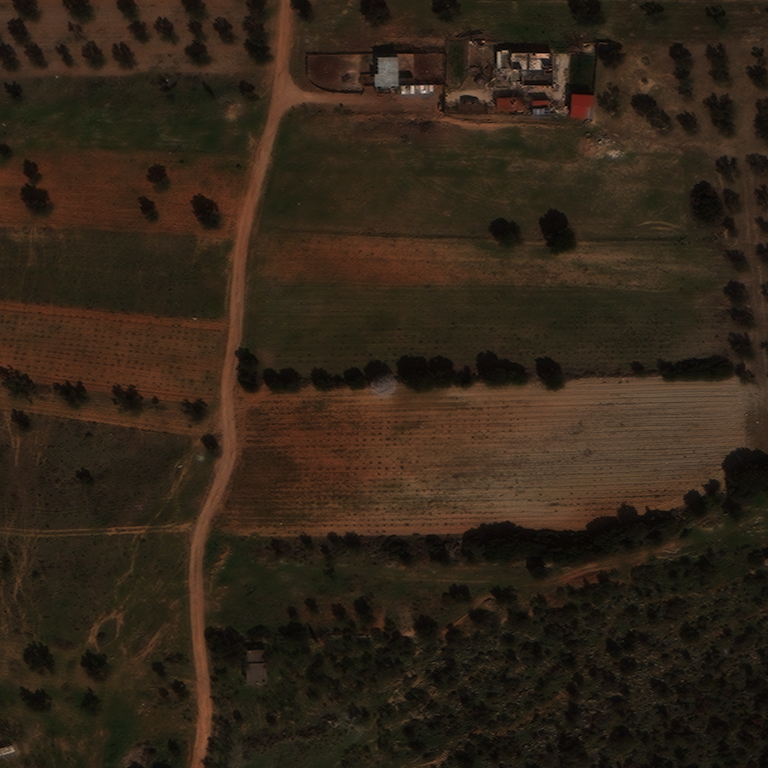}&
 \includegraphics[height=\factor\columnwidth,keepaspectratio]{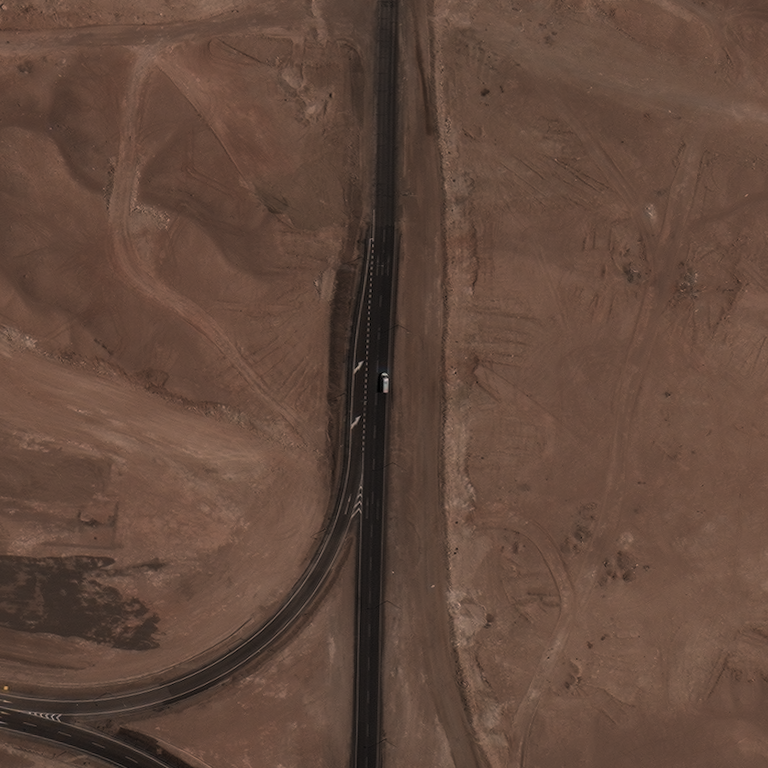}& 
 \includegraphics[height=\factor\columnwidth,keepaspectratio]{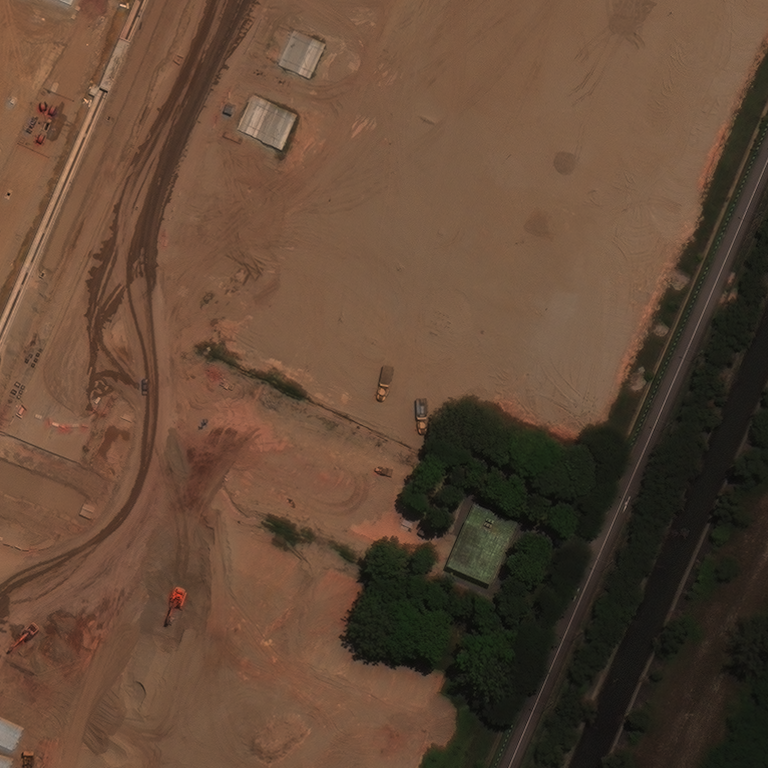}& 
 \includegraphics[height=\factor\columnwidth,keepaspectratio]{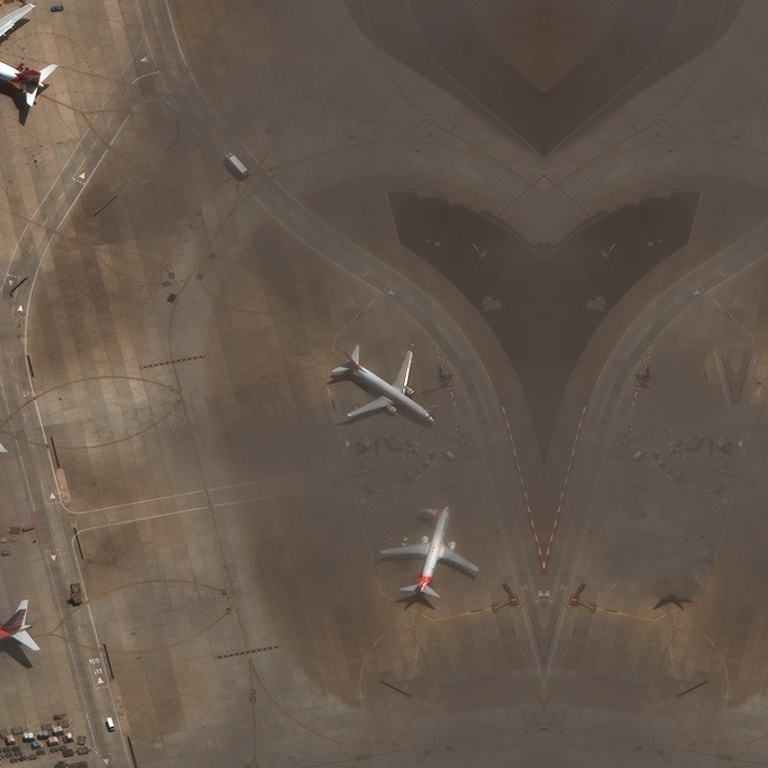} \\
 
 \includegraphics[height=\factor\columnwidth,keepaspectratio]{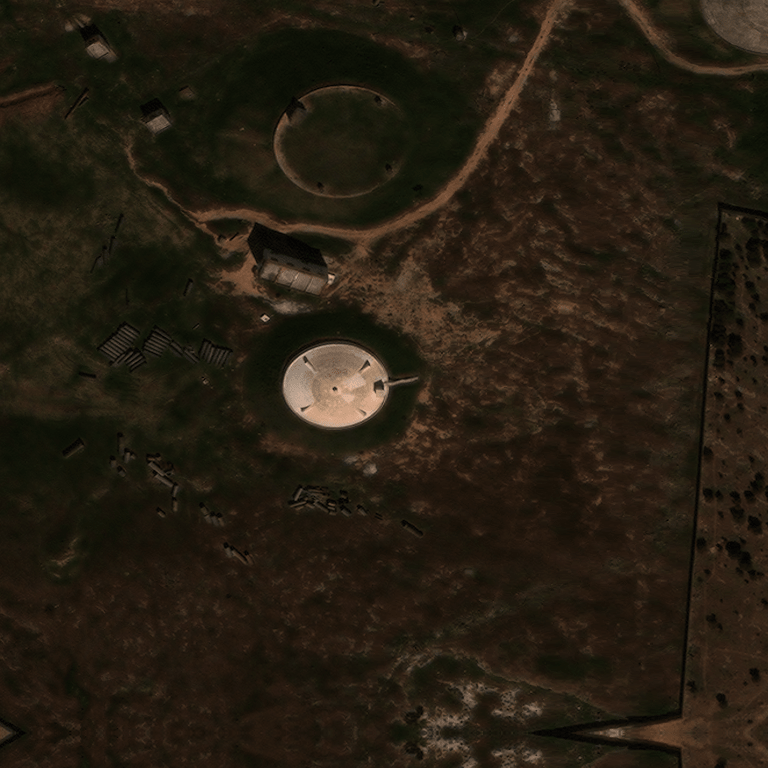}&
 \includegraphics[height=\factor\columnwidth,keepaspectratio]{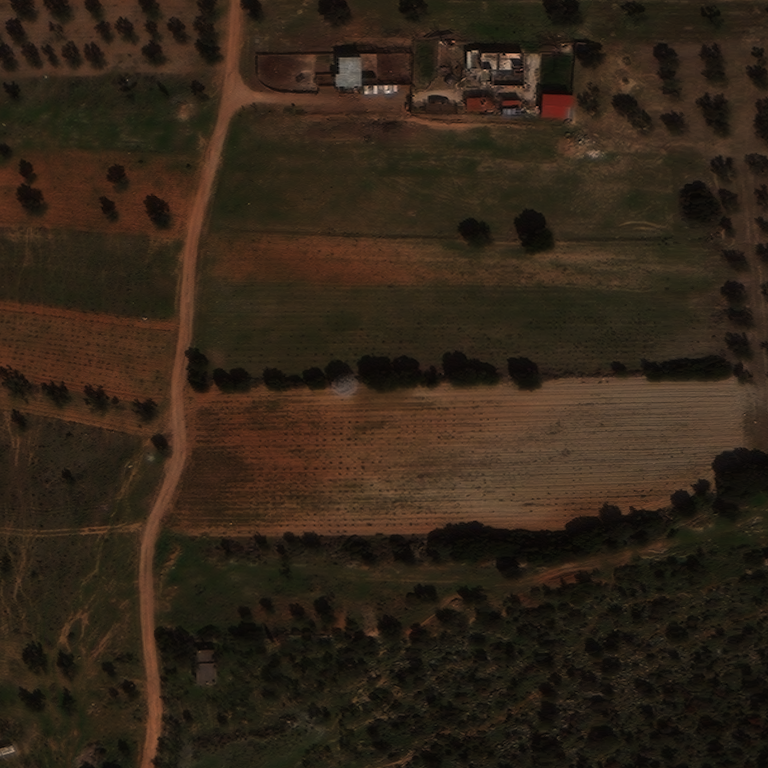}&
 \includegraphics[height=\factor\columnwidth,keepaspectratio]{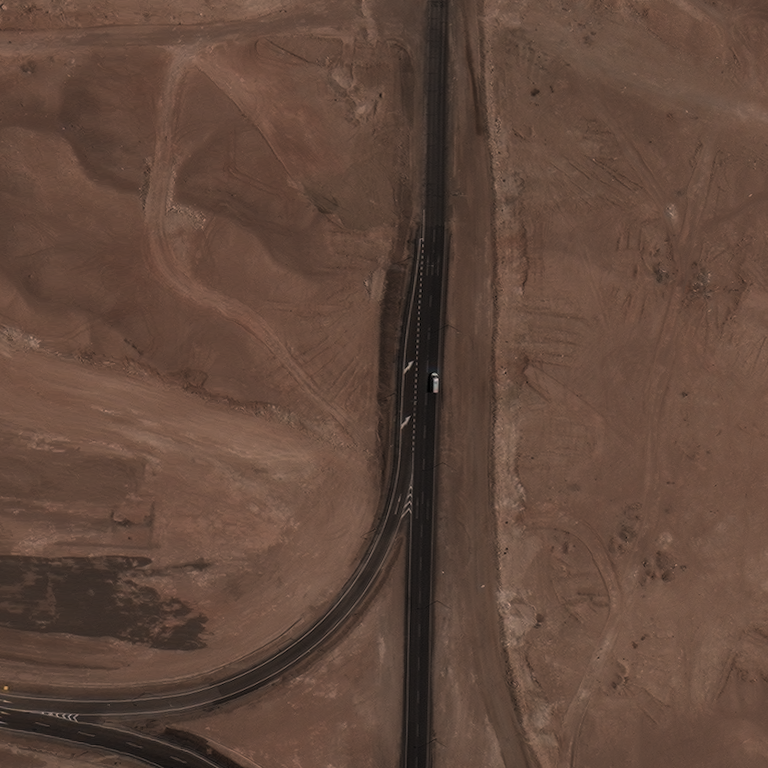}&
 \includegraphics[height=\factor\columnwidth,keepaspectratio]{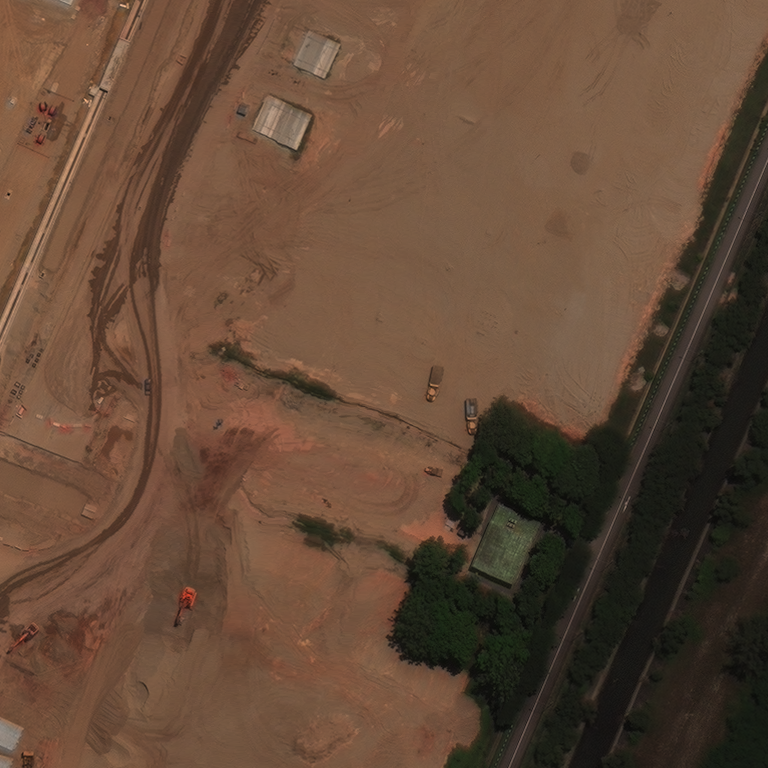}&
 \includegraphics[height=\factor\columnwidth,keepaspectratio]{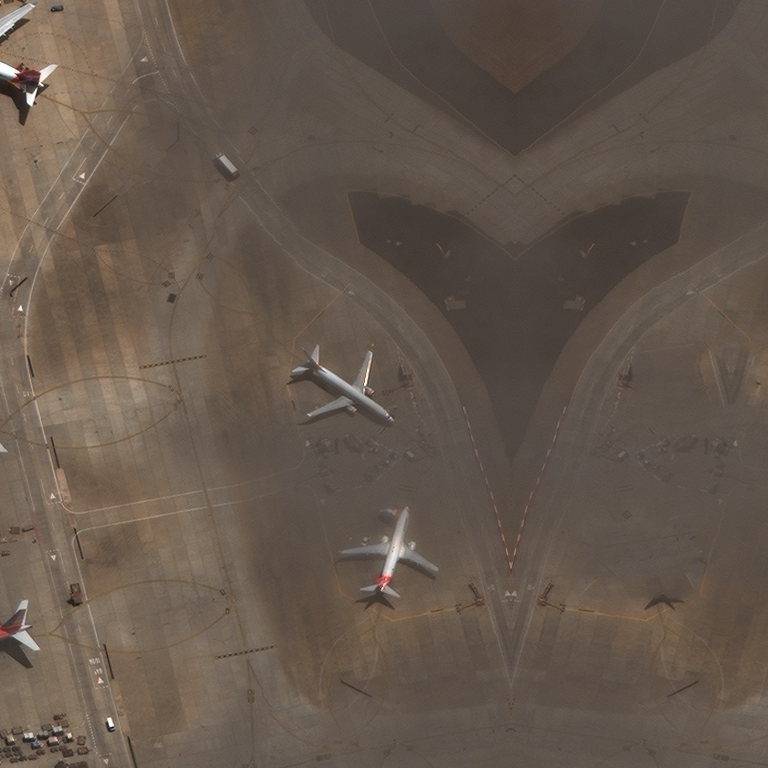}\\
 
 \includegraphics[height=\factor\columnwidth,keepaspectratio]{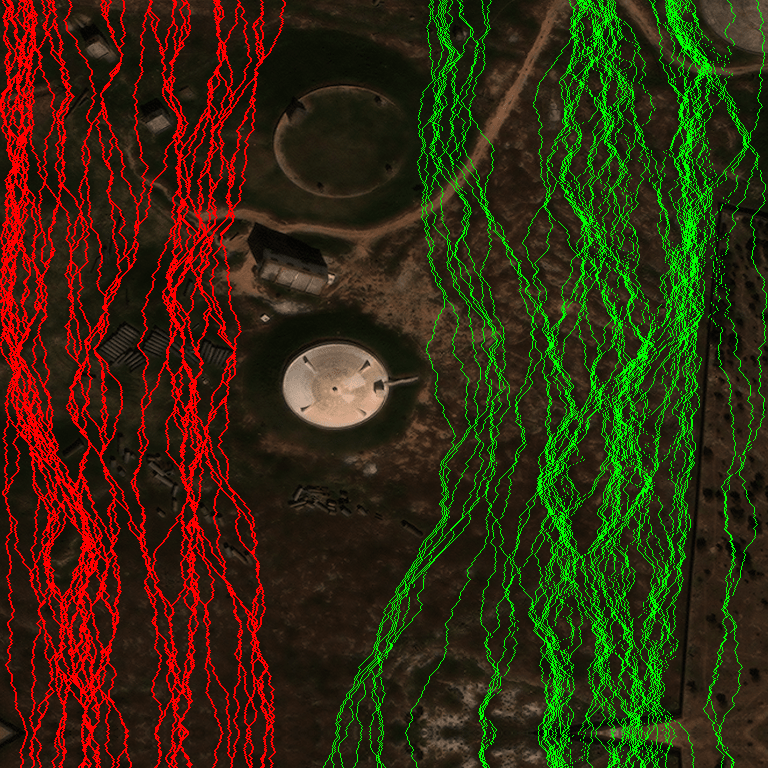}&
 \includegraphics[height=\factor\columnwidth,keepaspectratio]{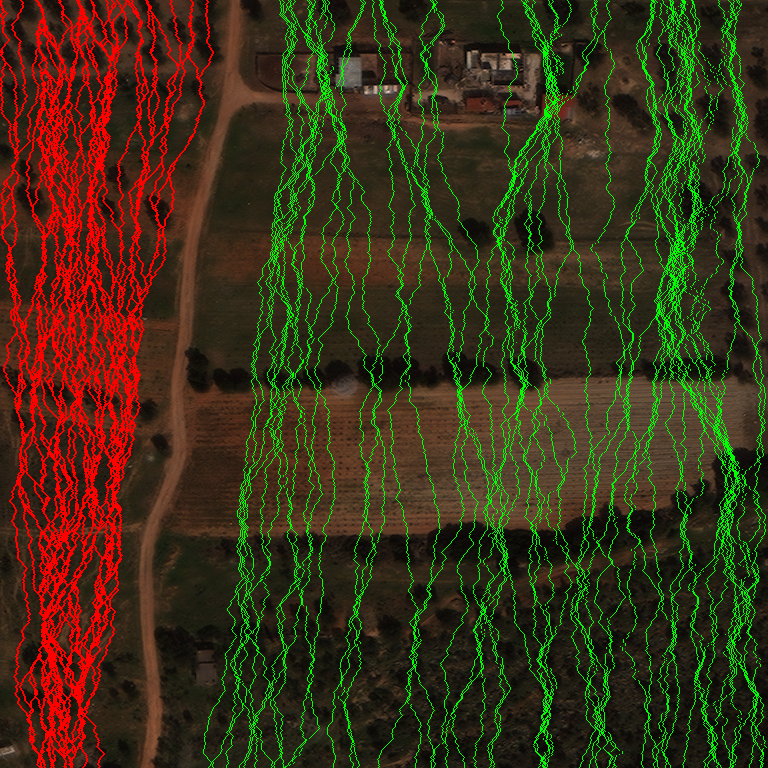}&
 \includegraphics[height=\factor\columnwidth,keepaspectratio]{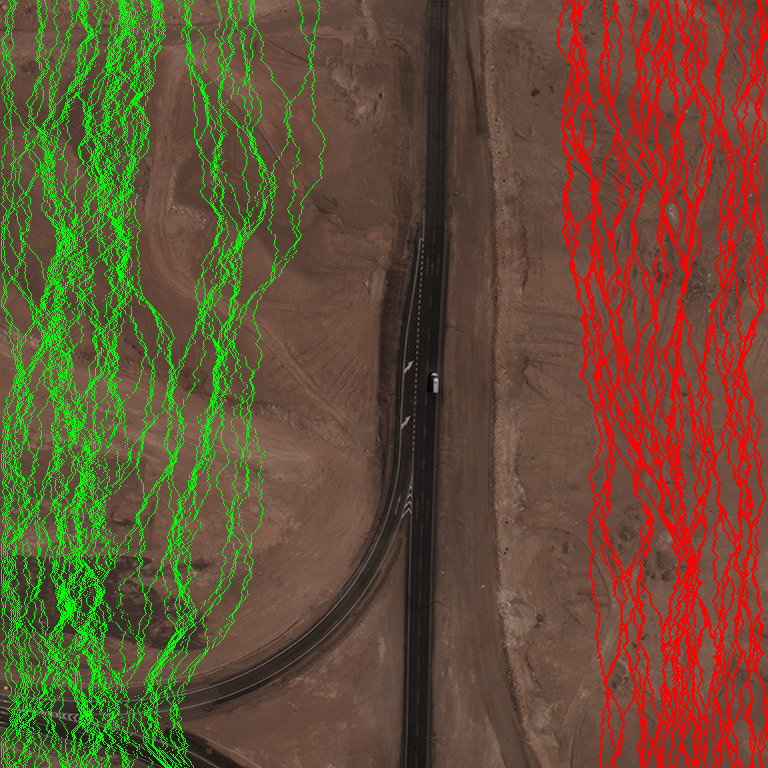}&
 \includegraphics[height=\factor\columnwidth,keepaspectratio]{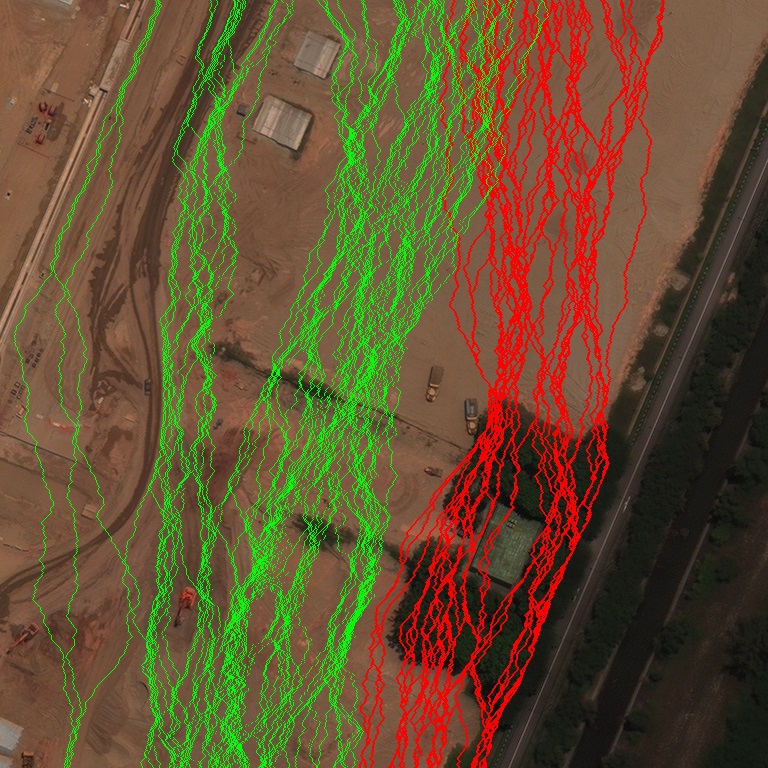}&
 \includegraphics[height=\factor\columnwidth,keepaspectratio]{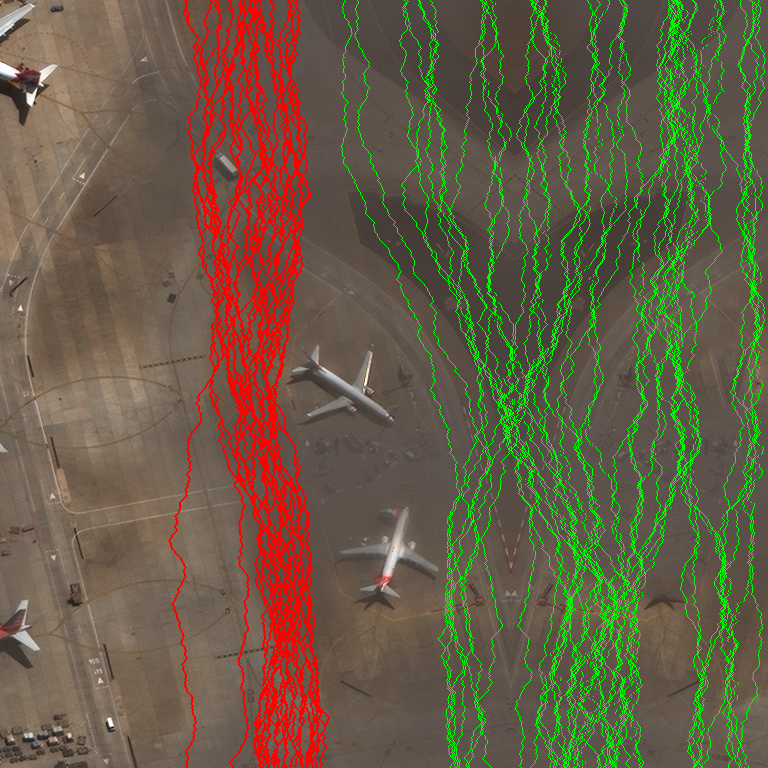}\\
 
 \includegraphics[height=\factor\columnwidth,keepaspectratio]{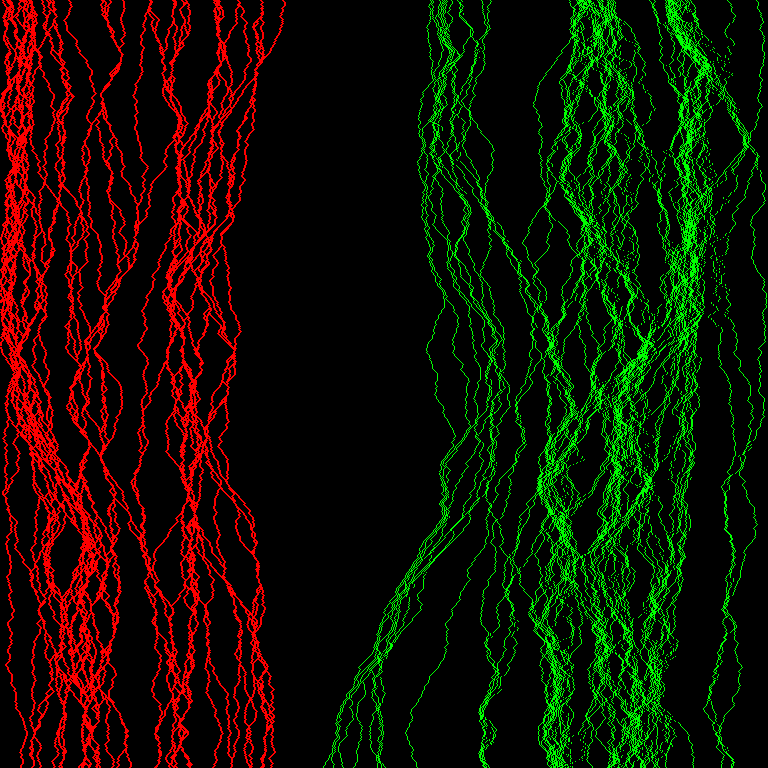}&
 \includegraphics[height=\factor\columnwidth,keepaspectratio]{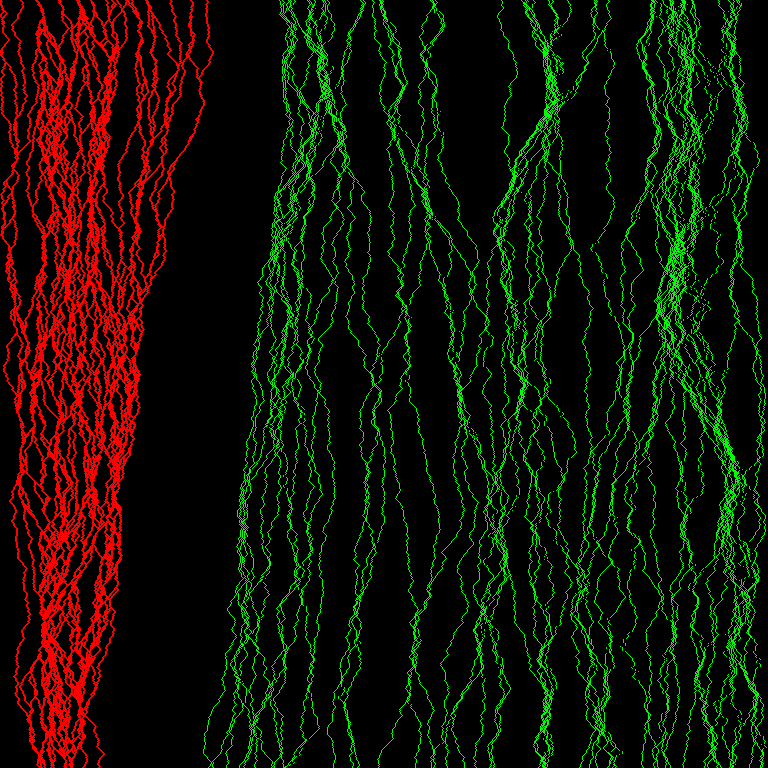}&
 \includegraphics[height=\factor\columnwidth,keepaspectratio]{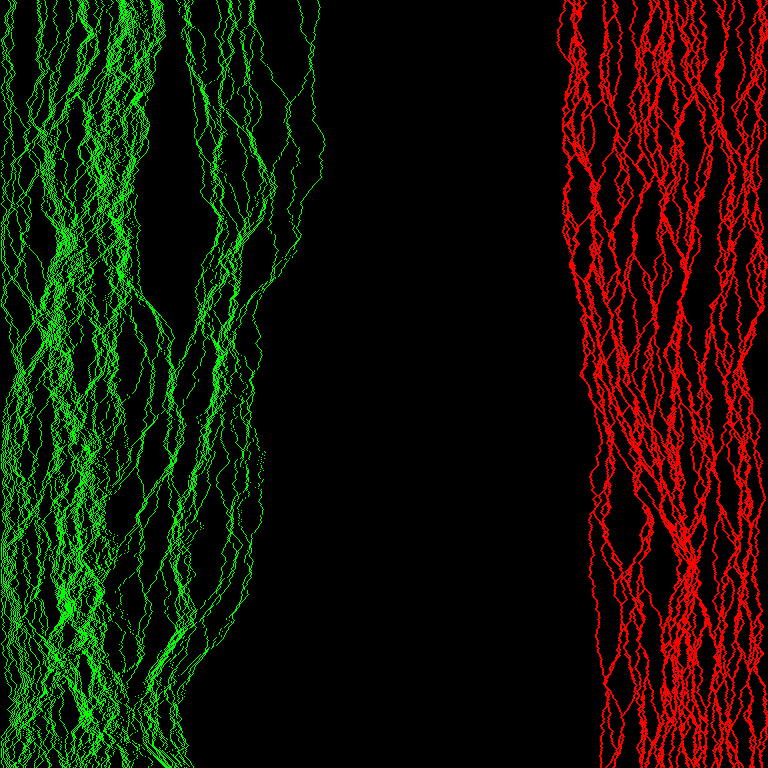}&
 \includegraphics[height=\factor\columnwidth,keepaspectratio]{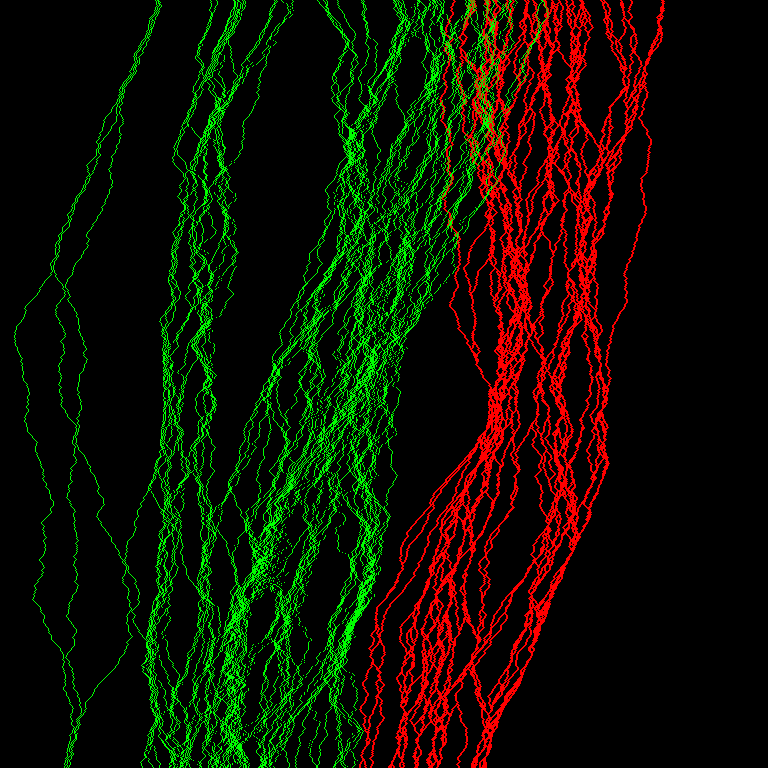}&
 \includegraphics[height=\factor\columnwidth,keepaspectratio]{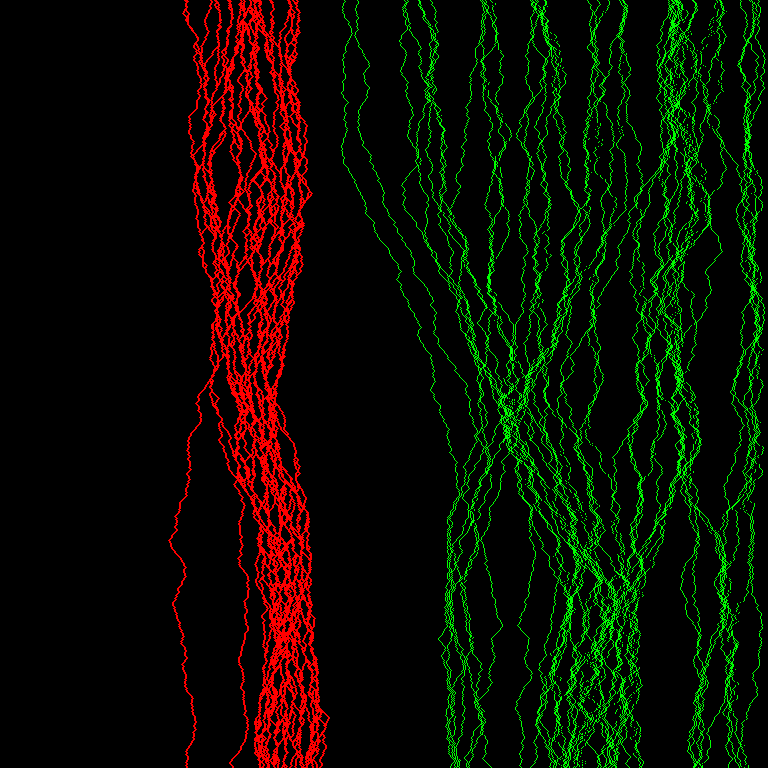} \\
 
 \includegraphics[height=\factor\columnwidth,keepaspectratio]{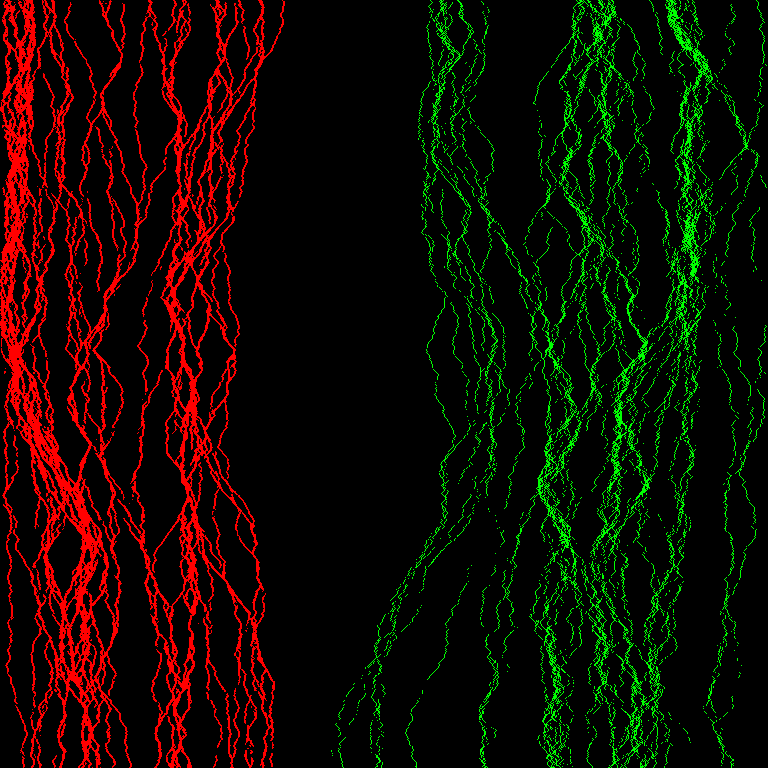}&
 \includegraphics[height=\factor\columnwidth,keepaspectratio]{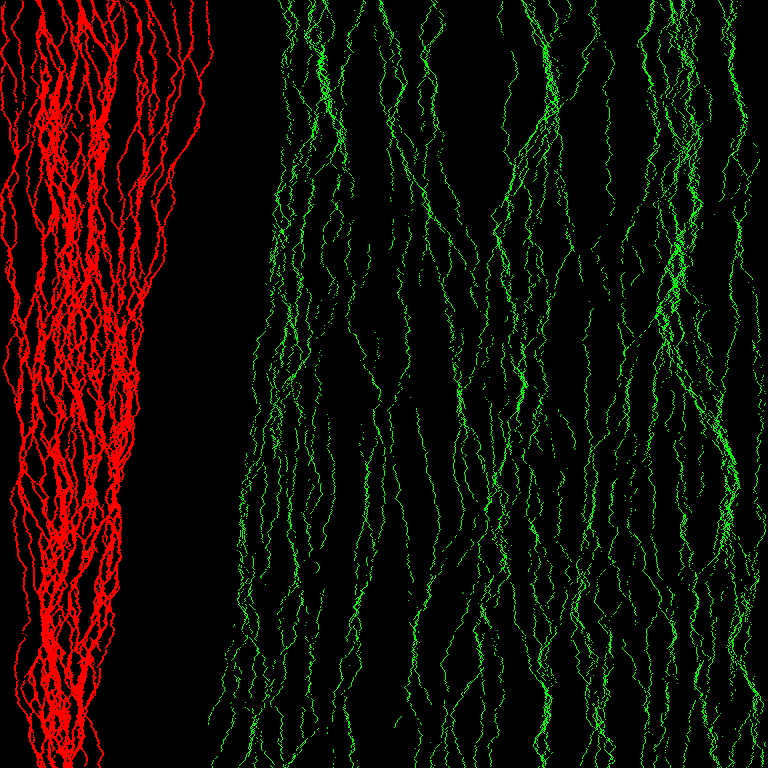}&
 \includegraphics[height=\factor\columnwidth,keepaspectratio]{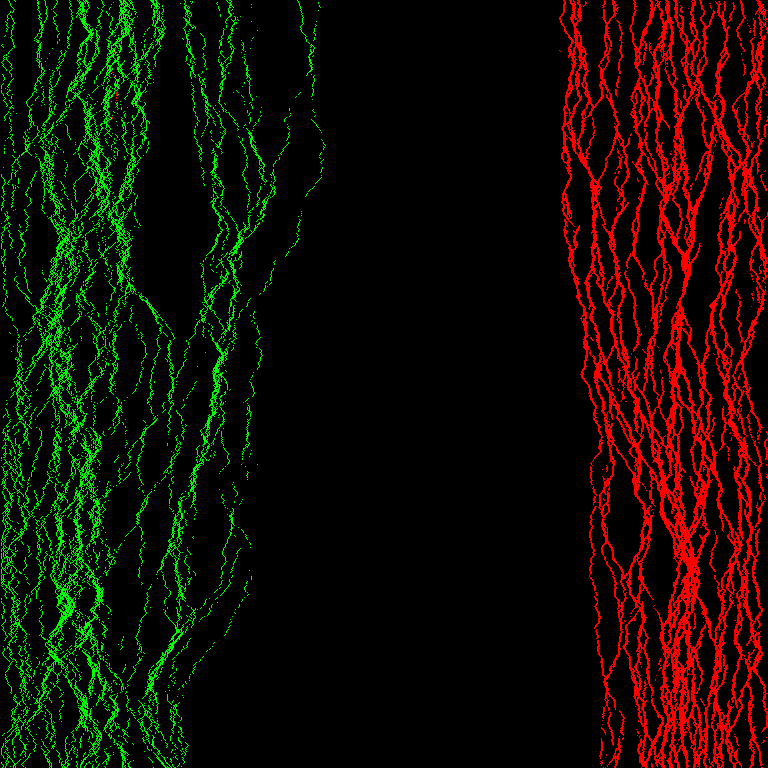}&
 \includegraphics[height=\factor\columnwidth,keepaspectratio]{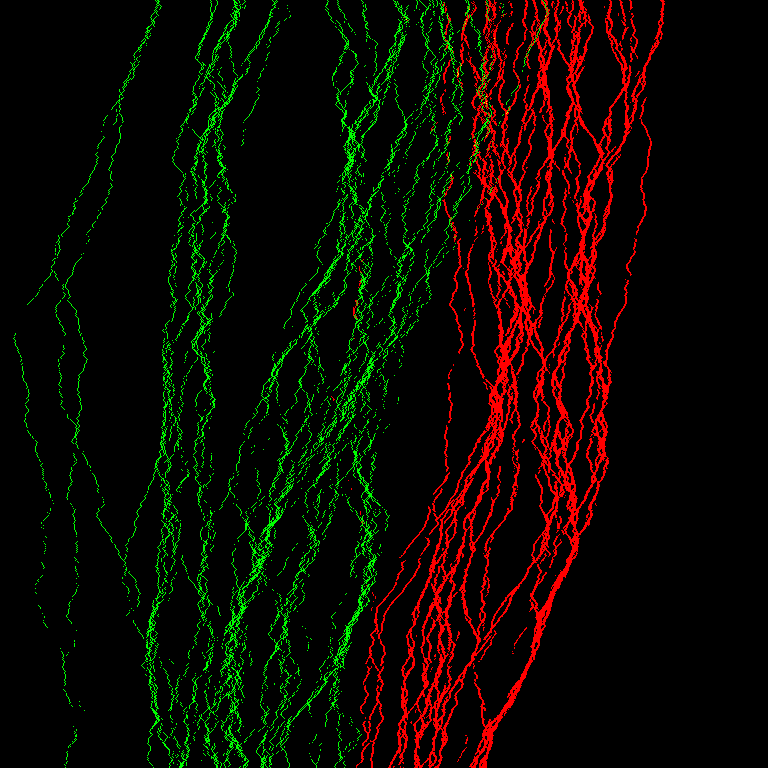}&
 \includegraphics[height=\factor\columnwidth,keepaspectratio]{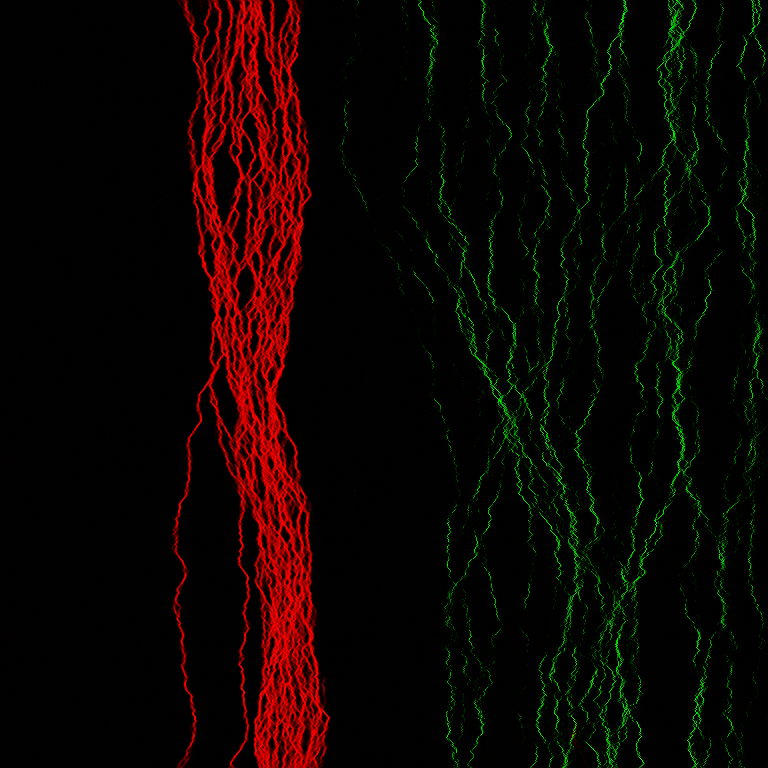}\\
 
 (a) &
 (b) &
 (c) &
 (d) &
 (e)
 
\end{tabular}

\caption{Object Dislocation Examples. From top to bottom: Pristine satellite image; Seam carved image with objects dislocated while retaining the original image size; Ground truth seam mask, with removed (red) and inserted (green) seams, overlaid on the seam carved image; Ground truth seam mask; Predicted seam mask generated by a stage 1 seam removal detector (red) and seam insertion detector (green).}

\label{fig:sc_obj_dis}
\end{figure*}

Besides content aware image resizing, seam carving can be used to remove or displace objects in a given image. In Figure \ref{fig:sc_obj_rem_supp} and Figure \ref{fig:sc_obj_dis}, we present the sample manipulations that one can do using seam carving and visualize the localization results of stage 1 pixelwise classifiers on manipulated images.

Figure \ref{fig:sc_obj_rem_supp} has five examples illustrating the application of seam carving to remove objects in satellite imagery while making sure that the seam carved satellite image looks authentic to the human eye and retains its original dimensions. In Figure \ref{fig:sc_obj_rem_supp}a, truck under the removal mask (red) is taken out and seams are inserted to restore the original image, all while leaving pixels in the protective mask (green) are left undisturbed. This is achieved by setting the energy map values at the removal mask locations to a low energy value, forcing seam carving algorithm to pass through. When inserting seams to restore the original dimensions, pixels at the locations of the protective mask are set to a high energy value, ensuring that the seam carving algorithm ignores them. Figure \ref{fig:sc_obj_rem_supp}b has an example in which an excavator under the red mask is removed from the image. Similar examples are shown in Figures \ref{fig:sc_obj_rem_supp}c, \ref{fig:sc_obj_rem_supp}d, and \ref{fig:sc_obj_rem_supp}e, where pixels under the red colored mask are removed while pixels under the green colored mask (if present) are left undisturbed.

Figure \ref{fig:sc_obj_dis} has five examples illustrating seam carving based manipulations to displace objects in a given image while retaining the visual authenticity and size of the original image. In Figure \ref{fig:sc_obj_dis}a, the white colored building at the center of the image is displaced by 50 pixels to left. This is achieved by forcing the seam carving algorithm to remove 50 seams from left side of the object and insert back same number of seams to right using removal and protective masks. In Figure \ref{fig:sc_obj_dis}b, the road is displaced by 50 pixels to left, whereas in Figure \ref{fig:sc_obj_dis}c, road is displaced by 50 pixels to right. An excavator is moved to right by 50 pixels in Figure \ref{fig:sc_obj_dis}d, and aeroplanes are moved to left by 40 pixels in Figure \ref{fig:sc_obj_dis}e.

\begin{figure*}[ht]
\begin{center}
\newcommand*{\factor}{0.48}
\subfigure{ 
\includegraphics[height=\factor\columnwidth,keepaspectratio]{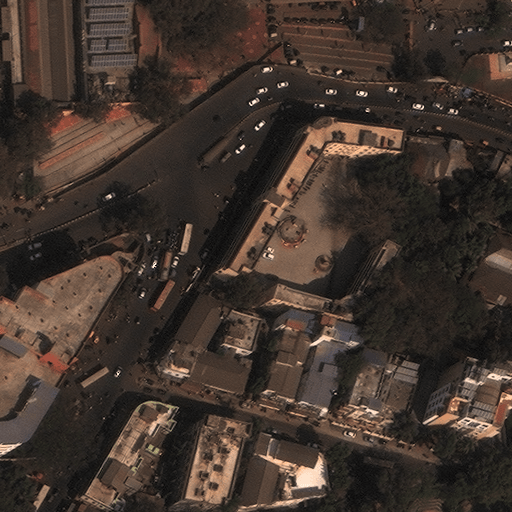}}
\subfigure{ 
\includegraphics[height=\factor\columnwidth,keepaspectratio]{SC-ICCV/images/orig_for_sc_methods.png}}
\subfigure{ 
\includegraphics[height=\factor\columnwidth,keepaspectratio]{SC-ICCV/images/orig_for_sc_methods.png}}
\subfigure{ 
\includegraphics[height=\factor\columnwidth,keepaspectratio]{SC-ICCV/images/orig_for_sc_methods.png}}
\newline
\vspace{-0.5cm}

\subfigure{ 
\includegraphics[height=\factor\columnwidth,keepaspectratio]{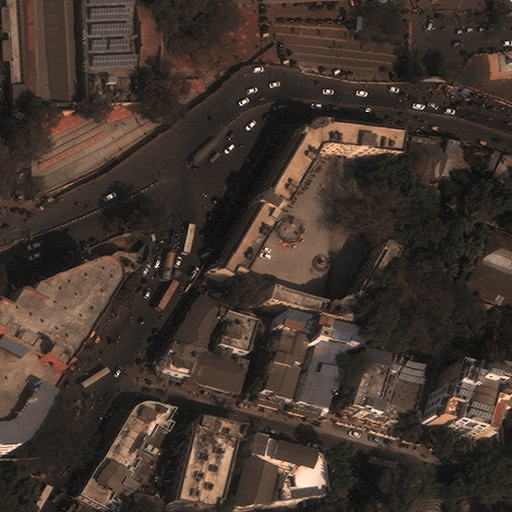}}
\subfigure{ 
\includegraphics[height=\factor\columnwidth,keepaspectratio]{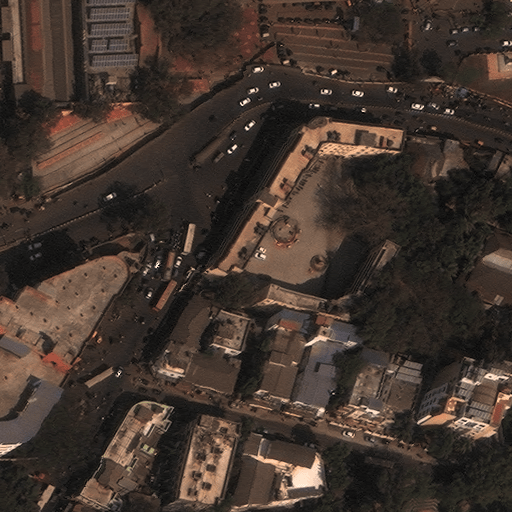}}
\subfigure{ 
\includegraphics[height=\factor\columnwidth,keepaspectratio]{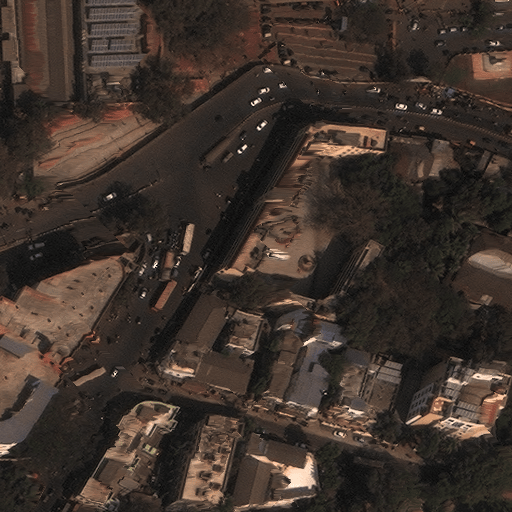}}
\subfigure{ 
\includegraphics[height=\factor\columnwidth,keepaspectratio]{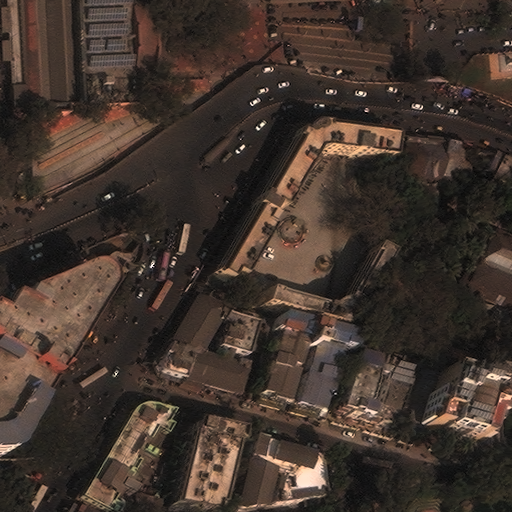}}
\newline
\vspace{-0.5cm}

\subfigure{ 
\includegraphics[height=\factor\columnwidth,keepaspectratio]{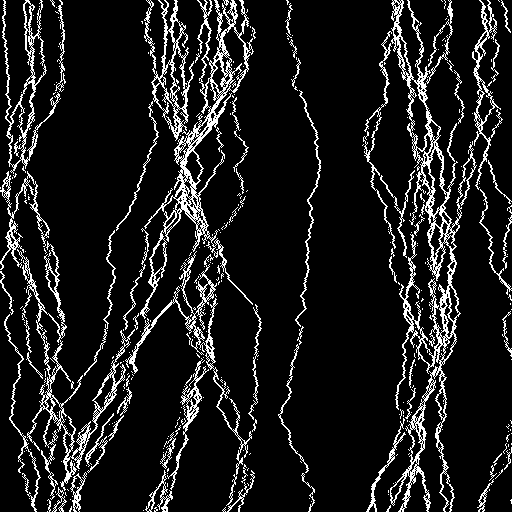}}
\subfigure{ 
\includegraphics[height=\factor\columnwidth,keepaspectratio]{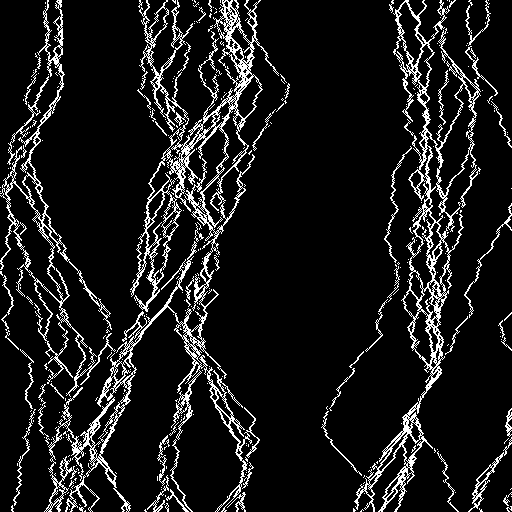}}
\subfigure{ 
\includegraphics[height=\factor\columnwidth,keepaspectratio]{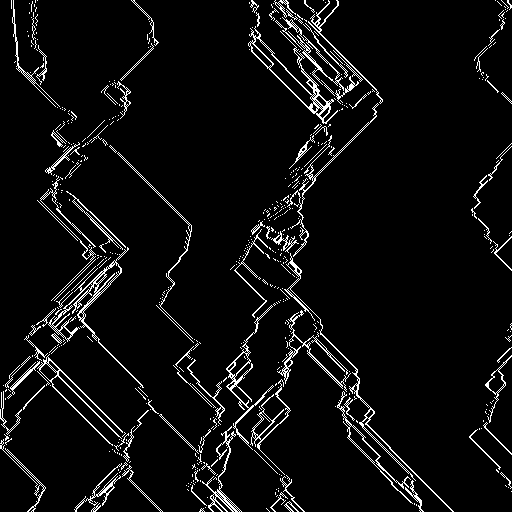}}
\subfigure{ 
\includegraphics[height=\factor\columnwidth,keepaspectratio]{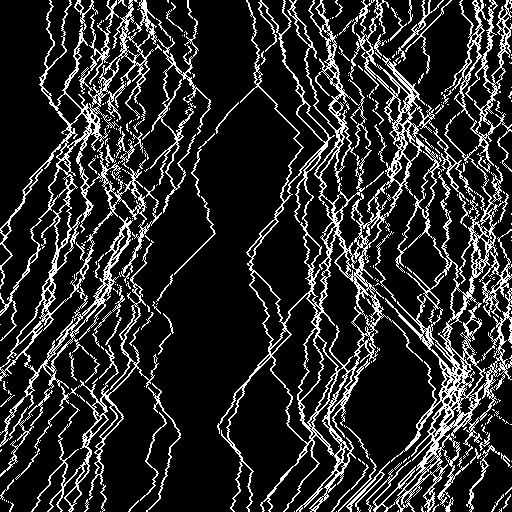}}
\newline
\vspace{-0.5cm}

\subfigure{  \includegraphics[height=\factor\columnwidth,keepaspectratio]{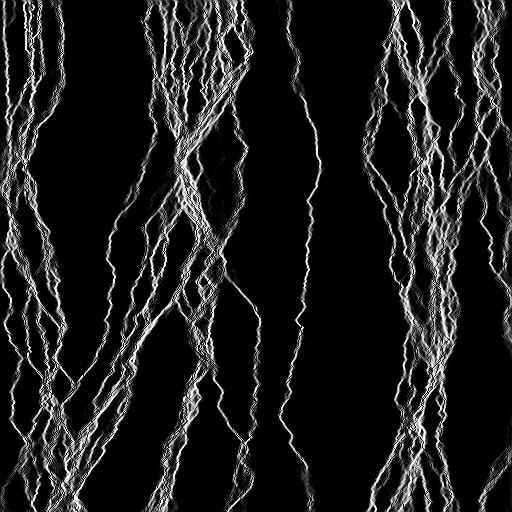}}
\subfigure{  \includegraphics[height=\factor\columnwidth,keepaspectratio]{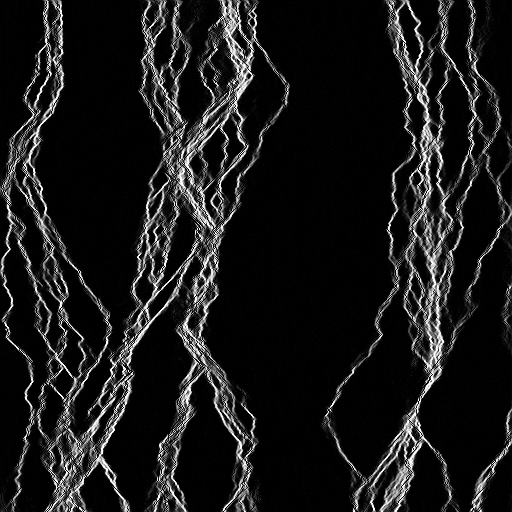}}
\subfigure{  \includegraphics[height=\factor\columnwidth,keepaspectratio]{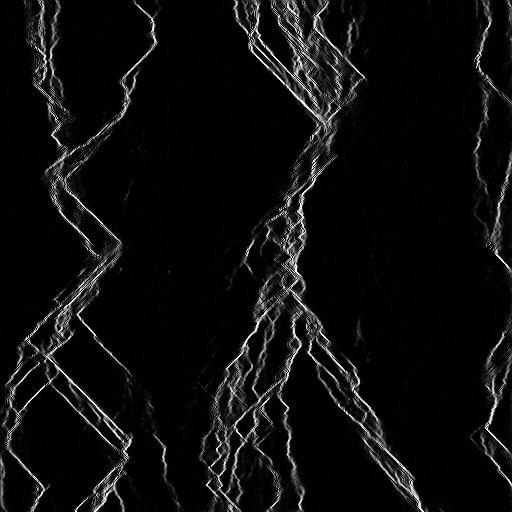}}
\subfigure{  \includegraphics[height=\factor\columnwidth,keepaspectratio]{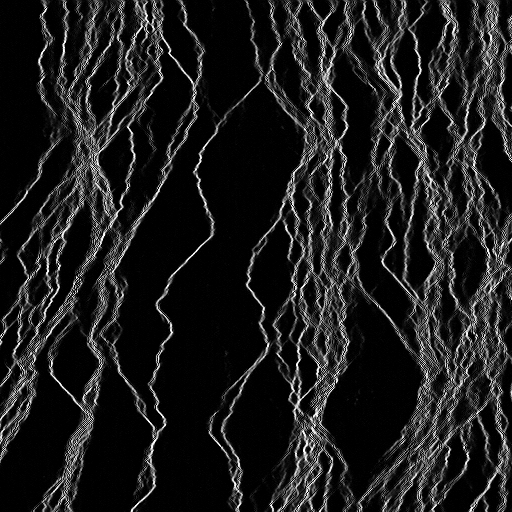}}
\newline

\end{center}
\caption{Seam Carving Ground Truth and Prediction Comparison using Various Seam Carving Algorithms. From left to right: Forward Energy, Backward Energy, Saliency Maps, Seam Merging. From top to bottom: Original image, seam carved image, ground truth seam removal mask, predicted seam removal mask}
\label{fig:sc_methods_comparison}
\end{figure*}

Figure \ref{fig:sc_methods_comparison} visualizes the distribution of removed seams in a given image that is seam carved using different seam carving algorithms described in  Section~\ref{sec:seam_carving_overview}. It can be observed that even though the seams are removed using different seam carving algorithms, our seam removal detector in stage 1 is able to predict the locations of the removed seams.


\vspace{-0.2cm}
\subsection{Comparison against Existing Approaches}
\label{sec:pxlwise_generalizability}
\vspace{-0.2cm}
We compare our method with two seam carving detection methods, that uses LBP~\cite{seam_yin2015detecting} and ILFNet~\cite{Nam2020DeepCN}. Although these methods are not specifically designed for localized detection in satellite imagery, we compare our image level stage 2 results with them as they are performing image level classification to detect seam carved images. Comparative results are shown in Table~\ref{tab:comaparative_exp}, where all of the forensic techniques have been trained on the xView training set generated as in Section~\ref{ssec:data_pxl_wise_clf}. We observe that although ILFNet achieves a comparable accuracy with the proposed method on the xView test set, it's generalizability performance drops around $5\%$ and $2\%$ when testing on xBD and Orbview-3 while our method drops less than $1\%$. Similarly, a method that utilizes local binary pattern based feature extraction combined with an SVM classifier performs reasonably well on the xView test set, but fails to generalize to other datasets.

\vspace{-0.2cm}
\begin{table}[h]
\begin{center}
\begin{tabular}{|c|c|c|c|}
\hline
\textbf{Forensic Technique} & \textbf{xView} & \textbf{xBD} & \textbf{Orbview-3} \\ \hline
LBP based detection~\cite{seam_yin2015detecting} & 91.72 & 82.71 & 78.41 \textbf{} \\ \hline
ILFNet~\cite{Nam2020DeepCN} & 99.03 & 94.96 & 96.19 \\ \hline
\textbf{Proposed method} & \textbf{99.29} & \textbf{98.86} & \textbf{98.43} \\ \hline
\end{tabular}
\end{center}
\vspace{-0.5cm}
\caption{Test accuracy(\%): Comparison with other image level classifiers, trained on xView, and tested on all datasets.}
\label{tab:comaparative_exp}
\vspace{-0.4cm}
\end{table}

\section{Conclusion}
\label{sec:conclusion}
In this paper, we proposed a method to detect and localize seam carving based manipulations in satellite images. 
We use a two stage approach that first localizes removed/inserted seams via pixelwise classification and then performs a final classification if an image has been seam carved.
We enable localization of seams as well as a generalizable framework across different datasets and seam carving techniques. Finally, we detailed the curation of three unique, large seam carving satellite image datasets that will be released to the public. 
Although the proposed method is not specifically restricted to satellite imagery, we present our findings on satellite images as a case study and leave further evaluation on more conventional images with varying compression schemes and preprocessing to be explored in future work. 

\section*{Acknowledgment}
\vspace{-0.2cm}
This material is based upon work supported by the Defense Advanced Research Projects Agency (DARPA), the National Geospatial-Intelligence Agency (NGA) and  the Air Force Research Laboratory (AFRL) 
under the contract number FA8750-16-C-0078. The views, opinions and/or findings expressed are those of the author and should not be interpreted as representing the official views or policies of the Department of Defense or the U.S. Government.

{\small
\bibliographystyle{ieee_fullname}
\bibliography{references}

\begin{thebibliography}{10}\itemsep=-1pt

\bibitem{OTHER_SC_METHODS_SALIENCY}
R. {Achanta} and S. {Süsstrunk}.
\newblock Saliency detection for content-aware image resizing.
\newblock In {\em 2009 16th IEEE International Conference on Image Processing
  (ICIP)}, pages 1005--1008, 2009.

\bibitem{ali2017identification}
Luqman Ali, Teerasit Kasetkasem, Faisal~Ghaffar Khan, Thitiporn Chanwimaluang,
  and Hiroki Nakahara.
\newblock Identification of inpainted satellite images using evalutionary
  artificial neural network (eann) and k-nearest neighbor (knn) algorithm.
\newblock In {\em 2017 8th International Conference of Information and
  Communication Technology for Embedded Systems (IC-ICTES)}, pages 1--6. IEEE,
  2017.

\bibitem{AS07}
Shai Avidan and Ariel Shamir.
\newblock Seam carving for content-aware image resizing.
\newblock {\em ACM Trans. Graph.}, 26(3):10, 2007.

\bibitem{baheti2020eff}
Bhakti Baheti, Shubham Innani, Suhas Gajre, and Sanjay Talbar.
\newblock Eff-unet: A novel architecture for semantic segmentation in
  unstructured environment.
\newblock In {\em Proceedings of the IEEE/CVF Conference on Computer Vision and
  Pattern Recognition Workshops}, pages 358--359, 2020.

\bibitem{sat_bartusiak2019splicing}
Emily~R Bartusiak, Sri~Kalyan Yarlagadda, David G{\"u}era, Paolo Bestagini,
  Stefano Tubaro, Fengqing~M Zhu, and Edward~J Delp.
\newblock Splicing detection and localization in satellite imagery using
  conditional gans.
\newblock In {\em 2019 IEEE Conference on Multimedia Information Processing and
  Retrieval (MIPR)}, pages 91--96. IEEE, 2019.

\bibitem{bayar2016deep}
Belhassen Bayar and Matthew~C Stamm.
\newblock A deep learning approach to universal image manipulation detection
  using a new convolutional layer.
\newblock In {\em Proceedings of the 4th ACM Workshop on Information Hiding and
  Multimedia Security}, pages 5--10, 2016.

\bibitem{nonsat_bondi2016first}
Luca Bondi, Luca Baroffio, David G{\"u}era, Paolo Bestagini, Edward~J Delp, and
  Stefano Tubaro.
\newblock First steps toward camera model identification with convolutional
  neural networks.
\newblock {\em IEEE Signal Processing Letters}, 24(3):259--263, 2016.

\bibitem{nonsat_bondi2017tampering}
Luca Bondi, Silvia Lameri, David Guera, Paolo Bestagini, Edward~J Delp, Stefano
  Tubaro, et~al.
\newblock Tampering detection and localization through clustering of
  camera-based cnn features.
\newblock In {\em CVPR Workshops}, pages 1855--1864, 2017.

\bibitem{sat_motivation_india}
Deborah Byrd.
\newblock Fake image of diwali still circulating.
\newblock
  https://earthsky.org/earth/fake-image-of-india-during-diwali-versus-the-real-thing,
  2018.
\newblock Accessed: 2021-03-16.

\bibitem{MCCADVANTAGES}
Davide Chicco and Giuseppe Jurman.
\newblock The advantages of the matthews correlation coefficient (mcc) over f1
  score and accuracy in binary classification evaluation.
\newblock {\em BMC Genomics}, 21, 2020.

\bibitem{cieslak2018seam}
Luiz Fernandoda~Silva Cieslak, Kelton~Augustopontara Da~Costa, and Joao
  PauloPapa.
\newblock Seam carving detection using convolutional neural networks.
\newblock In {\em 2018 IEEE 12th International Symposium on Applied
  Computational Intelligence and Informatics (SACI)}, pages 000195--000200.
  Ieee, 2018.

\bibitem{cozzolino2015efficient}
Davide Cozzolino, Giovanni Poggi, and Luisa Verdoliva.
\newblock Efficient dense-field copy--move forgery detection.
\newblock {\em IEEE Transactions on Information Forensics and Security},
  10(11):2284--2297, 2015.

\bibitem{sat_motivation_china}
Jane Edwards.
\newblock Nga’s todd myers: China uses gan technique to tamper with earth
  images.
\newblock
  https://www.executivegov.com/2019/04/ngas-todd-myers-china-uses-gan-technique-to-tamper-with-earth-images/,
  2019.
\newblock Accessed: 2021-03-16.

\bibitem{farid2009exposing}
Hany Farid.
\newblock Exposing digital forgeries from jpeg ghosts.
\newblock {\em IEEE transactions on information forensics and security},
  4(1):154--160, 2009.

\bibitem{fei2015detection}
Wei Fei, Yang Gaobo, Li Leida, Xia Ming, and Zhang Dengyong.
\newblock Detection of seam carving-based video retargeting using forensics
  hash.
\newblock {\em Security and Communication Networks}, 8(12):2102--2113, 2015.

\bibitem{fouad2020detection}
Mohamed~Mahmoud Fouad, Eslam~Magdy Mostafa, and Mohamed~Abdelmoneim Elshafey.
\newblock Detection and localization enhancement for satellite images with
  small forgeries using modified gan-based cnn structure.
\newblock {\em International Journal of Advances in Intelligent Informatics},
  6(3):278--289, 2020.

\bibitem{orbview3}
GeoEye.
\newblock Usgs eros archive - commercial satellites - orbview 3.
\newblock \url{https://doi.org/10.5066/F7J38R0R}, (accessed Oct 24, 2019).

\bibitem{gong2018detecting}
Qingmei Gong, Qingqing Shan, Yongzhen Ke, and Jing Guo.
\newblock Detecting the location of seam and recovering image for seam inserted
  image.
\newblock {\em Journal of Computational Methods in Sciences and Engineering},
  18(2):499--509, 2018.

\bibitem{guillemot2014image}
Christine Guillemot and Olivier Le~Meur.
\newblock Image inpainting: Overview and recent advances.
\newblock {\em Signal processing magazine}, 31(1):127--144, 2014.

\bibitem{XVIEW2}
Ritwik Gupta, Richard Hosfelt, Sandra Sajeev, Nirav Patel, Bryce Goodman, Jigar
  Doshi, Eric~T. Heim, Howie Choset, and Matthew~E. Gaston.
\newblock xbd: {A} dataset for assessing building damage from satellite
  imagery.
\newblock {\em CoRR}, abs/1911.09296, 2019.

\bibitem{han2018exploring}
Rong Han, Yongzhen Ke, Ling Du, Fan Qin, and Jing Guo.
\newblock Exploring the location of object deleted by seam-carving.
\newblock {\em Expert Systems with Applications}, 95:162--171, 2018.

\bibitem{resnet}
Kaiming He, Xiangyu Zhang, Shaoqing Ren, and Jian Sun.
\newblock Deep residual learning for image recognition.
\newblock {\em CoRR}, abs/1512.03385, 2015.

\bibitem{sat_ho_1525212}
A.~T.~S. {Ho}, {Xunzhan Zhu}, and W.~M. {Woon}.
\newblock A semi-fragile pinned sine transform watermarking system for content
  authentication of satellite images.
\newblock In {\em Proceedings. 2005 IEEE International Geoscience and Remote
  Sensing Symposium, 2005. IGARSS '05.}, volume~2, pages 4 pp.--, 2005.

\bibitem{sat_horvath2019anomaly}
J{\'a}nos Horv{\'a}th, David G{\"u}era, Sri~Kalyan Yarlagadda, Paolo Bestagini,
  Fengqing~Maggie Zhu, Stefano Tubaro, and Edward~J Delp.
\newblock Anomaly-based manipulation detection in satellite images.
\newblock {\em networks}, 29:21, 2019.

\bibitem{jassim2018automatic}
Sabah Jassim and Aras Asaad.
\newblock Automatic detection of image morphing by topology-based analysis.
\newblock In {\em 2018 26th European Signal Processing Conference (EUSIPCO)},
  pages 1007--1011. IEEE, 2018.

\bibitem{sat_motivation_russia}
Andrew Kramer.
\newblock Russian images of malaysia airlines flight 17 were altered, report
  finds.
\newblock
  https://www.nytimes.com/2016/07/16/world/europe/malaysia-airlines-flight-17-russia.html,
  2016.
\newblock Accessed: 2021-03-16.

\bibitem{XVIEW}
Darius Lam, Richard Kuzma, Kevin McGee, Samuel Dooley, Michael Laielli, Matthew
  Klaric, Yaroslav Bulatov, and Brendan McCord.
\newblock xview: Objects in context in overhead imagery.
\newblock {\em CoRR}, abs/1802.07856, 2018.

\bibitem{li2015segmentation}
Jian Li, Xiaolong Li, Bin Yang, and Xingming Sun.
\newblock Segmentation-based image copy-move forgery detection scheme.
\newblock {\em IEEE Transactions on Information Forensics and Security},
  10(3):507--518, 2015.

\bibitem{li2020identification}
Yanan Li, Ming Xia, Xin Liu, and Gaobo Yang.
\newblock Identification of various image retargeting techniques using hybrid
  features.
\newblock {\em Journal of Information Security and Applications}, 51:102459,
  2020.

\bibitem{liang2015efficient}
Zaoshan Liang, Gaobo Yang, Xiangling Ding, and Leida Li.
\newblock An efficient forgery detection algorithm for object removal by
  exemplar-based image inpainting.
\newblock {\em Journal of Visual Communication and Image Representation},
  30:75--85, 2015.

\bibitem{gimp_sc}
{Liquid Rescale GIMP plugin}.
\newblock http://liquidrescale.wikidot.com/en:examples.
\newblock Accessed: 2021-03-16.

\bibitem{lu2011seam}
Wenjun Lu and Min Wu.
\newblock Seam carving estimation using forensic hash.
\newblock In {\em 13th ACM multimedia workshop on Multimedia and security},
  pages 9--14, 2011.

\bibitem{luo2010jpeg}
Weiqi Luo, Jiwu Huang, and Guoping Qiu.
\newblock Jpeg error analysis and its applications to digital image forensics.
\newblock {\em IEEE Transactions on Information Forensics and Security},
  5(3):480--491, 2010.

\bibitem{forgery_2}
Babak Mahdian and Stanislav Saic.
\newblock A bibliography on blind methods for identifying image forgery.
\newblock {\em Signal Processing: Image Communication}, 25(6):389--399, 2010.

\bibitem{mishiba2012image}
Kazu Mishiba and Masaaki Ikehara.
\newblock Image resizing using improved seam merging.
\newblock In {\em 2012 IEEE International Conference on Acoustics, Speech and
  Signal Processing (ICASSP)}, pages 1261--1264. IEEE, 2012.

\bibitem{mnih2010learning}
Volodymyr Mnih and Geoffrey~E Hinton.
\newblock Learning to detect roads in high-resolution aerial images.
\newblock In {\em European Conference on Computer Vision}, pages 210--223.
  Springer, 2010.

\bibitem{montserrat2020generative}
Daniel~Mas Montserrat, J{\'a}nos Horv{\'a}th, SK Yarlagadda, Fengqing Zhu, and
  Edward~J Delp.
\newblock Generative autoregressive ensembles for satellite imagery
  manipulation detection.
\newblock {\em arXiv preprint arXiv:2010.03758}, 2020.

\bibitem{nam2019content}
Seung-Hun Nam, Wonhyuk Ahn, Seung-Min Mun, Jinseok Park, Dongkyu Kim, In-Jae
  Yu, and Heung-Kyu Lee.
\newblock Content-aware image resizing detection using deep neural network.
\newblock In {\em 2019 IEEE International Conference on Image Processing
  (ICIP)}, pages 106--110. IEEE, 2019.

\bibitem{Nam2020DeepCN}
Seung-Hun Nam, Wonhyuk Ahn, In-Jae Yu, Myung-Joon Kwon, M. Son, and H. Lee.
\newblock Deep convolutional neural network for identifying seam-carving
  forgery.
\newblock {\em ArXiv}, abs/2007.02393, 2020.

\bibitem{nataraj2021seam}
Lakshmanan Nataraj, Chandrakanth Gudavalli, Tajuddin~Manhar Mohammed, Shivkumar
  Chandrasekaran, and BS Manjunath.
\newblock Seam carving detection and localization using two-stage deep neural
  networks.
\newblock In {\em Machine Learning, Deep Learning and Computational
  Intelligence for Wireless Communication}, pages 381--394. Springer, 2021.

\bibitem{neubert2017face}
Tom Neubert.
\newblock Face morphing detection: An approach based on image degradation
  analysis.
\newblock In {\em International Workshop on Digital Watermarking}, pages
  93--106. Springer, 2017.

\bibitem{popescu-farid-resampling}
Alin~C Popescu and Hany Farid.
\newblock Exposing digital forgeries by detecting traces of resampling.
\newblock {\em IEEE Transactions on signal processing}, 53(2):758--767, 2005.

\bibitem{rao2016deep}
Yuan Rao and Jiangqun Ni.
\newblock A deep learning approach to detection of splicing and copy-move
  forgeries in images.
\newblock In {\em Information Forensics and Security (WIFS), International
  Workshop on}, pages 1--6. IEEE, 2016.

\bibitem{UNET}
Olaf Ronneberger, Philipp Fischer, and Thomas Brox.
\newblock U-net: Convolutional networks for biomedical image segmentation.
\newblock {\em CoRR}, abs/1505.04597, 2015.

\bibitem{Rubinstein08}
Michael Rubinstein, Ariel Shamir, and Shai Avidan.
\newblock Improved seam carving for video retargeting.
\newblock {\em ACM Transactions on Graphics (SIGGRAPH)}, 27(3):1--9, 2008.

\bibitem{ryu2014estimation}
Seung-Jin Ryu and Heung-Kyu Lee.
\newblock Estimation of linear transformation by analyzing the periodicity of
  interpolation.
\newblock {\em Pattern Recognition Letters}, 36:89--99, 2014.

\bibitem{salloum2018image}
Ronald Salloum, Yuzhuo Ren, and C-C~Jay Kuo.
\newblock Image splicing localization using a multi-task fully convolutional
  network (mfcn).
\newblock {\em Journal of Visual Communication and Image Representation},
  51:201--209, 2018.

\bibitem{sarkar2009detection}
Anindya Sarkar, Lakshmanan Nataraj, and Bangalore~S Manjunath.
\newblock Detection of seam carving and localization of seam insertions in
  digital images.
\newblock In {\em 11th ACM workshop on Multimedia and security}, pages
  107--116. ACM, 2009.

\bibitem{efficientnet}
Mingxing Tan and Quoc~V. Le.
\newblock Efficientnet: Rethinking model scaling for convolutional neural
  networks.
\newblock {\em CoRR}, abs/1905.11946, 2019.

\bibitem{sat_van2018you}
Adam Van~Etten.
\newblock You only look twice: Rapid multi-scale object detection in satellite
  imagery.
\newblock {\em arXiv preprint arXiv:1805.09512}, 2018.

\bibitem{forgery_5}
Savita Walia and Krishan Kumar.
\newblock Digital image forgery detection: a systematic scrutiny.
\newblock {\em Australian Journal of Forensic Sciences}, pages 1--39, 2018.

\bibitem{wegner2013higher}
Jan~D Wegner, Javier~A Montoya-Zegarra, and Konrad Schindler.
\newblock A higher-order crf model for road network extraction.
\newblock In {\em IEEE Conference on Computer Vision and Pattern Recognition},
  pages 1698--1705, 2013.

\bibitem{photoshop_sc}
What's new in adobe photoshop cs4 - photoshop 11.
\newblock
  http://www.photoshopsupport.com/photoshop-cs4/what-is-new-in-photoshop-cs4.html.
\newblock Accessed: 2021-03-16.

\bibitem{wu2008detection}
Qiong Wu, Shao-Jie Sun, Wei Zhu, Guo-Hui Li, and Dan Tu.
\newblock Detection of digital doctoring in exemplar-based inpainted images.
\newblock In {\em Machine Learning and Cybernetics, 2008 International
  Conference on}, volume~3, pages 1222--1226. IEEE, 2008.

\bibitem{sat_yarlagadda2018satellite}
Sri~Kalyan Yarlagadda, David G{\"u}era, Paolo Bestagini, Fengqing Maggie~Zhu,
  Stefano Tubaro, and Edward~J Delp.
\newblock Satellite image forgery detection and localization using gan and
  one-class classifier.
\newblock {\em Electronic Imaging}, 2018(7):214--1, 2018.

\bibitem{ye2018convolutional}
Jingyu Ye, Yuxi Shi, Guanshuo Xu, and Yun-Qing Shi.
\newblock A convolutional neural network based seam carving detection scheme
  for uncompressed digital images.
\newblock In {\em International Workshop on Digital Watermarking}, pages 3--13.
  Springer, 2018.

\bibitem{seam_yin2015detecting}
Ting Yin, Gaobo Yang, Leida Li, Dengyong Zhang, and Xingming Sun.
\newblock Detecting seam carving based image resizing using local binary
  patterns.
\newblock {\em computers \& security}, 55:130--141, 2015.

\bibitem{zhang2017detection}
Dengyong Zhang, Qingguo Li, Gaobo Yang, Leida Li, and Xingming Sun.
\newblock Detection of image seam carving by using weber local descriptor and
  local binary patterns.
\newblock {\em Journal of information security and applications}, 36:135--144,
  2017.

\bibitem{zhang2020detecting}
Dengyong Zhang, Gaobo Yang, Feng Li, Jin Wang, and Arun~Kumar Sangaiah.
\newblock Detecting seam carved images using uniform local binary patterns.
\newblock {\em Multimedia Tools and Applications}, 79(13):8415--8430, 2020.

\end{thebibliography}
}

\end{document}